\documentclass[11pt]{article}

\usepackage[final]{acl}

\usepackage{times}
\usepackage{latexsym}

\usepackage[T1]{fontenc}

\usepackage[utf8]{inputenc}

\usepackage{microtype}

\usepackage{inconsolata}

\usepackage{graphicx}
\usepackage{rotating}
\usepackage{placeins}
\usepackage{listings}
\usepackage{colortbl}

\usepackage{multirow}

\usepackage{booktabs}
\usepackage{pifont}
\lstdefinestyle{prompttemplate}{
  basicstyle=\scriptsize\ttfamily,
  breaklines=true,
  breakatwhitespace=false,
  columns=fullflexible,
  keepspaces=true,
  showstringspaces=false,
  frame=single,
  framerule=0.4pt,
  framesep=3pt,
  aboveskip=0.5\baselineskip,
  belowskip=0.5\baselineskip,
  literate={é}{{\'e}}1 {è}{{\`e}}1 {à}{{\`a}}1
}
\newcommand{\prompttemplate}[1]{\lstinputlisting[style=prompttemplate]{#1}}
\newcommand{\cmark}{\ding{51}}%
\newcommand{\xmark}{~}%
\newcommand{\ie}{i.e.,~}

\newcommand{\LangUnd}{\textbf{U}}      
\newcommand{\LangGen}{\textbf{G}}      
\newcommand{\Knowledge}{\textbf{K}}    
\newcommand{\Chat}{\textbf{Ch}}         
\newcommand{\ReasonMath}{\textbf{R}}   
\newcommand{\CodeSkill}{\textbf{Co}} 
\newcommand{\StructFormat}{\textbf{Fo}} 
\newcommand{\InstrFollow}{\textbf{I}}  
\newcommand{\FuncCall}{\textbf{Fu}}     
\newcommand{\LongContext}{\textbf{L}}  
\newcommand{\SafetyEthics}{\textbf{S}} 

\newcommand{\sr}[2]{#1 {\tiny (#2)}}

\definecolor{scoreTopBg}{HTML}{DFF2E5}
\definecolor{scoreHighBg}{HTML}{E8F1FA}
\definecolor{scoreMidBg}{HTML}{FFF4CF}
\definecolor{scoreLowBg}{HTML}{FBE4E4}
\newcommand{\scoretop}[1]{\cellcolor{scoreTopBg}#1}
\newcommand{\scorehigh}[1]{\cellcolor{scoreHighBg}#1}
\newcommand{\scoremid}[1]{\cellcolor{scoreMidBg}#1}
\newcommand{\scorelow}[1]{\cellcolor{scoreLowBg}#1}
\newcommand{\scorecolorlegend}{Red, yellow, blue, and green indicate ascending position in each column range w.r.t. slices of 15\%, 35\%, 35\%, and 15\%.}

\newsavebox{\fitwidthtablebox}
\newenvironment{fitwidthtable}[1][\linewidth]{%
  \def\fitwidthtablewidth{#1}%
  \begin{lrbox}{\fitwidthtablebox}%
  \setlength{\tabcolsep}{0pt}%
}{%
  \end{lrbox}%
  \ifdim\wd\fitwidthtablebox>\fitwidthtablewidth
    \resizebox{\fitwidthtablewidth}{!}{\usebox{\fitwidthtablebox}}%
  \else
    \usebox{\fitwidthtablebox}%
  \fi
}

\title{Wolof-Adapted Industry-Ready Open-Weight Language Models}

\author{
{\bf Gwénolé Lecorvé},
{\bf Delphine Charlet},
{\bf Pierre Adam},
{\bf Floriane Behanzin} \\
{\bf Lionel Delphin-Poulat},
{\bf Johannes Heinecke},
{\bf Géraldine Damnati},
\\
{\bf Pelin Kaytmaz},
{\bf Noureddine Khaous},
{\bf Charbel Kindji},
{\bf Meriam Maadi},
\\
{\bf Abdoulaye Mbaye},
{\bf Claire Perroux},
{\bf Ismaël Rousseau},
{\bf Papa Séga Wadé},
\\
{\bf Gilles Xiberras},
{\bf Olivier Zhang},
{\bf Malek Zitouni}
\\
Orange Research, France
\\
{\small Corresponding author: \texttt{gwenole.lecorve@orange.com}}}

\begin{document}
\maketitle

\begin{abstract}
Wolof is spoken by 23M people, mostly in Senegal, but remains weakly supported by open-weight language models, while API models raise cost, governance, and sovereignty constraints for local deployments. In this paper, we present 0.6B-14B open-weight models adapted to Wolof through instruction tuning, preference alignment, and model merging. After presenting the pipeline and data, extensive experiments on 63~public and in-house tasks show good performance on Wolof understanding and generation while remaining effective on other languages and cross-lingual skills, with the 8B and 14B models being particularly competitive with substantially larger commercial systems. The models are currently being evaluated in internal customer-care pilots and are publicly available\footnote{\url{https://huggingface.co/collections/Orange/wolof-lm}}.
\end{abstract}

\section{Introduction}

Wolof---a West-Atlantic language of the Niger-Congo family---is the lingua franca of Senegal with 23M speakers, and the native language for about 40\% of the population~\cite{mbaye2026nlp-senegalese-languages}. However, it remains under-served in current Large Language Models (LLMs) compared to English, French, and better-covered African languages \citep{hussen2026state-llms-african,adebara2025evaluate-african-llms}.
This limits applications in banking, telecommunications, and public services where users expect Wolof support.
Commercial API-based LLMs show some Wolof capability but are often too costly or incompatible with local data-governance and sovereignty requirements. This paper addresses this gap by adapting open-weight small language models to Wolof and evaluating them on industrially realistic use cases. A representative application is customer-relationship automation, spanning offline conversation analysis, FAQ answering, transactional support, and interactive chat. The models are currently being integrated into internal Wolof customer-care pilots, with a target customer base of several millions and the goal of reducing dependence on commercial systems. Adaptation pipeline mostly relies on Wolof-centric instruction tuning, which, depending on the model size, is complemented with preference alignment and model merging to balance the acquired Wolof capabilities with the pre-existing ones in the base models. Although these adaptation stages are well known in the literature, applying them to a severely under-resourced language requires substantial data engineering, empirical tuning, and a rigorous targeted evaluation methodology. Hence, the main contributions are: (\emph{i}) we provide a suite of Wolof-adapted Qwen models (from 0.6 to 14B) for deployment-constrained settings; (\emph{ii}) we propose a broad evaluation covering public and in-house tasks, with skill-based aggregations and French/English regression checks for multilingual operations; and (\emph{iii}) we release associated Wolof data: 222K instructions and some of the in-house test sets from our evaluation.

The paper is organised as follows: Section~\ref{sec:background} reviews African LLMs and Wolof/Senegalese Natural Language Processing (NLP); Section~\ref{sec:method} details the adaptation method; Section~\ref{sec:benchmark} introduces our evaluation benchmark while Section~\ref{sec:results} reports automatic and human evaluation results.

\section{Related Work}
\label{sec:background}

Recent African LLM surveys and benchmark studies report persistent coverage, tokenization, and evaluation gaps, with Wolof still rarely evaluated systematically \citep{hussen2026state-llms-african,adebara2025evaluate-african-llms,arxiv2024bridging-gap-africa}. Recent Wolof and Senegalese-language resources remain mostly task-specific, including banking-domain intent classification \citep{kandji2026wolbanking} and resource surveys \citep{mbaye2026nlp-senegalese-languages}. This leaves a gap for openly documented Wolof models designed and evaluated for general-purpose industrial use.

Regarding language modeling, existing work explores both training compact models from scratch and adapting pretrained models. InkubaLM follows the former approach for five African languages \citep{tonja2024inkubalm}. Aya uses multilingual instruction tuning and studies the composition of instruction data \citep{singh-etal-2024-aya,ustun2024aya}, while AfriqueLLM and Lugha-Llama investigate continued pre-training and the effect of data composition \citep{yu-etal-2026-afriquellm,buzaaba2025lugha}. DIPLomA instead combines target-language continued pre-training with delta-based merging \citep{sarasua2025diploma}. Given the scarcity of raw Wolof text, we rely primarily on Wolof-centric instruction tuning using translated public datasets, bilingual and native Wolof data, complemented by preference alignment and model merging.

Regarding evaluation, public Wolof evaluation resources remain sparse but are increasing: MasakhaNER 2.0 provides curated named-entity recognition data for Wolof \citep{adelani2022masakhaner2}, while AfrIFact covers evidence-grounded verification \citep{azime2026afrifact}. Broader multilingual benchmarks such as IrokoBench and Belebele further extend task coverage, although translated or English-pivoted data may not fully reflect in-language quality \citep{adelani2025irokobench,bandarkar2024belebele}. For instance, \citet{smart2026multiwikiqa} presents a question answering dataset with samples generated in Wolof by an LM that we benchmarked as non-Wolof-capable. We therefore complement automatic evaluation with manually checked in-house test sets, and blind human assessment by Wolof speakers. For model baselines, Aya-101 is multilingual but does not list Wolof \citep{ustun2024aya}, whereas Tiny Aya explicitly includes it \citep{salamanca2026tinyaya}. Oolel-v0.1 is a Wolof-dedicated 7B Qwen-2.5 model advertised for RAG, translation, generation, and math~\cite{soynade2025oolel}, but lacks public training data or a detailed evaluation, motivating our controlled comparison.

\section{Adaptation Method}
\label{sec:method}

The target adapted models are intended for public non-commercial release, with training data released under open licenses where possible. The main objective is strong Wolof understanding and generation while preserving French and English, with multi-turn conversation for customer-service use. We also track non-linguistic base model skills, especially instruction following, knowledge or safety for data processing or user interaction, while treating function calling, long contexts, and code as secondary checks for future agentic use. 

Our adaptation process first performs instruction tuning through Supervised Fine-Tuning (SFT) to acquire Wolof capabilities while seeking to preserve general behaviors through a mixed training corpus. We then evaluate preference alignment and model merging as complementary ways to improve this balance without adding new Wolof supervision. The contribution of these extra stages depends on model capacity: with larger positive effect as the model is large. These are usual post-training techniques, but implementing successful recipes in always a practical challenge.
Finally, let one note that all adaptations rely on full fine-tuning, rather than parameter-efficient approaches such as LoRA since the latter can be effective for domain adaptation but are less suited to adding a language.

\subsection{Instruction Tuning}
\label{sec:method-data}
Because native digital Wolof is scarce, our instruction tuning combines native bilingual resources with machine-translated public instructions to obtain both scale and task diversity. After comparing translation systems, Google Translate was selected for Wolof instruction generation.
The quality of the translations have been assessed by native speakers. First, random instructed were selected and checked. Then, more rigorously, we conducted a human annotation of Google Translate outputs with two native speakers, who evaluated 50 sentence translations w.r.t. the fidelity of the translated sentence's meaning (from 1 to 5) and the linguistic quality of the Wolof (from 1 to 5). The scores obtained---correct meaning: 4.59/5 and Wolof quality: 4.78/5---indicate that Google Translate produces translations that are both semantically faithful and linguistically correct. Details are given App.~\ref{sec:appendix-translation}.
We therefore obviously do not treat translated data as a substitute for fully native corpora, but consider it as a scalable imperfect-but-reliable supervision and further counterbalance it by native/bilingual sources and Wolof-specific evaluation.

Data mixture was guided by iterative studies of skills: Wolof understanding and generation, conversational coherence, multi-turn interaction, and function calling.
The resulting corpus is organized into four blocks, detailed in Table~\ref{tab:training-data}. The bilingual French-Wolof block uses Alwaly translation pairs to anchor the model in authentic Wolof text \citep{alwaly2026woloffrench}. The translated general block mixes instruction following, dialogue summarization, QA, and masked-turn dialogue completion from Alpaca \citep{taori2023alpaca}, Dolly-15k \citep{databricks2023dolly15k}, SAMSum \citep{gliwa2019samsum}, Helpful Instructions \citep{huggingfaceh42023helpful}, Hermes-217k \citep{hermes217k}, Natural Instructions \citep{wang2022supernaturalinstructions}, and Claire Dialogue \citep{hunter2023claire}. Multi-turn Wolof dialogue is injected through the Claire one-to-one split \citep{hunter2023claire}. Preserving of structured tool calls is aimed via English function calling data~\citep{younissk2024toolcallingmix,zhang2024xlam,liu2024apigen}.
Let one note that the massively multilingual dataset Aya (with a few thousands Wolof instructions) was voluntarily excluded from our training and reserved for evaluation to avoid contaminating one of the few public Wolof benchmarks.
The mixture contains 446K samples and 109.4M tokens. All details of our training corpus, along with the associated validations sets used for monitoring and development are given in~App.~\ref{sec:appendix-data}.

For completeness, we also investigated the use of a Continued Pre-Training (CPT) stage prior to SFT. However, only a raw corpus of 13M Wolof tokens could be collected, which was insufficient. The corresponding experiments are reported in App.~\ref{app:ongoing-cpt}. Tokenization adaptation has also been experimented using the CPT data. Interesting results are reported in App.~\ref{app:ongoing-tokenization}, but this is neither included in main recipe of the paper.

\begin{table}[t]
  \centering
  \scriptsize
  \setlength{\tabcolsep}{3pt}
  \begin{tabular}{@{}clcccc@{}}
    \hline
     & \textbf{Dataset source} & \textbf{Samples} & \textbf{Tokens} & \textbf{Lang.} & \textbf{N. turns} \\
    \hline
    \multirow{3}{*}{\rotatebox[origin=c]{90}{\textbf{Biling.  }}}
      & Alwaly FR$\rightarrow$WO & 91K & 5.0M & FR/WO & Single \\
      & Alwaly WO$\rightarrow$FR & 91K & 5.0M & WO/FR & Single \\
      & \textit{Total} & \textit{182K} & \textit{10.0M} &  & \\
    \hline
    \multirow{8}{*}{\rotatebox[origin=c]{90}{\textbf{Translated general}}}
      & Alpaca & 52K & 10.4M & WO & Single \\
      & Dolly-15k & 15K & 4.4M & WO & Single \\
      & SAMSum & 15K & 4.0M & WO & Single \\
      & Helpful instructions & 65K & 15.2M & WO & Single \\
      & Hermes-217k (33k subset) & 33K & 17M & WO & Single \\
      & WiseNUT & 20K & 5.3M & WO & Single \\
      & Claire multilogue & 15K & 22.5M & WO & Single \\
      & \textit{Total} & \textit{215K} & \textit{78.8M} &  & \\
    \hline
    \textbf{Conv.}
      & Claire one-to-one & 7K & 5.7M & WO & Multiple \\
    \hline
    \multirow{4}{*}{\rotatebox[origin=c]{90}{\textbf{Func. call.}}}
      & tool-calling-mix & 20K & 13.5M & EN & Multiple \\
      & xlamk60k & 20K & 13.5M & EN & Multiple \\
      & generated & 4K & 1.4M & EN & Multiple \\
      & \textit{Total} & \textit{44K} & \textit{14.9M} & & \\
    \hline
     & \textbf{TOTAL} & \textbf{446K} & \textbf{109.4M} & &  \\
    \hline
  \end{tabular}
  \caption{Training data. WO, FR, and EN stand for Wolof, French, and English, resp.. Turns are user-assistant pairs of messages (+ optional system message). }
  \label{tab:training-data}
  \vspace{-5mm}
\end{table}

\subsection{Preference Alignment and Model Merging}
Preference alignment and model merging were tested to balance after the SFT step.

Regarding preference alignment, we applied \emph{Direct Preference Optimization (DPO)}~\cite{rafailov2023direct} using the DOLCI preference dataset from OLMo~3~\citep{olmo2025olmo3}. The corpus contains 233,804 preference pairs in its training set, covering instruction-following, general knowledge, coding, mathematics, and safety. $98,\%$ of the samples are in English, and none in Wolof: the purpose here is to reinforce general assistant behavior after language adaptation while retaining the Wolof capabilities acquired through SFT. Appendix~\ref{app:alignment} provides the data composition and training configuration.

Regarding model merging, the idea is to avoid the adapted model to move too far from the original one to decrease the risk of catastrophic forgetting. In practice, we also use the simple \emph{linear interpolation} technique, also often referred to as \emph{model soup}~\cite{wortsman2022model}. In the experiments, model merging is experimented either after SFT or after DPO. In the first case, a sweep over interpolation coefficients selected weights of 0.4 for the base model and 0.6 for the adapted model. In the second case, the merging mixes the base, post-SFT and post-DPO model. The best coefficients selected are 1/3.
More advanced recent techniques~\cite{yang2024adamerging,zhou2024metagpt,touayouch2026divmerge} could be investigated to optimize these coefficients but this is left for future work since this is not the central scope of the paper.

\subsection{Training Setup}

We target open-weight small models to support on-premise inference on reasonable infrastructure, an important constraint when access to high-capability cloud services can be legally restricted in Senegal.
After testing several families (Llama, Gemma, Phi, Mistral, GPT-OSS), Qwen proved the best tradeoff between performance and stability. Development then used Qwen-2.5 7B models, while final experiments are applied on 0.6-14B Qwen3 models.

SFT and DPO trainings were conducted on internal single nodes with 2 NVIDIA H100 80GB or 1~H200 140GB GPUs.
SFT models were trained for 3 epochs, peak learning rate of $2 \times 10^{-5}$, and an effective batch size of 1024. The maximum sequence length was 2048 tokens.
DPO models were trained for 2 epochs with an effective batch size of 512, a maximum sequence length of 1024 tokens, with regularization coefficient $\beta=0.2$ and a peak learning rate of $5 \times 10^{-6}$.
Final checkpoints for SFT and DPO were selected by on loss on validation sets.
All hyperparameters for both stages are provided in App.~\ref{app:training-setup} and~\ref{app:alignment}.

\begin{table*}[t!]
\scriptsize
\centering
\begin{fitwidthtable}[\textwidth]
\begin{tabular}{@{}p{0.27\linewidth}@{~~~~}p{0.18\linewidth}@{~}p{0.14\linewidth}@{~~}|@{~~}l@{}l@{~~}|@{~~}l@{~}l@{~}l@{~}l@{}l@{~}l@{}l@{~}l@{~}l@{}l@{~}l@{\,}|@{~}r@{~}r@{}}

\textbf{Task} & \textbf{Input} & \textbf{Output}
& \rotatebox[origin=l]{90}{\textbf{Wolof}} & \rotatebox[origin=l]{90}{\textbf{In-house}}
& \rotatebox[origin=l]{90}{\textbf{Lang. underst.}} & \rotatebox[origin=l]{90}{\textbf{Lang. gen.}} & \rotatebox[origin=l]{90}{\textbf{Knowledge}} & \rotatebox[origin=l]{90}{\textbf{Chat}} & \rotatebox[origin=l]{90}{\textbf{Reason./Math}} & \rotatebox[origin=l]{90}{\textbf{Code}} & \rotatebox[origin=l]{90}{\textbf{Formating}} & \rotatebox[origin=l]{90}{\textbf{Instr. following}} & \rotatebox[origin=l]{90}{\textbf{Func. calling}} & \rotatebox[origin=l]{90}{\textbf{Long context}} & \rotatebox[origin=l]{90}{\textbf{Safety/ethics}}
& \rotatebox[origin=l]{90}{\textbf{Samples}} & \textbf{Metric} \\
\midrule

\multicolumn{18}{l}{\textbf{Knowledge \& QA \quad (6 tasks)}} \\
\texttt{mmlu\_mcqa\_EN} & Question (EN) & Choice & \xmark & \xmark & \LangUnd & & \Knowledge & & & & \StructFormat & & & & & 1,140 & Accuracy \\
\texttt{mmlu\_mcqa\_FR} & Question (FR) & Choice & \xmark & \xmark & \LangUnd & & \Knowledge & & & & \StructFormat & & & & & 1,140 & Accuracy \\
\texttt{irokobench\_mmlu} & Text (WO) & Choice & \cmark & \xmark & \LangUnd & & \Knowledge & & & & \StructFormat & & & & & 500 & Accuracy \\
\texttt{aya\_WO\_culture\_qa} & Question (WO) & Answer (WO) & \cmark & \xmark & \LangUnd & \LangGen & \Knowledge & & & & & & & & & 865 & LLM score \\
\texttt{corporate\_faq\_french\_mrqa} & Question + document (FR) & Answer (FR) & \xmark & \cmark & \LangUnd & \LangGen & \Knowledge & & & & & & & & & 122 & LLM score \\
\texttt{corporate\_faq\_wolof\_mrqa} & Question + document (WO) & Answer (WO) & \cmark & \cmark & \LangUnd & \LangGen & \Knowledge & & & & & & & & & 122 & METEOR \\
\midrule

\multicolumn{18}{l}{\textbf{Translation \& Generation \quad (16 tasks)}} \\
\texttt{aya\_WO\_translation\_from\_WO} & Sentence (WO) & Transl. (FR/EN) & \cmark & \xmark & \LangUnd & \LangGen & & & & & & & & & & 16 & METEOR \\
\texttt{aya\_WO\_translation\_to\_WO} & Sentence (FR/EN) & Transl. (WO) & \cmark & \xmark & \LangUnd & \LangGen & & & & & & & & & & 31 & METEOR \\
\texttt{aya\_WO\_orthographic\_correction} & Text (WO) & Corrected text (WO) & \cmark & \xmark & \LangUnd & \LangGen & & & & & & & & & & 244 & Edit distance \\
\texttt{flores\_*\_to\_*} (6 tasks) & Sentence (EN/FR/WO) & Transl. (EN/FR/WO) & \cmark (4/6) & \xmark & \LangUnd & \LangGen & & & & & & & & & & 1,012 & METEOR \\
\texttt{wolbanking77\_text\_*\_to\_*} (6 tasks) & Sentence (EN/FR/WO) & Transl. (EN/FR/WO) & \cmark (4/6) & \xmark & \LangUnd & \LangGen & & & & & & & & & & 500 & METEOR \\
\texttt{wolbanking77\_audio\_WO\_ASR\_correction} & ASR transcript (WO) & Corrected text (WO) & \cmark & \xmark & \LangUnd & \LangGen & & & & & & & & & & 641 & WER change \\
\midrule

\multicolumn{18}{l}{\textbf{Classification \quad (9 tasks)}} \\
\texttt{irokobench\_xnli} & Text pair (WO) & Class & \cmark & \xmark & \LangUnd & & & & & & \StructFormat & & & & & 600 & Accuracy \\
\texttt{wolbanking77\_text\_*\_intent} (3 tasks)& Sentence (EN/FR/WO) & Class & \cmark (1/3) & \xmark & \LangUnd & & & & & & \StructFormat & & & & & 500 & Macro F1 \\
\texttt{wolbanking77\_audio\_WO\_manual\_intent} & Manual transcript (WO) & Class & \cmark & \xmark & \LangUnd & & & & & & \StructFormat & & & & & 641 & Macro F1 \\
\texttt{wolbanking77\_audio\_WO\_ASR\_intent} & ASR transcript (WO) & Class & \cmark & \xmark & \LangUnd & & & & & & \StructFormat & & & & & 641 & Macro F1 \\
\texttt{corporate\_dialog\_act\_labelling} & Conv. (WO) & Labels & \cmark & \cmark & \LangUnd & & & & & & \StructFormat & & & & & 150 & Macro F1 \\
\texttt{corporate\_resolution\_classification} & Conv. (WO) & Class & \cmark & \cmark & \LangUnd & & & & & & \StructFormat & & & & & 150 & Macro F1 \\
\texttt{corporate\_sentiment\_classification} & Conv. (WO) & Labels & \cmark & \cmark & \LangUnd & & & & & & \StructFormat & & & & & 150 & Macro F1 \\
\midrule

\multicolumn{18}{l}{\textbf{Information Extraction \quad (7 tasks)}} \\
\texttt{aya\_WO\_pos\_tagging} & Text (WO) & POS tags & \cmark & \xmark & \LangUnd & & & & & & \StructFormat & & & & & 159 & Macro F1 \\
\texttt{masakha\_ner} & Text (WO) & BIO labels & \cmark & \xmark & \LangUnd & & & & & & \StructFormat & & & & & 536 & Exact F1 \\
\texttt{afrifact\_claim\_classification} & Text (WO) + claim (WO) & Class & \cmark & \xmark & \LangUnd & & & & & & \StructFormat & & & & & 1,658 & Macro F1 \\
\texttt{afrifact\_evidence\_extraction} & Text (WO) + claim (WO) & Span & \cmark & \xmark & \LangUnd & & & & & & & & & & & 1,658 & Partial F1 \\
\texttt{corporate\_gdpr\_labelling} & Conv. (WO) & Entity list & \cmark & \cmark & \LangUnd & & & & & & \StructFormat & & & & & 150 & Exact F1 \\
\texttt{corporate\_ner} & Conv. (WO) & Entity list & \cmark & \cmark & \LangUnd & & & & & & \StructFormat & & & & & 150 & Exact F1 \\
\texttt{corporate\_full\_analysis} & Conv. (WO) & Composite (JSON) & \cmark & \cmark & \LangUnd & \LangGen & & & & & \StructFormat & & & & & 150 & Composite \\
\midrule

\multicolumn{18}{l}{\textbf{Conversation \quad (8 tasks)}} \\
\texttt{claire\_WO\_conversation\_coherence} & Conv. (WO) & Response (WO) & \cmark & \xmark & \LangUnd & \LangGen & & \Chat & & & & & & & & 292 & LLM score \\
\texttt{dialogsum\_conversation\_summarization} & Conv. (EN) & Summary (EN) & \xmark & \xmark & \LangUnd & \LangGen & & & & & & & & & & 200 & ROUGE-1 \\
\texttt{personachat\_conversation\_coherence} & Conv. + persona (EN) & Response (EN) & \xmark & \xmark & \LangUnd & \LangGen & & \Chat & & & & & & & & 500 & LLM score \\
\texttt{webnlg\_qa\_conversational\_mrqa} & Question + conv. (EN) & Response (EN) & \xmark & \xmark & \LangUnd & & & \Chat & & & & & & & & 232 & LLM score \\
\texttt{corporate\_faq\_french\_conv\_mrqa} & Question + conv. (FR) & Response (FR) & \xmark & \cmark & \LangUnd & \LangGen & & \Chat & & & & & & & & 122 & LLM score \\
\texttt{corporate\_faq\_wolof\_conv\_mrqa} & Question + conv. (WO) & Response (WO) & \cmark & \cmark & \LangUnd & \LangGen & & \Chat & & & & & & & & 122 & METEOR \\
\texttt{corporate\_conversation\_coherence} & Conv. (WO) & Response (WO) & \cmark & \cmark & \LangUnd & \LangGen & & \Chat & & & & & & & & 287 & LLM score \\
\texttt{corporate\_summarization} & Conv. (WO) & Summary (WO) & \cmark & \cmark & \LangUnd & \LangGen & & & & & & & & & & 150 & LLM score \\
\midrule

\multicolumn{18}{l}{\textbf{Reasoning, Math \& Code \quad (6 tasks)}} \\
\texttt{arc\_challenge\_reasoning} & Science question (EN) & Choice & \xmark & \xmark & \LangUnd & & \Knowledge & & \ReasonMath & & & & & & & 2,600 & Accuracy \\
\texttt{gpqa\_diamond\_EN\_mcqa} & Science question (EN) & Choice & \xmark & \xmark & \LangUnd & & \Knowledge & & \ReasonMath & & \StructFormat & & & & & 198 & Accuracy \\
\texttt{gpqa\_diamond\_FR\_mcqa} & Science question (FR) & Choice & \xmark & \xmark & \LangUnd & & \Knowledge & & \ReasonMath & & \StructFormat & & & & & 197 & Accuracy \\
\texttt{gsm8k\_reasoning} & Math problem (EN) & Integer & \xmark & \xmark & \LangUnd & & & & \ReasonMath & & & & & & & 1,319 & Accuracy \\
\texttt{irokobench\_mgsm} & Math problem (WO) & Integer & \cmark & \xmark & \LangUnd & & & & \ReasonMath & & & & & & & 250 & Exact match \\
\texttt{humaneval\_code\_generation} & Code problem (EN) & Code (Python) & \xmark & \xmark & \LangUnd & \LangGen & & & & \CodeSkill & \StructFormat & & & & & 164 & pass@1 \\
\midrule

\multicolumn{18}{l}{\textbf{Safety \& Ethics \quad (4 tasks)}} \\
\texttt{bbq\_demographic\_bias\_mcqa} & Question (EN) & Label & \xmark & \xmark & \LangUnd & & & & & & \StructFormat & & & & \SafetyEthics & 500 & Accuracy \\
\texttt{halueval\_qa\_hallucination\_detection} & Question (EN) & Label & \xmark & \xmark & \LangUnd & & & & & & \StructFormat & & & & \SafetyEthics & 500 & Accuracy \\
\texttt{toxigen\_implicit\_hate\_detection} & Text (EN) & Label & \xmark & \xmark & \LangUnd & & & & & & \StructFormat & & & & \SafetyEthics & 500 & Accuracy \\
\texttt{wildjailbreak\_harmful\_request\_refusal} & Text (EN) & Refusal indicator & \xmark & \xmark & \LangUnd & & & & & & & & & & \SafetyEthics & 500 & Refusal rate \\
\midrule

\multicolumn{18}{l}{\textbf{Instruction Following \& Function Calling \quad (5 tasks)}} \\
\texttt{multi\_if\_instruction\_following} & Instruction (EN/FR) & Text & \xmark & \xmark & \LangUnd & & & & & & & \InstrFollow & & & & 808 & Accuracy \\
\texttt{ifbench\_instruction\_following} & Instruction (EN) & Text & \xmark & \xmark & \LangUnd & & & & & & & \InstrFollow & & & & 300 & Accuracy \\
\texttt{mifeval\_EN\_instruction\_following} & Instruction (EN) & Text & \xmark & \xmark & \LangUnd & & & & & & & \InstrFollow & & & & 541 & Accuracy \\
\texttt{mifeval\_FR\_instruction\_following} & Instruction (FR) & Text & \xmark & \xmark & \LangUnd & & & & & & & \InstrFollow & & & & 235 & Accuracy \\
\texttt{bfcl\_function\_calling} & Instruction (EN) & Tool call & \xmark & \xmark & \LangUnd & & & & & & \StructFormat & & \FuncCall & & & 3,641 & Micro accuracy \\
\midrule

\multicolumn{18}{l}{\textbf{Long-Context QA \quad (2 tasks)}} \\
\texttt{ruler\_8k\_context\_mrqa} & Question + documents (EN) & Answer & \xmark & \xmark & \LangUnd & & & & & & & & & \LongContext & & 60 & Exact match \\
\texttt{corporate\_faq\_wolof\_long\_contx\_mrqa} & Question + documents (WO) & Answer & \cmark & \cmark & \LangUnd & & & & & & & & & \LongContext & & 99 & METEOR \\
\bottomrule
\end{tabular}
\end{fitwidthtable}
\caption{Evaluation suite grouped by task family with contribution to target skills. Detailed task descriptions and dataset references are provided in App.~\ref{app:tasks}.}
\label{tab:all_tasks}
\vspace{-4mm}
\end{table*}

\section{Benchmarking Tasks and Skills}
\label{sec:benchmark}

\begin{table*}[t!]
\centering
\scriptsize
\begin{fitwidthtable}
\begin{tabular}{@{}l@{}l@{~}|@{~~}l@{~~}l@{~~}l@{~~}l@{~~}l@{~~}l@{~~}l@{~~}l@{~~}l@{~~}l@{~~}l@{~~}l@{~~}l@{~~}l@{~~}l@{}}
\toprule
 & Avg. sc.$^\uparrow{}$ &\multicolumn{2}{@{}c@{}}{English} &\multicolumn{2}{@{}c@{}}{French} &\multicolumn{2}{@{}c@{}}{Wolof} &\multicolumn{1}{@{}c@{}}{Know.} & \multicolumn{1}{@{}c@{}}{Conv.} & \multicolumn{1}{@{}c@{}}{Reas.} & \multicolumn{1}{@{}c@{}}{Code} & \multicolumn{1}{@{}c@{}}{Format.} & \multicolumn{1}{@{}c@{}}{Instr.} & \multicolumn{1}{@{}c@{}}{Func.} & \multicolumn{1}{@{}c@{}}{Lg ctx.} & \multicolumn{1}{@{}c@{}}{Safet.} \\
\cline{3-8}
& {\tiny (\& rank$^\downarrow{}$)} & \multicolumn{1}{@{}c@{}}{Und.{\tiny (22)}} & \multicolumn{1}{@{}c@{}}{Gen.{\tiny (8)}} & \multicolumn{1}{@{}c@{}}{Und.{\tiny (16)}} & \multicolumn{1}{@{}c@{}}{Gen.{\tiny (7)}} & \multicolumn{1}{@{}c@{}}{Und.{\tiny (30)}} & \multicolumn{1}{@{}c@{}}{Gen.{\tiny (14)}} & \multicolumn{1}{@{}c@{}}{{\tiny (7)}} & \multicolumn{1}{@{}c@{}}{{\tiny (6)}} & \multicolumn{1}{@{}c@{}}{{\tiny (6)}} & \multicolumn{1}{@{}c@{}}{{\tiny (1)}} & \multicolumn{1}{@{}c@{}}{{\tiny (22)}} & \multicolumn{1}{@{}c@{}}{{\tiny (4)}} & \multicolumn{1}{@{}c@{}}{{\tiny (1)}} & \multicolumn{1}{@{}c@{}}{{\tiny (2)}} & \multicolumn{1}{@{}c@{}}{{\tiny (4)}} \\
\midrule
\multicolumn{17}{@{}l}{\textbf{Baseline open-weight models}}\\
Qwen3-14B & \scoremid{\sr{.32}{10.7}} & \scorehigh{\sr{.52}{10}} & \scoremid{\sr{.46}{13}} & \scorehigh{\sr{.40}{10}} & \scoremid{\sr{.45}{12}} & \scorelow{\sr{.10}{15}} & \scorelow{\sr{.00}{16}} & \scoremid{\sr{.25}{12}} & \scoremid{\sr{.33}{13}} & \scoremid{\sr{.37}{11}} & \scorehigh{\sr{.74}{10}} & \scoremid{\sr{.30}{12}} & \scorehigh{\sr{.67}{6}} & \scoretop{\sr{.70}{2}} & \scorehigh{\sr{.45}{10}} & \scorehigh{\sr{.54}{9}} \\

Qwen3-8B & \scoremid{\sr{.29}{12.6}} & \scorehigh{\sr{.49}{11}} & \scoremid{\sr{.46}{13}} & \scoremid{\sr{.38}{12}} & \scoremid{\sr{.44}{13}} & \scorelow{\sr{.07}{16}} & \scorelow{\sr{-.03}{18}} & \scoremid{\sr{.20}{14}} & \scoremid{\sr{.31}{14}} & \scoremid{\sr{.30}{14}} & \scoremid{\sr{.54}{14}} & \scoremid{\sr{.26}{13}} & \scorehigh{\sr{.63}{9}} & \scoretop{\sr{.67}{6}} & \scorehigh{\sr{.45}{10}} & \scoremid{\sr{.45}{12}} \\

Qwen3-4B & \scoremid{\sr{.28}{13.5}} & \scoremid{\sr{.46}{13}} & \scoremid{\sr{.45}{16}} & \scoremid{\sr{.37}{13}} & \scoremid{\sr{.44}{13}} & \scorelow{\sr{.07}{16}} & \scorelow{\sr{-.02}{17}} & \scoremid{\sr{.24}{13}} & \scorelow{\sr{.26}{17}} & \scoremid{\sr{.33}{13}} & \scorehigh{\sr{.62}{13}} & \scoremid{\sr{.25}{14}} & \scorehigh{\sr{.59}{10}} & \scoretop{\sr{.68}{5}} & \scorehigh{\sr{.42}{14}} & \scoremid{\sr{.38}{15}} \\

Qwen3-1.7B & \scorelow{\sr{.23}{16.2}} & \scoremid{\sr{.40}{14}} & \scoremid{\sr{.41}{17}} & \scoremid{\sr{.32}{16}} & \scoremid{\sr{.39}{17}} & \underline{\scorelow{\sr{.04}{19}}} & \underline{\scorelow{\sr{-.04}{19}}} & \scorelow{\sr{.10}{18}} & \underline{\scorelow{\sr{.23}{19}}} & \scorelow{\sr{.26}{17}} & \scoremid{\sr{.53}{16}} & \scoremid{\sr{.20}{17}} & \scorehigh{\sr{.50}{13}} & \scorehigh{\sr{.61}{9}} & \scorehigh{\sr{.37}{16}} & \scoremid{\sr{.32}{16}} \\

Qwen3-0.6B & \scorelow{\sr{.20}{17.3}} & \scoremid{\sr{.33}{18}} & \scorelow{\sr{.37}{18}} & \scorelow{\sr{.27}{18}} & \underline{\scorelow{\sr{.32}{19}}} & \scorelow{\sr{.06}{18}} & \scoremid{\sr{.03}{15}} & \underline{\scorelow{\sr{.07}{19}}} & \scorelow{\sr{.25}{18}} & \scorelow{\sr{.21}{18}} & \scoremid{\sr{.37}{18}} & \scorelow{\sr{.15}{18}} & \scoremid{\sr{.38}{16}} & \scorehigh{\sr{.53}{12}} & \scoremid{\sr{.18}{18}} & \scorelow{\sr{.30}{17}} \\

\midrule
\multicolumn{17}{@{}l}{\textbf{Commercial API models}}\\
Claude-sonnet-4-5 & \scorehigh{\sr{.47}{4.0}} & \scorehigh{\sr{.60}{3}} & \scorehigh{\sr{.55}{5}} & \scorehigh{\sr{.50}{4}} & \scorehigh{\sr{.55}{4}} & \scorehigh{\sr{.31}{6}} & \scorehigh{\sr{.27}{6}} & \scorehigh{\sr{.54}{2}} & \scoretop{\sr{.57}{2}} & \scorehigh{\sr{.65}{2}} & \scoretop{\sr{.98}{2}} & \scorehigh{\sr{.45}{4}} & \scorehigh{\sr{.71}{5}} & \textbf{\scoretop{\sr{.72}{1}}} & \scoretop{\sr{.47}{7}} & \scorehigh{\sr{.57}{7}} \\

Claude-haiku-4-5 & \scorehigh{\sr{.41}{6.8}} & \scorehigh{\sr{.56}{6}} & \scorehigh{\sr{.50}{9}} & \scorehigh{\sr{.46}{6}} & \scorehigh{\sr{.50}{8}} & \scoremid{\sr{.23}{9}} & \scorehigh{\sr{.18}{10}} & \scorehigh{\sr{.42}{6}} & \scorehigh{\sr{.51}{4}} & \scorehigh{\sr{.53}{6}} & \scoretop{\sr{.93}{6}} & \scorehigh{\sr{.40}{6}} & \scorehigh{\sr{.67}{6}} & \scoretop{\sr{.70}{2}} & \scorehigh{\sr{.45}{10}} & \scorehigh{\sr{.55}{8}} \\

Gemini-2.5-pro & \scoretop{\sr{.54}{2.9}} & \scorehigh{\sr{.59}{4}} & \scoretop{\sr{.61}{2}} & \scoretop{\sr{.52}{2}} & \scoretop{\sr{.59}{3}} & \textbf{\scoretop{\sr{.45}{1}}} & \textbf{\scoretop{\sr{.35}{1}}} & \scorehigh{\sr{.51}{3}} & \textbf{\scoretop{\sr{.59}{1}}} & \scorehigh{\sr{.65}{2}} & \scoretop{\sr{.94}{5}} & \scoretop{\sr{.54}{2}} & \scoretop{\sr{.75}{3}} & \scoretop{\sr{.65}{7}} & \scoretop{\sr{.49}{4}} & \scorehigh{\sr{.68}{4}} \\

Gemini-2.5-flash & \scoretop{\sr{.53}{3.1}} & \scorehigh{\sr{.59}{4}} & \scoretop{\sr{.60}{3}} & \scoretop{\sr{.52}{2}} & \scoretop{\sr{.60}{2}} & \scoretop{\sr{.44}{2}} & \textbf{\scoretop{\sr{.35}{1}}} & \scorehigh{\sr{.51}{3}} & \scoretop{\sr{.57}{2}} & \scorehigh{\sr{.63}{5}} & \scoretop{\sr{.96}{4}} & \scoretop{\sr{.53}{3}} & \scorehigh{\sr{.73}{4}} & \scoretop{\sr{.69}{4}} & \scoretop{\sr{.50}{2}} & \scorehigh{\sr{.64}{5}} \\

GPT-5 & \scoretop{\sr{.56}{2.6}} & \textbf{\scoretop{\sr{.70}{1}}} & \textbf{\scoretop{\sr{.62}{1}}} & \textbf{\scoretop{\sr{.56}{1}}} & \textbf{\scoretop{\sr{.61}{1}}} & \scoretop{\sr{.41}{3}} & \scorehigh{\sr{.27}{6}} & \textbf{\scoretop{\sr{.70}{1}}} & \scoremid{\sr{.40}{9}} & \textbf{\scoretop{\sr{.76}{1}}} & \textbf{\scoretop{\sr{.99}{1}}} & \textbf{\scoretop{\sr{.58}{1}}} & \textbf{\scoretop{\sr{.85}{1}}} & \scorehigh{\sr{.59}{10}} & \textbf{\scoretop{\sr{.52}{1}}} & \textbf{\scoretop{\sr{.86}{1}}} \\

GPT-5-nano & \scorehigh{\sr{.44}{5.6}} & \scoretop{\sr{.65}{2}} & \scorehigh{\sr{.51}{7}} & \scorehigh{\sr{.47}{5}} & \scorehigh{\sr{.52}{7}} & \scoremid{\sr{.22}{10}} & \scoremid{\sr{.14}{12}} & \scorehigh{\sr{.48}{5}} & \scorehigh{\sr{.42}{7}} & \scorehigh{\sr{.65}{2}} & \scoretop{\sr{.98}{2}} & \scorehigh{\sr{.40}{6}} & \scoretop{\sr{.80}{2}} & \scorehigh{\sr{.58}{11}} & \scoretop{\sr{.49}{4}} & \scoretop{\sr{.82}{2}} \\

Mistral-medium-3 & \scoremid{\sr{.37}{8.8}} & \scorehigh{\sr{.54}{8}} & \scorehigh{\sr{.49}{10}} & \scorehigh{\sr{.45}{7}} & \scorehigh{\sr{.50}{8}} & \scoremid{\sr{.17}{12}} & \scoremid{\sr{.06}{14}} & \scoremid{\sr{.36}{8}} & \scoremid{\sr{.37}{11}} & \scorehigh{\sr{.52}{7}} & \scoretop{\sr{.88}{7}} & \scorehigh{\sr{.37}{9}} & \scorehigh{\sr{.64}{8}} & \scoretop{\sr{.62}{8}} & \scoretop{\sr{.49}{4}} & \scoremid{\sr{.48}{11}} \\

\midrule
\multicolumn{17}{@{}l}{\textbf{Wolof-adapted open-weight models (ours)}}\\
Wolof-Qwen3-14B & \scorehigh{\sr{.47}{6.3}} & \scorehigh{\sr{.56}{6}} & \scorehigh{\sr{.56}{4}} & \scorehigh{\sr{.44}{8}} & \scorehigh{\sr{.55}{4}} & \scorehigh{\sr{.36}{4}} & \scoretop{\sr{.31}{3}} & \scorehigh{\sr{.41}{7}} & \scorehigh{\sr{.44}{6}} & \scoremid{\sr{.46}{8}} & \scorehigh{\sr{.79}{8}} & \scorehigh{\sr{.43}{5}} & \scorehigh{\sr{.58}{11}} & \scoremid{\sr{.35}{15}} & \scoretop{\sr{.50}{2}} & \scorehigh{\sr{.73}{3}} \\

Wolof-Qwen3-8B & \scorehigh{\sr{.44}{7.8}} & \scorehigh{\sr{.54}{8}} & \scorehigh{\sr{.53}{6}} & \scorehigh{\sr{.42}{9}} & \scorehigh{\sr{.53}{6}} & \scorehigh{\sr{.33}{5}} & \scorehigh{\sr{.28}{5}} & \scoremid{\sr{.36}{8}} & \scorehigh{\sr{.42}{7}} & \scoremid{\sr{.44}{9}} & \scorehigh{\sr{.69}{11}} & \scorehigh{\sr{.40}{6}} & \scorehigh{\sr{.57}{12}} & \scorehigh{\sr{.46}{13}} & \scoretop{\sr{.47}{7}} & \scorehigh{\sr{.64}{5}} \\

Wolof-Qwen3-4B & \scorehigh{\sr{.40}{10.7}} & \scoremid{\sr{.47}{12}} & \scoremid{\sr{.48}{11}} & \scoremid{\sr{.39}{11}} & \scorehigh{\sr{.48}{10}} & \scorehigh{\sr{.30}{8}} & \scorehigh{\sr{.27}{6}} & \scoremid{\sr{.30}{11}} & \scoremid{\sr{.39}{10}} & \scoremid{\sr{.38}{10}} & \scorehigh{\sr{.64}{12}} & \scorehigh{\sr{.36}{10}} & \scorehigh{\sr{.49}{14}} & \scorehigh{\sr{.37}{14}} & \scorehigh{\sr{.45}{10}} & \scoremid{\sr{.45}{12}} \\

Wolof-Qwen3-1.7B & \scoremid{\sr{.31}{14.6}} & \scoremid{\sr{.37}{17}} & \scoremid{\sr{.46}{13}} & \scoremid{\sr{.31}{17}} & \scoremid{\sr{.43}{16}} & \scoremid{\sr{.22}{10}} & \scorehigh{\sr{.22}{9}} & \scoremid{\sr{.20}{14}} & \scoremid{\sr{.36}{12}} & \scoremid{\sr{.29}{15}} & \scoremid{\sr{.48}{17}} & \scoremid{\sr{.24}{15}} & \scoremid{\sr{.25}{17}} & \scoremid{\sr{.24}{16}} & \scorehigh{\sr{.42}{14}} & \scorelow{\sr{.30}{17}} \\

Wolof-Qwen3-0.6B & \underline{\scorelow{\sr{.20}{17.4}}} & \underline{\scorelow{\sr{.26}{19}}} & \underline{\scorelow{\sr{.35}{19}}} & \underline{\scorelow{\sr{.24}{19}}} & \scorelow{\sr{.33}{18}} & \scoremid{\sr{.12}{14}} & \scorehigh{\sr{.17}{11}} & \scorelow{\sr{.14}{17}} & \scorelow{\sr{.27}{15}} & \underline{\scorelow{\sr{.20}{19}}} & \underline{\scorelow{\sr{.20}{19}}} & \underline{\scorelow{\sr{.12}{19}}} & \underline{\scorelow{\sr{.13}{19}}} & \scoremid{\sr{.23}{17}} & \scoremid{\sr{.26}{17}} & \underline{\scorelow{\sr{.22}{19}}} \\

\midrule
\multicolumn{17}{@{}l}{\textbf{Wolof-aware models from the literature}}\\
Oolel-v0.1 (7B) & \scoremid{\sr{.37}{10.7}} & \scoremid{\sr{.40}{14}} & \scorehigh{\sr{.51}{7}} & \scoremid{\sr{.35}{14}} & \scoremid{\sr{.44}{13}} & \scorehigh{\sr{.31}{6}} & \scoretop{\sr{.31}{3}} & \scoremid{\sr{.31}{10}} & \scorehigh{\sr{.45}{5}} & \scoremid{\sr{.34}{12}} & \scorehigh{\sr{.77}{9}} & \scorehigh{\sr{.35}{11}} & \scorelow{\sr{.20}{18}} & \underline{\scorelow{\sr{.00}{18}}} & \scoretop{\sr{.47}{7}} & \scoremid{\sr{.40}{14}} \\

tiny-aya-earth (3B) & \scoremid{\sr{.29}{14.5}} & \scoremid{\sr{.38}{16}} & \scoremid{\sr{.47}{12}} & \scoremid{\sr{.33}{15}} & \scorehigh{\sr{.47}{11}} & \scoremid{\sr{.17}{12}} & \scoremid{\sr{.12}{13}} & \scoremid{\sr{.18}{16}} & \scorelow{\sr{.27}{15}} & \scorelow{\sr{.27}{16}} & \scoremid{\sr{.54}{14}} & \scoremid{\sr{.22}{16}} & \scorehigh{\sr{.49}{14}} & \underline{\scorelow{\sr{.00}{18}}} & \underline{\scorelow{\sr{.06}{19}}} & \scoremid{\sr{.50}{10}} \\

\bottomrule
\end{tabular}
\end{fitwidthtable}
\caption{Automatic results by language and skill: normalised scores (higher, better) and ranks (lower, better). \scorecolorlegend}
\label{tab:benchmark_results_abbrev_noprovider}
\end{table*}

We designed an evaluation suite of 63~tasks around the paper's main deployment objective: improving Wolof understanding and generation while preserving English and French capability. The suite combines public academic benchmarks with in-house customer-service tasks. In total, 35 tasks include Wolof and 13 use in-house data.
Table~\ref{tab:all_tasks} summarises each task (input, output, samples, metric), Wolof and in-house coverage, and contribution to skills. Details are provided in App.~\ref{app:tasks}. The in-house tasks use internal documents and customer conversations annotated by Wolof-speaking professionals, and cover FAQ retrieval, intent routing, privacy-sensitive entity extraction, summarization, and response generation.

Wolof understanding and generation are the primary adaptation target skills. English and French monitor multilingual behavior, especially for Senegal. Knowledge, chat, formatting, instruction following, and safety capture assistant requirements for interaction with users, while function calling and reasoning target agentic AI. Finally, code and long context serve as broader regression checks.

We use deterministic metrics for outputs that can be matched reliably and LLM-as-a-judge for free-form generation: GPT-4o-mini for English/French and Gemini-2.5-flash for Wolof. For the latter, on 450 Wolof responses, native speakers
accepted 95\% of the scores assigned by Gemini-2.5-flash (99\% for
Gemini-2.5-pro), as detailed in App.~\ref{app:llm-as-a-judge}. These LLM scores should still be treated as comparative signals rather than used as standalone evidence. This is why automatic evaluation of Sec.~\ref{sec:automatic_results} is complemented by human evaluation in Sec.~\ref{sec:human_results}.

Finally, all evaluations use multiple prompt templates (listed in App.~\ref{app:tasks}) sampled with fixed seed for reproducibility. Decoding uses temperature zero, and on-premise models run in 16-bit precision.

\section{Results}
\label{sec:results}

This section reports final automatic and human results. 
Whereas post-SFT gains vary by model size, this section only reports the best setup for each size (the final released model) due to space constraints. Stage-by-stage comparisons are provided in App.~\ref{app:stage-results}.

\subsection{Automatic evaluation}
\label{sec:automatic_results}

Table~\ref{tab:benchmark_results_abbrev_noprovider} reports the overall and skill-aggregated results for four model families: unadapted open-weight Qwen3 baseline models, our Wolof-adapted Qwen3 models, commercial API systems used as upper reference points, and Wolof-aware open-weight models from prior releases (Oolel and tiny-aya-earth).
Scores are normalised per task based on their [\textit{min}, \textit{max}] range, such that \textit{min} is the score of a random predictor (random classification or random text ; not reported here) and \textit{max} depends on the metric (e.g., 1 for METEOR, 5 for LLM-as-a-judge, etc.). All scores are the higher, the better. Normalized values below zero indicate performance below the lower (random) score on that skill.
Explicit \texttt{<think>} traces are removed from Qwen3 outputs before scoring.
A graphical summary of Table~\ref{tab:benchmark_results_abbrev_noprovider} is also provided in App.~\ref{app:score-params}, showing overall scores as a function of model size.

First, the clearest result is that adaptation produces a genuine acquisition of Wolof. Across sizes, the unadapted Qwen3 models remain close to the random baseline both in Wolof understanding and generation, whereas adaptation significantly raises scores for these skills. Our models fall within the range of the commercial APIs. As compared to prior Wolof-aware open models, Oolel demonstrates similar linguistic performance in Wolof but ranks lower in most other skills. Aya-tiny's results show that it is not dedicated to Wolof and thus performs poorly.

Then, at 4B-14B scales, these gains coexist with good preservation of the base models' broader capabilities like English, French, and most other skills like safety or reasoning, or even long context and coding which were yet not part of the originally targeted skills for our models.
The gains are nevertheless uneven at the smallest sizes. In particular, the 0.6B model improves on Wolof but degrades on several skills. 
Likewise, whereas the targeted applications are not agentic ones, the function calling skill declines. After manual investigations, this is mainly due to false positives, \ie the models predict tool calls too frequently, even when none is needed.

Deeper investigations lead to other interesting conclusions. First, task-by-task results (detailed in App.~\ref{app:skill-detailed-results}) show that improvements on Wolof capabilities span all types of tasks: classification, labelling, generation, etc. Second, the stage-by-stage analysis (see App.~\ref{app:stage-results}) identifies SFT as the main driver for adaptation, while the effect of DPO and merging is more variable and depending on the model size. A possible explanation is that the hyperparameters---which have been tuned on validation data only for the 8B size---should be set differently according to the size.

\subsection{Human Evaluation via Arena}
\label{sec:human_results}

Automatic metrics are complemented with an arena-style blind human evaluation on Wolof-facing tasks. Twelve Wolof-speaking annotators compared outputs from 5~models: Gemini-2.5-pro, Gemini-2.5-flash, our final Wolof-Qwen3-8B model, its unadapted Qwen3-8B base model, and Oolel-v0.1. The annotators were not informed of the actual model identities during evaluation. The evaluation used 100~samples from each of 9~tasks.\footnote{\texttt{claire\_\-WO\_\-conversation\_\-coherence}, \texttt{corporate\_\-faq\_\-WO\_\-mrqa}, \texttt{flores\_\-translation\_\-EN\_\-to\_\-WO}, \texttt{flores\_\-translation\_\-WO\_\-to\_\-FR}, \texttt{corporate\_\-summarization}, \texttt{wolbanking77\_\-WO\_\-intent\_\-classification}, \texttt{wolbanking77\_\-FR\_\-to\_\-WO\_\-translation}, \texttt{wolbanking77\_\-WO\_\-to\_\-EN\_\-translation}, and \texttt{wolbanking77\_\-audio\_\-transcription\_\-correction}} The four criteria were correctness, formatting, Wolof quality, and overall preference. Annotators could select either system, mark both outputs as good or bad, or skip the comparison. Pairwise outcomes were converted to Elo scores, splitting ties equally, as well as win probabilities for a better intuition (0.5 indicates equal chances to win or lose). For the Wolof quality, samples whose expected output was English or French or a classification label were excluded.

\begin{table}[t]
\centering
\scriptsize
\begin{tabular}{@{}l@{~~~~}r@{~~~~}r@{~~~~}r@{~~~~}r@{~~~~}}
\toprule
Model & Overall & Correctness & Formatting & Wolof quality \\
\midrule
Gemini-2.5-flash & \textbf{1112} \tiny(.67) & \textbf{1104} \tiny(.66) & 1072 \tiny(.61) & \textbf{1066} \tiny(.60) \\
Gemini-2.5-pro & 1093 \tiny(.64) & 1082 \tiny(.63) & \textbf{1090} \tiny(.64) & 1062 \tiny(.60) \\
Wolof-Qwen3-8B & 1051 \tiny(.57) & 1035 \tiny(.55) & 1030 \tiny(.54) & 1058 \tiny(.59) \\
Oolel-v0.1 & 1002 \tiny(.48) & 994 \tiny(.48) & 1030 \tiny(.54) & 1038 \tiny(.55) \\
Qwen3-8B & 742 \tiny(.14) & 785 \tiny(.18) & 777 \tiny(.17) & 777 \tiny(.16) \\
\bottomrule
\end{tabular}
\caption{Human evaluation Elo scores and win probability by criterion. The Wolof quality column excludes tasks whose expected output is English or French or a classification label.}
\label{tab:human_eval_elo}
\end{table}

Table~\ref{tab:human_eval_elo} provides the results.
Overall, Gemini models are slightly preferred but very close to our 8B model, while Oolel is slightly further behind and the non-adapted model is clearly worse, which confirms the results of automatic evaluations.
In terms of pure Wolof quality (\ie, independent of correctness), our model is clearly comparable to closed models.
Consequently, these results are a positive indication of our models' deployability.

\section{Conclusion and Further Improvements}
\label{sec:current-work}

In this paper, we introduced open-weight models adapted to Wolof, ranging from 0.6B to 14B parameters, along with detailed descriptions of the training pipeline, datasets, and evaluation tasks. Automatic and human evaluations show that our models have developed strong Wolof capabilities while retaining their multilingual and operational abilities, especially for customer service applications. The 8B and 14B models, in particular, narrow the gap with commercial APIs. These models, training datasets, and selected in-house test sets are now ready to be publicly released for free non-commercial use.

Our results also provide practical insights into the effects of model size, the role of SFT, and the variability of preference alignment and model merging, leaving room for further improvements.
In parallel, targeting agentic AI capabilities represents an important next step, requiring a particular focus on function calling. Work on Group Relative Policy Optimization (GRPO)~\citep{shao2024deepseekmath} with verifiable rewards has begun in pursuit of this new objective.

\section*{Limitations}

Our final experiments use the Qwen family. Altough preliminary experiments had been conducted to select the Qwen familiy, cross-family generalisation should be tested again now that full recipes have been built.

Then, as shown in the paper, Wolof adaptation can produce uneven changes across English, French, instruction-following, and tool-use tasks, particularly at the smallest model sizes. Multilingual or task-specific deployments therefore still require task-specific validation.

Third, the evaluation suite could still be augmented with industrial failure modes (adversarial inputs, privacy leakage, ASR noise, low-literacy users),  agentic scenarios, and larger user studies. Regarding safety, specific tests related to the socio-cultural environment in Senegal should complement the standard safety tasks used so far. We hope that the internal deployment of these models in our service will help discover such new quality indicators.

While LLM-as-a-judge is still a research question for high-resource languages, it should even more be cautiously interpreted for Wolof. In spite of our human assessments of LLM judges, this is an area of improvement.

Finally, computational constraints (single node trainings on 1 or 2 GPUs) limit ablation scope and multi-seed runs. Nonetheless, we think that the diversity of evaluation tasks counterbalances this aspect to provide a reliable view of our models' performance.

\section*{Ethics and Societal Impact}

LLMs were used in this paper to automate the generation of repetitive \LaTeX\ structures (tasks details, skill tables, etc.), as well as improve English quality wherever required.

Annotators involved in this work are also employees of the company and were thus paid for the work that they accomplished.

Publicly sourced datasets use permissive licenses (CC-BY, MIT, Apache-2.0).

Translated and web-scraped data risk reinforcing gender, ethnic, or regional biases; systematic Wolof-specific fairness audits (in collaboration with local linguists) should be conducted, as well as alignment work to mitigate these biases. Likewise, safety evaluation includes English-centric checks (WildJailbreak, HaluEval); Wolof-specific red-teaming (phishing, culturally inappropriate responses) should be conducted.

Deployment may displace customer-service labor; we advocate human-in-the-loop workflows where models handle routine queries and escalate complex cases to preserve employment.

Fine-tuning and inference for the benchmark evolves a large quantity of carbon emission. These emissions are measured and taken into consideration when planning experiments to limit them.

\section*{Acknowledgments}

The authors would like to thank the human annotators involved in the numerous stages of this work.

\section*{Appendix Contents}
\begingroup
\small
\setlength{\parindent}{0pt}
\setlength{\parskip}{0.2\baselineskip}
\newcommand{\appendixcontentsentry}[2]{%
  \hyperref[#1]{Appendix~\ref*{#1}. #2}\dotfill\pageref{#1}\par}
\newcommand{\appendixcontentssubentry}[2]{%
  \hspace*{1em}\hyperref[#1]{\ref*{#1}. #2}\dotfill\pageref{#1}\par}
\appendixcontentsentry{app:score-params}{Aggregate Score by Model Size}
\appendixcontentsentry{app:instruction-tuning}{Instruction Tuning Details}
\appendixcontentssubentry{sec:appendix-data}{Data}
\appendixcontentssubentry{app:training-setup}{Training Setup}
\appendixcontentsentry{app:alignment}{Preference Alignment Details}
\appendixcontentssubentry{app:alignment-data}{Data}
\appendixcontentssubentry{app:alignment-setup}{Training Setup}
\appendixcontentsentry{app:human-assessment}{Human Assessment of Translation and LLM Judging}
\appendixcontentssubentry{sec:appendix-translation}{Translation Quality}
\appendixcontentssubentry{app:llm-as-a-judge}{LLM-as-a-Judge Reliability}
\appendixcontentsentry{app:developmental-results}{Developmental Results}
\appendixcontentssubentry{app:ablation}{Instruction Ablation}
\appendixcontentssubentry{app:ongoing-cpt}{Continued Pre-training}
\appendixcontentssubentry{app:ongoing-tokenization}{Tokenization}
\appendixcontentsentry{app:stage-results}{Stage-by-Stage Results}
\appendixcontentsentry{app:skill-detailed-results}{Detailed Final Results}
\appendixcontentsentry{app:tasks}{Detailed Tasks}
\endgroup

\bibliography{custom}

\clearpage
\appendix

\section{Overall Score as a Function of Model Size}
\label{app:score-params}

\begin{figure*}[b!]
\centering
\includegraphics[width=\textwidth]{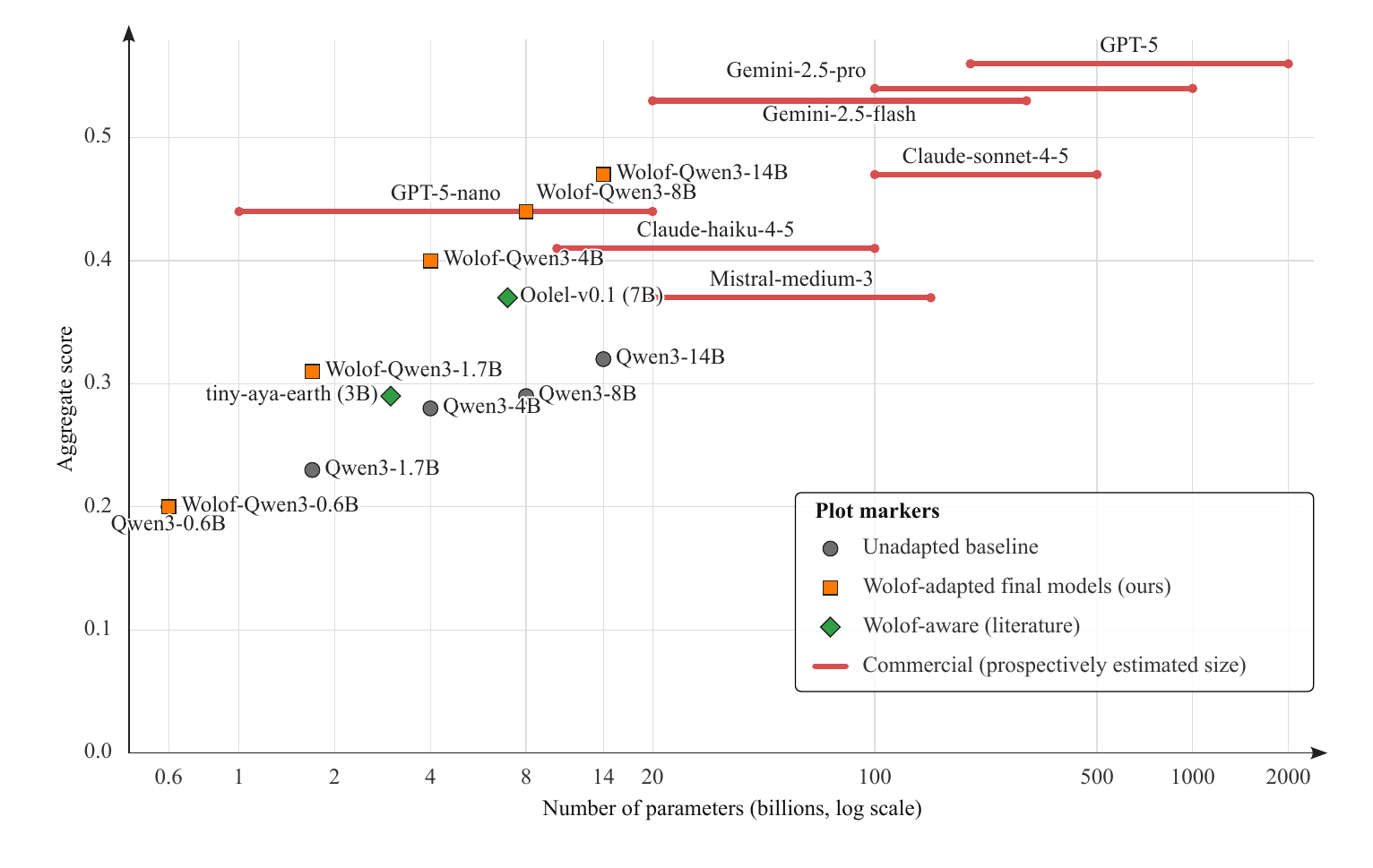}
\caption{Aggregate scores as a function of model size. Orange squares denote our Wolof-adapted models. Circles denote unadapted baselines, and diamonds denote Wolof-aware models from the literature. Commercial API systems are shown using purely indicative, prospectively estimated parameter ranges based on unofficial online information, rather than vendor-confirmed specifications.}
\label{fig:score-vs-params}
\end{figure*}

Figure~\ref{fig:score-vs-params} plots the overall scores against model size. For commercial API systems, the horizontal segment represents a prospectively estimated parameter-count range, not a disclosed or verified model size. These purely indicative estimates are derived from unofficial, publicly accessible online information and should not be interpreted as vendor-confirmed specifications.

\clearpage

\section{Instruction tuning details}
\label{app:instruction-tuning}

\subsection{Data}
\label{sec:appendix-data}

Table~\ref{tab:data-splits-detailed} provides complete statistics for training, validation, and test splits across all datasets, broken down by language and task type.

\begin{table*}[t]
  \centering
  \scriptsize
  \begin{tabular}{@{}llrrrll@{}}
    \hline
    \textbf{Dataset} & \textbf{Split} & \textbf{Examples} & \textbf{Tokens} & \textbf{Avg Len} & \textbf{Language} & \textbf{Type} \\
    \hline
    \multicolumn{7}{@{}l}{\emph{Translated General Instruction (Wolof)}} \\
    Alpaca (Wolof) & Train & 51,939 & 10,420,068 & 200.6 & Wolof & Single turn \\
     & Validation & 30 & 6,456 & 215.2 & Wolof & Single turn \\
    Dolly-15k (Wolof) & Train & 14,985 & 4,352,286 & 290.4 & Wolof & Single turn \\
     & Validation & 30 & 9,723 & 324.1 & Wolof & Single turn \\
    SAMSum (Wolof) & Train & 14,696 & 4,006,869 & 272.7 & Wolof & Single turn \\
     & Validation & 30 & 7,718 & 257.3 & Wolof & Single turn \\
    Anthropic Helpful (Wolof) & Train & 65,399 & 15,257,134 & 233.3 & Wolof & Single turn \\
     & Validation & 100 & 15,467 & 154.7 & Wolof & Single turn \\
    Hermes-217k (33k subset) & Train & 33,000 & 17,083,410 & 517.7 & Wolof & Single turn \\
     & Validation & 100 & 51,751 & 517.5 & Wolof & Single turn \\
    WiseNUT (Wolof) & Train & 19,700 & 5,315,879 & 269.8 & Wolof & Single turn \\
     & Validation & 100 & 11,428 & 114.3 & Wolof & Single turn \\
    Claire multilogue & Train & 15,317 & 22,485,029 & 1,468.0 & Wolof & Single turn \\
     & Validation & 100 & 202,862 & 2,028.6 & Wolof & Single turn \\
    \hline
    \multicolumn{7}{@{}l}{\emph{Bilingual French-Wolof}} \\
    Alwaly French$\rightarrow$Wolof & Train & 90,956 & 4,998,528 & 55.0 & French/Wolof & Single turn \\
     & Validation & 100 & 5,242 & 52.4 & French/Wolof & Single turn \\
    Alwaly Wolof$\rightarrow$French & Train & 90,956 & 4,999,316 & 55.0 & Wolof/French & Single turn \\
     & Validation & 100 & 5,258 & 52.6 & Wolof/French & Single turn \\
    \hline
    \multicolumn{7}{@{}l}{\emph{Native Conversational (Wolof)}} \\
    Claire one-to-one & Train & 6,757 & 5,692,874 & 842.5 & Wolof & Multiple turn \\
     & Validation & 100 & 102,599 & 1,026.0 & Wolof & Multiple turn \\
    In-house conversations & Validation & 50 & 33,883 & 677.7 & Wolof & Multiple turn \\
     & Test & 149 & 99,657 & 668.8 & Wolof & Multiple turn \\
    \hline
    \multicolumn{7}{@{}l}{\emph{Function Calling (English)}} \\
    English function calling & Train & 42,150 & 14,866,461 & 352.7 & English & Multiple turn \\
     & Validation & 2,218 & 791,172 & 356.7 & English & Multiple turn \\
    \hline
    \multicolumn{7}{@{}l}{\emph{Validation Only (English/French)}} \\
    OASST1-English & Validation & 188 & 81,840 & 435.3 & English & Multiple turn \\
    OASST1-French & Validation & 16 & 7,490 & 468.1 & French & Multiple turn \\
    Synthetic-French & Validation & 179 & 91,310 & 510.1 & French & Single turn \\
    Tulu-3-SFT-personas-IF & Validation & 26,982 & 567,693 & 21.0 & English & Single turn \\
    \hline
    \multicolumn{7}{@{}l}{\emph{Aggregates by Language}} \\
    \multicolumn{2}{@{}l}{Total Wolof (mono)} & 222,582 & 85,155,093 & 382.6 & Wolof & \\
    \multicolumn{2}{@{}l}{Total bilingual Wolof} & 182,112 & 10,008,344 & 55.0 & French/Wolof, Wolof/French & \\
    \multicolumn{2}{@{}l}{Total English} & 71,538 & 16,307,166 & 228.0 & English & \\
    \multicolumn{2}{@{}l}{Total French} & 195 & 98,800 & 506.7 & French & \\
    \hline
    \multicolumn{2}{@{}l}{\textbf{Total Training}} & \textbf{445,855} & \textbf{109,477,854} & \textbf{245.5} &  & \\
    \multicolumn{2}{@{}l}{\textbf{Total Validation}} & \textbf{30,423} & \textbf{1,991,892} & \textbf{65.5} &  & \\
    \multicolumn{2}{@{}l}{\textbf{Total Test}} & \textbf{149} & \textbf{99,657} & \textbf{668.8} &  & \\
    \hline
  \end{tabular}
  \caption{Dataset statistics by split. \textbf{Avg Len} is the average number of tokens per example. Wolof-containing data (monolingual and bilingual) accounts for approximately 90\% of training examples.}
  \label{tab:data-splits-detailed}
\end{table*}

The qualitative descriptions below follow the same source families as Table~\ref{tab:training-data}, while Table~\ref{tab:data-splits-detailed} reports the corresponding split-level statistics.

\subsubsection{Datasets}

\paragraph{Alwaly.}
Alwaly provides native bilingual French-Wolof text derived from educational and religious sources \citep{alwaly2026woloffrench}. We use the corpus in both directions: French inputs with Wolof outputs, and Wolof inputs with French outputs. These examples provide authentic Wolof text and help preserve translation ability between the two languages most relevant to the target deployment setting.

\paragraph{Alpaca.}
Alpaca is a general-purpose instruction-following dataset covering question answering, classification, creative generation, and factual knowledge queries \citep{taori2023alpaca}. We machine-translated the instruction and response fields into Wolof and used the resulting examples to broaden single-turn instruction coverage beyond translation-style supervision.

\paragraph{Dolly-15k.}
Dolly-15k is a human-written instruction corpus released by Databricks \citep{databricks2023dolly15k}. Its categories include brainstorming, classification, closed question answering, generation, information extraction, open question answering, and summarization. After translation into Wolof, it contributes diverse single-turn supervision with relatively natural human-authored prompts.

\paragraph{SAMSum.}
SAMSum contains messenger-like conversations paired with human-written summaries \citep{gliwa2019samsum}. We translated the data into Wolof and formatted it as instruction-style dialogue summarization, which adds supervision for compressing conversational content while still fitting the single-turn training format used in the main mixture.

\paragraph{Helpful instructions.}
The Helpful instructions subset is derived from the HuggingFaceH4 helpfulness data \citep{huggingfaceh42023helpful}, including examples from Anthropic helpfulness data \citep{bai2022training}. It contains assistant-style interactions that were translated into Wolof and converted into instruction examples. This source complements Alpaca and Dolly by emphasizing helpful response behavior and natural instruction following.

\paragraph{Hermes-217k.}
Hermes-217k is a large synthetic instruction collection with coverage of coding, mathematics, reasoning, knowledge, and general QA-style tasks \citep{hermes217k}. We use a curated translated subset to add task diversity and reasoning-oriented supervision to the Wolof instruction mixture.

\paragraph{WiseNUT.}
WiseNUT contributes Natural-Instructions-style examples translated into Wolof \citep{wang2022supernaturalinstructions}. The retained subset provides QA-oriented and speaker-identification prompts, adding factual and structured instruction patterns that complement the broader general-purpose sources.

\paragraph{Claire multilogue.}
Claire multilogue is built from the OpenLLM-France Claire-Dialogue-French-0.1 dataset \citep{hunter2023claire} by extracting dialogues with at least three speakers. For each dialogue, one speaker is randomly selected, one of that speaker's turns is masked, and the model is trained to reconstruct the masked turn from the full dialogue context. The resulting examples are dialogue-derived but are used as a single-turn masked-turn completion objective.

\paragraph{Claire one-to-one.}
Claire one-to-one is built from the same Claire source by extracting dialogues with exactly two speakers. These samples are formatted as standard multi-turn conversations, with a system message providing short speaker profiles and dialogue context, followed by alternating user and assistant turns. The profiles, context, and messages were machine-translated into Wolof, giving the model direct supervision for turn-taking and persona-consistent Wolof dialogue.

\paragraph{In-house conversations.}
The in-house conversation split contains de-identified Wolof customer-care dialogues. It is used to monitor deployment-relevant conversational behavior, including service terminology, turn-taking, and customer-intent patterns. Raw logs are not released, but the split informs validation and test coverage for the industrial customer-service setting.

\paragraph{Function calling instructions.}
The English function-calling block aggregates examples from \texttt{tool-calling-mix} \citep{younissk2024toolcallingmix}, \texttt{xLAM k60k} \citep{zhang2024xlam,liu2024apigen}, and synthetic examples generated with Gemini 2.5 Pro. We deduplicated and filtered the public sources to keep examples where at least one tool is actually invoked and to exclude BFCL-derived examples. The synthetic portion targets Python, Java, and JavaScript function signatures with 1--20 available tools per example, following the Berkeley Function Calling Leaderboard distribution of tool-count complexity. This block is kept in English to preserve structured tool-use behavior and support cross-lingual transfer to Wolof-facing assistant use.

\subsubsection{Validation data}

First, we included Bilingual French-Wolof and Translated General validation sets. Then, we added stratified validation sets by language and task type. For English and French conversations, we use established conversational benchmarks including OASST1 \citep{kopf2023openassistant}, synthetic French dialogues generated by ourselves, and Tulu-3-SFT-personas \citep{ge2024personas} to assess conversational quality and instruction-following in these languages. Wolof conversational validation relies on the held-out split of \texttt{claire\_one\_to\_one\_dialogue} and an in-house customer-care conversation dataset to evaluate dialogue coherence and turn-taking in Wolof. Validation for structured output and tool use employs the held-out split of \texttt{en\_function\_calling} and \texttt{tulu-3-sft-personas-if} \citep{zhou2023ifeval} (instruction-following variant). A composite validation score aggregating performance across these dimensions guided model checkpoint selection.

\subsection{Training Setup}
\label{app:training-setup}

As detailed in Table~\ref{tab:training-setup}, training was conducted on a single-node setup with 2 NVIDIA H100 80GB and H200 140GB GPUs. Main experiments ran for 3 epochs and typically lasted from a few hours to several days, depending on model size and corpus scale. Optimization used paged AdamW (\texttt{paged\_adamw\_32bit}) with a cosine learning-rate schedule, 5\,\% warmup, and a peak learning rate of $2 \times 10^{-5}$. Regularization included gradient clipping at 1.0 and weight decay of 0.1.

Models were trained in bfloat16 with DeepSpeed ZeRO-3 and gradient checkpointing. The batch size was 16 per device with gradient accumulation over 32 steps, yielding an effective batch size of 1024. The maximum sequence length was set to 2048 tokens, and attention was accelerated with FlashAttention2 (\texttt{flash\_attention\_2}). Data loading used 4 workers per GPU, with persistent workers, pinned memory, and a prefetch factor of~2.

Checkpoints were saved at the end of each epoch, keeping only the most recent one and selecting models based on validation loss. Validation was performed every 10 steps on EN/FR/WO and function-calling splits using a composite score for model selection. For reproducibility, training relied on YAML configurations, fixed seeds (42), and TensorBoard logging. Carbon emissions were tracked with CodeCarbon, reporting 1.6 kg CO$_2$eq for the main run.

\begin{table}[t!]
\centering
\scriptsize
\begin{tabular}{@{}lp{5cm}@{}}
\toprule
\textbf{Category} & \textbf{Details} \\
\midrule
Hardware & Single node; 2$\times$ NVIDIA H100 80GB and H200 140GB GPUs \\
Training duration & 3 epochs; a few hours to a few days depending on model and corpus size \\
Optimizer & Paged AdamW (\texttt{paged\_adamw\_32bit}) \\
Learning rate schedule & Cosine decay with 5\% warmup \\
Peak learning rate & $2 \times 10^{-5}$ \\
Regularization & Gradient clipping (max-norm 1.0); weight decay 0.1 \\
Precision / parallelism & bfloat16; DeepSpeed ZeRO-3; gradient checkpointing \\
Batching & Batch size 16/device; gradient accumulation 32; effective batch size 1024 \\
Sequence length & 2048 tokens; padding free \\
Attention & FlashAttention2 (\texttt{flash\_attention\_2}) \\
Data loading & 4 workers/GPU; persistent workers; pinned memory; prefetch factor 2 \\
Checkpointing & End-of-epoch saves; keep latest only; selected by validation loss \\
Validation & Every 10 steps on EN/FR/WO + function-calling splits; composite score used \\
Reproducibility & YAML configs; fixed seed (42); TensorBoard logs \\
Emissions tracking & CodeCarbon; 1.6 kg CO$_2$eq for main run \\
\bottomrule
\end{tabular}
\caption{Instruction tuning (SFT) training setup.}
\label{tab:training-setup}
\end{table}

\section{Preference Alignment Details}
\label{app:alignment}

\subsection{Data}
\label{app:alignment-data}

\begin{table*}[t!]
\centering
\scriptsize
\begin{tabular}{@{}lrrrrr@{}}
\hline
\textbf{Dataset} & \textbf{Split} & \textbf{Examples} & \textbf{Avg Len Prompt} & \textbf{Avg Len Chosen} & \textbf{Avg Len Rejected} \\
\hline
\multicolumn{6}{@{}l}{\emph{Instruction-Following}} \\
Chat \& WildChat & Train & 27,284 & 404.3 & 383.4 & 803.9 \\
\phantom{X} & Validation & 2896 & 366.5 & 378.3 & 895.1 \\
Dolci Instruct Precise IF & Train & 31,449 & 197.1 & 555.3 & 537.2 \\
\phantom{X} & Validation & 3534 & 194.1 & 558.0 & 554.7 \\
Dolci Instruct Persona Precise IF & Train & 5,994 & 58.9 & 413.1 & 322.9 \\
\phantom{X} & Validation & 644 & 61.4 & 453.4 & 366.9 \\
OpenAssistant & Train & 447 & 35.1 & 444.0 & 867.8 \\ 
\phantom{X} & Validation & 48 & 37.2 & 492.8 & 1,096.0 \\ \hline
\multicolumn{6}{@{}l}{\emph{Mathematics}} \\
Math Tülu 3 Persona MATH & Train & 13,225 & 288.6 & 1,739.6 & 1,257.9 \\
\phantom{X} & Validation & 1450 & 290.9 & 1,703.9 & 1,258.5 \\
Tülu 3 Persona Algebra & Train & 1,812 & 197.7 & 1,265.0 & 957.9 \\
\phantom{X} & Validation & 178 & 203.0 & 1,142.5 & 939.5 \\
Tülu 3 Persona GSM & Train & 4,504 & 114.1 & 328.3 & 278.3 \\
\phantom{X} & Validation & 536 & 112.8 & 302.4 & 262.5 \\
OpenMathInstruct & Train & 4,842 & 59.3 & 295.2 & 285.9 \\ 
\phantom{X} & Validation & 498 & 59.4 & 294.2 & 308.7 \\ \hline
\multicolumn{6}{@{}l}{\emph{Coding}} \\
Coding Dolci Instruct Python Algorithms & Train & 21,695 & 168.0 & 525.0 & 357.6 \\
\phantom{X} & Validation & 2430 & 168.8 & 538.7 & 383.6 \\
Tülu 3 Persona Python & Train & 4,145 & 239.4 & 406.7 & 262.9 \\
\phantom{X} & Validation & 446 & 250.7 & 404.1 & 238.0 \\
Evol CodeAlpaca & Train & 11,668 & 162.2 & 946.0 & 649.9 \\
\phantom{X} & Validation & 1348 & 172.2 & 950.9 & 723.7 \\ \hline
\multicolumn{6}{@{}l}{\emph{Safety}} \\
Safety CoCoNot & Train & 1,966 & 16.1 & 506.5 & 314.9 \\
\phantom{X} & Validation & 268 & 16.9 & 651.2 & 384.4 \\
WildGuardMix & Train & 10,836 & 98.0 & 651.4 & 405.9 \\
\phantom{X} & Validation & 1178 & 99.4 & 624.5 & 404.1 \\
WildJailbreak & Train & 11,198 & 17.4 & 594.5 & 364.1 \\
\phantom{X} & Validation & 1260 & 17.1 & 564.5 & 336.7 \\ \hline
\multicolumn{6}{@{}l}{\emph{General and science knowledge}} \\
Science SciRiff & Train & 7,973 & 1,026.9 & 157.5 & 160.3 \\
\phantom{X} & Validation & 910 & 983.7 & 128.7 & 149.3 \\
Dolci Instruct OpenThought3+ Science & Train & 23,458 & 280.1 & 1,432.2 & 877.5 \\
\phantom{X} & Validation & 2668 & 282.5 & 1,446.0 & 874.8 \\
UltraFeedback & Train & 20,096 & 208.6 & 213.5 & 431.8 \\
\phantom{X} & Validation & 2214 & 222.9 & 202.3 & 408.2 \\ \hline
\multicolumn{6}{@{}l}{\emph{Multilingual}} \\
Multilingual Aya & Train & 5,902 & 115.5 & 560.9 & 2,831.5 \\
\phantom{X} & Validation & 598 & 96.4 & 664.0 & 2,987.9 \\ \hline
\multicolumn{6}{@{}l}{\emph{Long-context prompts}} \\
Multiturn Dolci Instruct Self-Talk & Train & 4,449 & 816.5 & 178.7 & 221.4 \\
\phantom{X} & Validation & 542 & 853.8 & 192.6 & 180.8 \\
Dolci Instruct Synthetic Context & Train & 4,512 & 730.2 & 175.3 & 270.8 \\
\phantom{X} & Validation & 472 & 694.3 & 177.4 & 278.9 \\ \hline
\multicolumn{6}{@{}l}{\emph{Others}} \\
Other TableGPT & Train & 1,085 & 769.2 & 182.3 & 147.7 \\
\phantom{X} & Validation & 152 & 848.2 & 206.5 & 114.9 \\
FLAN & Train & 14,486 & 310.5 & 67.5 & 146.5 \\
\phantom{X} & Validation & 1622 & 306.0 & 66.7 & 120.6 \\ 
DaringAnteater & Train & 778 & 463.8 & 299.3 & 473.1 \\
\phantom{X} & Validation & 88 & 532.0 & 261.9 & 458.5 \\ \hline
\hline
\multicolumn{6}{@{}l}{\emph{Aggregates by category}} \\
Instruction-Following & All & 72,296 & 268.2 & 470.0 & 636.2 \\
Mathematics & All & 27,045 & 204.3 & 1,153.2 & 861.4 \\
Coding & All & 41,732 & 174.5 & 644.2 & 442.0 \\
Safety & All & 26,706 & 53.6 & 611.6 & 378.1 \\
General and science knowledge & All & 57,319 & 368.0 & 760.1 & 591.7 \\
Multilingual & All & 6,500 & 113.7 & 570.4 & 2,845.9 \\
Long-context prompts & All & 9,975 & 773.7 & 177.9 & 244.3 \\
Others & All & 18,211 & 349.5 & 86.3 & 159.5 \\
\hline
\textbf{Total} & \textbf{Train} & \textbf{233,804} & -- & -- & -- \\
\textbf{Total} & \textbf{Validation} & \textbf{25,980} & -- & -- & -- \\
\textbf{Total} & \textbf{All} & \textbf{259,784} & -- & -- & -- \\
\hline
\end{tabular}
\caption{Composition and Qwen3 token-length statistics for Dolci-Instruct-DPO. Token averages are computed using Qwen3 tokenizer.}
\label{tab:dolci-instruct-dpo-token-lengths}
\end{table*}

Table~\ref{tab:dolci-instruct-dpo-token-lengths} details the composition, split sizes, and sequence-length profiles across the main categories of the Dolci-Instruct-DPO dataset. The corpus contains 259,784 total pairwise comparisons partitioned into a 90/10 train/validation split (233,804 training and 25,980 validation pairs) which are stratified consistently across all individual sources.
\paragraph{Instruction-Following.} The dataset counts for 72,296 preference pairs (27.8\% of the corpus), it combines real-world user queries with strict constraint-following tasks with natural user interactions from Chat and WildChat \citep{zhao2024wildchat} alongside volunteer-annotated dialog trees from OpenAssistant Conversations (OASST1; \citealt{köpf2023openassistant}). These interactions are complemented by synthetic instances from Dolci Instruct Precise IF and Persona Precise IF, which use verifiable constraints (e.g., specific word counts, negative constraints, and exact formatting tags).

\paragraph{General and Scientific Knowledge.} The corpus covers 57,319 pairs (22.1\%), this partition provides supervision for factual correctness, multi-step explanations, and domain-specific knowledge. It incorporates SciRiff \citep{wadden2024sciriff}, an instruction benchmark covering biomedical and scientific information extraction, as well as Dolci Instruct OpenThought3, with detailed reasoning traces. These examples are combined with UltraFeedback \citep{cui2023ultrafeedback}, where outputs from different models scored across instruction-following, truthfulness, and honesty metrics using GPT-4 as a judge. 

\paragraph{Coding.} The coding subset contains 41,732 pairs (16.1\%) focused on program summaries, algorithmic problem-solving, and code execution correctness. It integrates Evol-CodeAlpaca \citep{luo2023wizardcoder}, produced via iterative prompting (e.g., adding constraints, improving complexity), with Tülu~3 Persona Python \citep{lambert2024tulu3} and Dolci Instruct Python Algorithms. Tülu~3 Persona tasks present modular problems across different programmer skills. 

\paragraph{Mathematics.} This category contains 27,045 pairs (10.4\%) designed to strengthen quantitative reasoning. It uses data from OpenMathInstruct-1 \citep{toshev2024openmathinstruct} and the math parts of Tülu~3 Persona \citep{lambert2024tulu3}, like GSM8K, MATH, and algebra problems. Pairs are typically formed by ranking correct final answers and valid intermediate steps over hallucinated reasoning paths. 

\paragraph{Safety and Alignment.} The dataset presents 26,706 pairs (10.3\%), The pairs teach the model to clearly say no to dangerous requests (such as cyberattacks), while still giving helpful answers to safe questions that only contain sensitive words.

\paragraph{Multilingual} This part has 6,500 pairs (2.5\% of the data) taken from the Aya collection \citep{ahmed2024aya}, covering many different languages. Even though Wolof is not included here, this data helps the model remember how to work in multiple languages and stop it from forgetting other languages after training. 

\paragraph{Long-Context Reasoning.} This category counts 9,975 pairs (3.8\% of the data) made from the Dolci Instruct Self-Talk and Synthetic Context datasets. The inputs are quite long (averaging 773.7 tokens and over 850 in validation) and include long chat histories, text snippets, and complex instructions. These pairs teach the model not to get lost or ignore parts of long texts. Instead, it learns to find key details in long prompts and give short, direct answers (averaging 177.9 chosen tokens).

\paragraph{Others.} This last group contains 18,211 pairs (7.0\% of the data) covering tables, classic NLP tasks, and general questions. It includes table data from TableGPT \citep{zha2023tablegpt}, standard tasks from the FLAN collection \citep{longpre2023flan}, and multi-turn conversations from NVIDIA's DaringAnteater. Good answers here follow special formats and extract data cleanly, which helps the model stay reliable across many different kinds of everyday tasks.

\subsection{Training Setup}
\label{app:alignment-setup}

As detailed in Table~\ref{tab:alignment-setup}, experiments were conducted on a single node with either $2 \times$ NVIDIA H100 (80GB) GPUs for the Qwen3-14B model, or $1 \times$ H200 (140GB) GPU for the 8B models. Training durations ranged from approximately one day for smaller models (0.6B-1.7B) to eight days for the largest models. All trainings targeted an effective batch size of 512, with the same learning-rate and scheduler. Evaluations as well as checkpointing were done each 40 steps, and the best model was selected using validation loss (\texttt{eval\_validation\_loss}).

To save energy and time, hyper-parameters tuning was performed on Qwen3-0.6B model, iterating over learning rates ($5 \times 10^{-5}, 5 \times 10^{-6}, 5 \times 10^{-7}, 5 \times 10^{-8}$) and $\beta$ values ($0.01, 0.02, 0.05, 0.1, 0.2, 0.3$), extending recommendations from \citet{ivison2024unpackingdpoppodisentangling}. The two best-performing configurations were then validated on an 8B model. Finally, the optimal configuration ($\beta = 0.2$ and a peak learning rate of $5 \times 10^{-6}$) was selected for all final training runs.

\begin{table}[b!]
\centering
\scriptsize
\begin{tabular}{@{}lp{5cm}@{}}
\toprule
\textbf{Category} & \textbf{Details} \\
\midrule
Hardware & Single node; 2$\times$ NVIDIA H100 80GB and 1$\times$ H200 140GB GPUs \\
Training duration & 2 epochs; from a day to a few days depending on model size \\
Beta & 0.2 \\
Learning rate schedule & Cosine decay with 5\% warmup \\
Peak learning rate & $5 \times 10^{-6}$ \\
Regularization & Gradient clipping (max-norm 1.0); weight decay 0.01 \\
Precision / parallelism & bfloat16; DeepSpeed ZeRO-3; gradient checkpointing \\
Batching & effective batch size 512; Batch size 16/device; gradient accumulation 32 when 1 GPU was used \\
Sequence length & 1024 tokens; padding free \\
Attention & FlashAttention2 (\texttt{flash\_attention\_2}) \\
Data loading & 4 workers/GPU; persistent workers; pinned memory; prefetch factor 2 \\
Checkpointing & 40 steps saves; best model selected by validation loss \\
Validation & Every 40 steps on validation split \\
Reproducibility & YAML configs; fixed seed (42); TensorBoard logs \\
\bottomrule
\end{tabular}
\caption{Preference alignment (DPO) training setup.}
\label{tab:alignment-setup}
\end{table}

\section{Human Assessment of Translation and LLM Judging}
\label{app:human-assessment}

\subsection{Translation Quality}
\label{sec:appendix-translation}
To assess the quality of the automatically translated Wolof data, we conducted a manual evaluation on a randomly selected, domain-stratified sample of 50 English-Wolof sentence pairs from FLORES-200. The translations were produced by Google Translate and evaluated by two native Wolof speakers using two five-point scales: semantic adequacy (``correctness of meaning'') and linguistic quality of the Wolof output. The translations received an average score of 4.59/5 for semantic adequacy and 4.78/5 for linguistic quality, indicating that they were generally faithful to the source sentences and grammatically well formed. The inter-annotator agreement for semantic adequacy and linguistic quality are $64\%$ and $68\%$, respectively, rising to $88\%$ and $100\%$ when a $\mp{}1$ margin is tolerated.

On the same samples, automatic evaluation yielded a METEOR score of 0.24. The Pearson correlation between sentence-level METEOR scores and human semantic-adequacy ratings was $r = 0.12$, suggesting little relationship between METEOR and perceived translation quality in this setting. This discrepancy is due to specific aspects of Wolof, that penalize lexical-based scores such as METEOR, whereas the translation quality is judged acceptable by human:
\begin{itemize}
    \item Syntactic flexibility: Wolof may allow more than one valid syntactic realization of the same proposition. A translation can therefore preserve the meaning while using a word order that differs from the FLORES-200 reference. 
    \item Orthographic variation: Wolof orthography is not always completely standardized across resources and speakers. Variants such as ñoo and ñu, or bi and bu, may be acceptable depending on the grammatical and contextual usage, while still producing limited character or word overlap with the reference. 
    \item Accepted lexical borrowings and alternatives: The system may use lexical borrowings or alternative terms that are readily understood and accepted by Wolof speakers. For example, the translation may use pied for “feet” or tournoi for “tournament,” whereas the FLORES-200 reference uses different Wolof equivalents. 
\end{itemize}

These results motivate the use of human assessment alongside automatic metrics when evaluating machine-translated Wolof data; accordingly, we retain the automatically generated data as silver data and explicitly document its provenance.

\subsection{LLM-as-a-Judge Reliability}
\label{app:llm-as-a-judge}

While LLM-as-a-judge is rather a common practice for the most popular languages, the question of using an LLM to judge free text in Wolof is less documented. Hence, gemini-2.5 models have been validated with 3~Wolof speakers on held-out data.

In this experiment, users had to on agree or disagree with the scores provided by the LLM judge for 450~predictions of various models on Aya tasks. The agreement rate was 95\% for gemini-2.5-flash and 99\,\% for gemini-2.5-pro.

\clearpage
\onecolumn

\section{Developmental results}
\label{app:developmental-results}

This appendix reports the intermediate development experiments using the same normalized skill aggregates as the main automatic evaluation.

\subsection{Instruction Ablation}
\label{app:ablation}

Table~\ref{tab:development_ablation_results} compares the Qwen2.5-7B-Instruct base model with staged Wolof-adapted variants. The comparison isolates the contribution of instruction tuning, function-calling data, conversational data, and the final combined recipe before transferring the setup to Qwen3.

\begin{table*}[ht]
\centering
\scriptsize
\begin{fitwidthtable}
\begin{tabular}{@{}l@{}l@{~}|@{~~}l@{~~}l@{~~}l@{~~}l@{~~}l@{~~}l@{~~}l@{~~}l@{~~}l@{~~}l@{~~}l@{~~}l@{~~}l@{~~}l@{~~}l@{}}
\toprule
 & Avg. sc.$^\uparrow{}$ &\multicolumn{2}{@{}c@{}}{English} &\multicolumn{2}{@{}c@{}}{French} &\multicolumn{2}{@{}c@{}}{Wolof} &\multicolumn{1}{@{}c@{}}{Know.} & \multicolumn{1}{@{}c@{}}{Conv.} & \multicolumn{1}{@{}c@{}}{Reas.} & \multicolumn{1}{@{}c@{}}{Code} & \multicolumn{1}{@{}c@{}}{Format.} & \multicolumn{1}{@{}c@{}}{Instr.} & \multicolumn{1}{@{}c@{}}{Func.} & \multicolumn{1}{@{}c@{}}{Lg ctx.} & \multicolumn{1}{@{}c@{}}{Safet.} \\
\cline{3-8}
& {\tiny (\& rank$^\downarrow{}$)} & \multicolumn{1}{@{}c@{}}{Und.{\tiny (22)}} & \multicolumn{1}{@{}c@{}}{Gen.{\tiny (8)}} & \multicolumn{1}{@{}c@{}}{Und.{\tiny (16)}} & \multicolumn{1}{@{}c@{}}{Gen.{\tiny (7)}} & \multicolumn{1}{@{}c@{}}{Und.{\tiny (30)}} & \multicolumn{1}{@{}c@{}}{Gen.{\tiny (14)}} & \multicolumn{1}{@{}c@{}}{{\tiny (7)}} & \multicolumn{1}{@{}c@{}}{{\tiny (6)}} & \multicolumn{1}{@{}c@{}}{{\tiny (6)}} & \multicolumn{1}{@{}c@{}}{{\tiny (1)}} & \multicolumn{1}{@{}c@{}}{{\tiny (22)}} & \multicolumn{1}{@{}c@{}}{{\tiny (4)}} & \multicolumn{1}{@{}c@{}}{{\tiny (1)}} & \multicolumn{1}{@{}c@{}}{{\tiny (2)}} & \multicolumn{1}{@{}c@{}}{{\tiny (4)}} \\
\midrule
\multicolumn{17}{@{}l}{\textbf{Baseline open-weight models}}\\
Qwen2.5-7B-Instruct & \scoretop{\sr{.31}{2.5}} & \textbf{\scoretop{\sr{.48}{1}}} & \scoretop{\sr{.45}{3}} & \textbf{\scoretop{\sr{.37}{1}}} & \scoretop{\sr{.42}{3}} & \scoretop{\sr{.12}{4}} & \scoretop{\sr{.06}{4}} & \scorehigh{\sr{.23}{4}} & \scorehigh{\sr{.32}{4}} & \textbf{\scoretop{\sr{.37}{1}}} & \textbf{\scoretop{\sr{.80}{1}}} & \scorehigh{\sr{.28}{3}} & \textbf{\scoretop{\sr{.54}{1}}} & \textbf{\scoretop{\sr{.62}{1}}} & \scoremid{\sr{.44}{3}} & \scorehigh{\sr{.44}{4}} \\

\midrule
\multicolumn{17}{@{}l}{\textbf{Wolof-adapted open-weight models (ours)}}\\
Wolof-Qwen2.5-7B (SFT on single + conv + func. call.) & \scoretop{\sr{.36}{1.9}} & \scorehigh{\sr{.43}{2}} & \scoretop{\sr{.47}{2}} & \scoretop{\sr{.35}{2}} & \textbf{\scoretop{\sr{.47}{1}}} & \textbf{\scoretop{\sr{.26}{1}}} & \textbf{\scoretop{\sr{.26}{1}}} & \scoretop{\sr{.28}{2}} & \scoretop{\sr{.42}{2}} & \scoretop{\sr{.36}{2}} & \scorehigh{\sr{.67}{2}} & \textbf{\scoretop{\sr{.31}{1}}} & \scoremid{\sr{.28}{2}} & \scoremid{\sr{.01}{4}} & \scoremid{\sr{.43}{4}} & \textbf{\scoretop{\sr{.50}{1}}} \\

Wolof-Qwen2.5-7B (SFT on single only) & \scoretop{\sr{.34}{2.3}} & \scorehigh{\sr{.43}{2}} & \textbf{\scoretop{\sr{.48}{1}}} & \scoretop{\sr{.33}{3}} & \scoretop{\sr{.46}{2}} & \textbf{\scoretop{\sr{.26}{1}}} & \textbf{\scoretop{\sr{.26}{1}}} & \scoretop{\sr{.28}{2}} & \textbf{\scoretop{\sr{.44}{1}}} & \underline{\scorelow{\sr{.28}{5}}} & \scorelow{\sr{.28}{4}} & \scorehigh{\sr{.27}{4}} & \scoremid{\sr{.24}{3}} & \scoremid{\sr{.06}{3}} & \textbf{\scoretop{\sr{.50}{1}}} & \scoretop{\sr{.49}{2}} \\

Wolof-Qwen2.5-7B (SFT on single + func. call.) & \scoretop{\sr{.33}{3.0}} & \scorehigh{\sr{.42}{4}} & \scoretop{\sr{.45}{3}} & \scorehigh{\sr{.29}{4}} & \scorehigh{\sr{.37}{4}} & \scoretop{\sr{.24}{3}} & \scoretop{\sr{.23}{3}} & \textbf{\scoretop{\sr{.30}{1}}} & \scorehigh{\sr{.35}{3}} & \scoremid{\sr{.32}{3}} & \underline{\scorelow{\sr{.25}{5}}} & \scoretop{\sr{.29}{2}} & \scoremid{\sr{.21}{4}} & \scorehigh{\sr{.39}{2}} & \textbf{\scoretop{\sr{.50}{1}}} & \scorehigh{\sr{.45}{3}} \\

Wolof-Qwen2.5-7B (SFT on single + conv) & \underline{\scorelow{\sr{-2.24}{4.7}}} & \underline{\scorelow{\sr{.20}{5}}} & \underline{\scorelow{\sr{.07}{5}}} & \underline{\scorelow{\sr{.07}{5}}} & \underline{\scorelow{\sr{.05}{5}}} & \underline{\scorelow{\sr{-4.89}{5}}} & \underline{\scorelow{\sr{-10.56}{5}}} & \underline{\scorelow{\sr{.15}{5}}} & \underline{\scorelow{\sr{-.04}{5}}} & \scoremid{\sr{.32}{3}} & \scorehigh{\sr{.57}{3}} & \underline{\scorelow{\sr{.12}{5}}} & \underline{\scorelow{\sr{.03}{5}}} & \underline{\scorelow{\sr{-.45}{5}}} & \underline{\scorelow{\sr{.41}{5}}} & \underline{\scorelow{\sr{.35}{5}}} \\

\bottomrule
\end{tabular}
\end{fitwidthtable}
\caption{Training-data ablation comparison on Qwen2.5-7B-Instruct: normalised scores (higher, better) and ranks (lower, better). \scorecolorlegend}
\label{tab:development_ablation_results}
\end{table*}

\subsection{Continued Pre-training}
\label{app:ongoing-cpt}

Table~\ref{tab:development_cpt_results} compares the Qwen2.5-7B baseline, the SFT-only recipe, and a continued-pre-training plus SFT variant. This experiment evaluates whether the available raw Wolof text is sufficient to justify a CPT stage before supervised adaptation.

\begin{table*}[ht]
\centering
\scriptsize
\begin{fitwidthtable}
\begin{tabular}{@{}l@{}l@{~}|@{~~}l@{~~}l@{~~}l@{~~}l@{~~}l@{~~}l@{~~}l@{~~}l@{~~}l@{~~}l@{~~}l@{~~}l@{~~}l@{~~}l@{~~}l@{}}
\toprule
 & Avg. sc.$^\uparrow{}$ &\multicolumn{2}{@{}c@{}}{English} &\multicolumn{2}{@{}c@{}}{French} &\multicolumn{2}{@{}c@{}}{Wolof} &\multicolumn{1}{@{}c@{}}{Know.} & \multicolumn{1}{@{}c@{}}{Conv.} & \multicolumn{1}{@{}c@{}}{Reas.} & \multicolumn{1}{@{}c@{}}{Code} & \multicolumn{1}{@{}c@{}}{Format.} & \multicolumn{1}{@{}c@{}}{Instr.} & \multicolumn{1}{@{}c@{}}{Func.} & \multicolumn{1}{@{}c@{}}{Lg ctx.} & \multicolumn{1}{@{}c@{}}{Safet.} \\
\cline{3-8}
& {\tiny (\& rank$^\downarrow{}$)} & \multicolumn{1}{@{}c@{}}{Und.{\tiny (22)}} & \multicolumn{1}{@{}c@{}}{Gen.{\tiny (8)}} & \multicolumn{1}{@{}c@{}}{Und.{\tiny (16)}} & \multicolumn{1}{@{}c@{}}{Gen.{\tiny (7)}} & \multicolumn{1}{@{}c@{}}{Und.{\tiny (30)}} & \multicolumn{1}{@{}c@{}}{Gen.{\tiny (14)}} & \multicolumn{1}{@{}c@{}}{{\tiny (7)}} & \multicolumn{1}{@{}c@{}}{{\tiny (6)}} & \multicolumn{1}{@{}c@{}}{{\tiny (6)}} & \multicolumn{1}{@{}c@{}}{{\tiny (1)}} & \multicolumn{1}{@{}c@{}}{{\tiny (22)}} & \multicolumn{1}{@{}c@{}}{{\tiny (4)}} & \multicolumn{1}{@{}c@{}}{{\tiny (1)}} & \multicolumn{1}{@{}c@{}}{{\tiny (2)}} & \multicolumn{1}{@{}c@{}}{{\tiny (4)}} \\
\midrule
\multicolumn{17}{@{}l}{\textbf{Baseline open-weight models}}\\
Qwen2.5-7B-Instruct & \scoremid{\sr{.31}{1.9}} & \textbf{\scoretop{\sr{.48}{1}}} & \scorehigh{\sr{.45}{2}} & \textbf{\scoretop{\sr{.37}{1}}} & \scoremid{\sr{.42}{2}} & \underline{\scorelow{\sr{.12}{3}}} & \underline{\scorelow{\sr{.06}{3}}} & \underline{\scorelow{\sr{.23}{3}}} & \underline{\scorelow{\sr{.32}{3}}} & \textbf{\scoretop{\sr{.37}{1}}} & \textbf{\scoretop{\sr{.80}{1}}} & \textbf{\scoretop{\sr{.28}{1}}} & \textbf{\scoretop{\sr{.54}{1}}} & \textbf{\scoretop{\sr{.62}{1}}} & \underline{\scorelow{\sr{.44}{3}}} & \scorehigh{\sr{.44}{2}} \\

\midrule
\multicolumn{17}{@{}l}{\textbf{Wolof-adapted open-weight models (ours)}}\\
Wolof-Qwen2.5-7B (SFT) & \scoretop{\sr{.35}{1.5}} & \scorehigh{\sr{.43}{2}} & \textbf{\scoretop{\sr{.46}{1}}} & \scorehigh{\sr{.34}{2}} & \textbf{\scoretop{\sr{.45}{1}}} & \textbf{\scoretop{\sr{.26}{1}}} & \scoretop{\sr{.26}{2}} & \textbf{\scoretop{\sr{.30}{1}}} & \textbf{\scoretop{\sr{.43}{1}}} & \scoremid{\sr{.29}{2}} & \scoremid{\sr{.30}{2}} & \textbf{\scoretop{\sr{.28}{1}}} & \scorelow{\sr{.25}{2}} & \scoremid{\sr{.12}{2}} & \textbf{\scoretop{\sr{.54}{1}}} & \textbf{\scoretop{\sr{.49}{1}}} \\

Wolof-Qwen2.5-7B (CPT + SFT) & \underline{\scorelow{\sr{.30}{2.5}}} & \underline{\scorelow{\sr{.37}{3}}} & \underline{\scorelow{\sr{.41}{3}}} & \underline{\scorelow{\sr{.31}{3}}} & \underline{\scorelow{\sr{.39}{3}}} & \scorehigh{\sr{.22}{2}} & \textbf{\scoretop{\sr{.27}{1}}} & \scoretop{\sr{.29}{2}} & \textbf{\scoretop{\sr{.43}{1}}} & \underline{\scorelow{\sr{.26}{3}}} & \underline{\scorelow{\sr{.16}{3}}} & \underline{\scorelow{\sr{.22}{3}}} & \underline{\scorelow{\sr{.22}{3}}} & \underline{\scorelow{\sr{.01}{3}}} & \scorehigh{\sr{.51}{2}} & \underline{\scorelow{\sr{.34}{3}}} \\

\bottomrule
\end{tabular}
\end{fitwidthtable}
\caption{Continued-pretraining comparison on Qwen2.5-7B-Instruct: normalised scores (higher, better) and ranks (lower, better). \scorecolorlegend}
\label{tab:development_cpt_results}
\end{table*}

\subsection{Tokenization}
\label{app:ongoing-tokenization}

Table~\ref{tab:development_tokenization_results} compares the current Qwen2.5-7B instruction-tuned setup with a tokenizer-extension variant. This experiment tests whether vocabulary adaptation improves Wolof coverage while preserving the broader skill profile.

\begin{table*}[ht]
\centering
\scriptsize
\begin{fitwidthtable}
\begin{tabular}{@{}l@{}l@{~}|@{~~}l@{~~}l@{~~}l@{~~}l@{~~}l@{~~}l@{~~}l@{~~}l@{~~}l@{~~}l@{~~}l@{~~}l@{~~}l@{~~}l@{~~}l@{}}
\toprule
 & Avg. sc.$^\uparrow{}$ &\multicolumn{2}{@{}c@{}}{English} &\multicolumn{2}{@{}c@{}}{French} &\multicolumn{2}{@{}c@{}}{Wolof} &\multicolumn{1}{@{}c@{}}{Know.} & \multicolumn{1}{@{}c@{}}{Conv.} & \multicolumn{1}{@{}c@{}}{Reas.} & \multicolumn{1}{@{}c@{}}{Code} & \multicolumn{1}{@{}c@{}}{Format.} & \multicolumn{1}{@{}c@{}}{Instr.} & \multicolumn{1}{@{}c@{}}{Func.} & \multicolumn{1}{@{}c@{}}{Lg ctx.} & \multicolumn{1}{@{}c@{}}{Safet.} \\
\cline{3-8}
& {\tiny (\& rank$^\downarrow{}$)} & \multicolumn{1}{@{}c@{}}{Und.{\tiny (22)}} & \multicolumn{1}{@{}c@{}}{Gen.{\tiny (8)}} & \multicolumn{1}{@{}c@{}}{Und.{\tiny (16)}} & \multicolumn{1}{@{}c@{}}{Gen.{\tiny (7)}} & \multicolumn{1}{@{}c@{}}{Und.{\tiny (30)}} & \multicolumn{1}{@{}c@{}}{Gen.{\tiny (14)}} & \multicolumn{1}{@{}c@{}}{{\tiny (7)}} & \multicolumn{1}{@{}c@{}}{{\tiny (6)}} & \multicolumn{1}{@{}c@{}}{{\tiny (6)}} & \multicolumn{1}{@{}c@{}}{{\tiny (1)}} & \multicolumn{1}{@{}c@{}}{{\tiny (22)}} & \multicolumn{1}{@{}c@{}}{{\tiny (4)}} & \multicolumn{1}{@{}c@{}}{{\tiny (1)}} & \multicolumn{1}{@{}c@{}}{{\tiny (2)}} & \multicolumn{1}{@{}c@{}}{{\tiny (4)}} \\
\midrule
\multicolumn{17}{@{}l}{\textbf{Wolof-adapted open-weight models (ours)}}\\
Wolof-Qwen2.5-7B (+380 tokens) & \scoretop{\sr{.39}{1.1}} & \textbf{\scoretop{\sr{.49}{1}}} & \textbf{\scoretop{\sr{.50}{1}}} & \textbf{\scoretop{\sr{.36}{1}}} & \textbf{\scoretop{\sr{.48}{1}}} & \textbf{\scoretop{\sr{.28}{1}}} & \textbf{\scoretop{\sr{.27}{1}}} & \underline{\scorelow{\sr{.27}{2}}} & \textbf{\scoretop{\sr{.44}{1}}} & \textbf{\scoretop{\sr{.34}{1}}} & \textbf{\scoretop{\sr{.71}{1}}} & \textbf{\scoretop{\sr{.35}{1}}} & \textbf{\scoretop{\sr{.36}{1}}} & \textbf{\scoretop{\sr{.59}{1}}} & \underline{\scorelow{\sr{.51}{2}}} & \textbf{\scoretop{\sr{.51}{1}}} \\

Wolof-Qwen2.5-7B & \underline{\scorelow{\sr{.35}{1.9}}} & \underline{\scorelow{\sr{.43}{2}}} & \underline{\scorelow{\sr{.46}{2}}} & \underline{\scorelow{\sr{.34}{2}}} & \underline{\scorelow{\sr{.45}{2}}} & \underline{\scorelow{\sr{.26}{2}}} & \underline{\scorelow{\sr{.26}{2}}} & \textbf{\scoretop{\sr{.30}{1}}} & \underline{\scorelow{\sr{.43}{2}}} & \underline{\scorelow{\sr{.29}{2}}} & \underline{\scorelow{\sr{.30}{2}}} & \underline{\scorelow{\sr{.28}{2}}} & \underline{\scorelow{\sr{.25}{2}}} & \underline{\scorelow{\sr{.12}{2}}} & \textbf{\scoretop{\sr{.54}{1}}} & \underline{\scorelow{\sr{.49}{2}}} \\

\bottomrule
\end{tabular}
\end{fitwidthtable}
\caption{Vocabulary-augmentation comparison on Qwen2.5-7B-Instruct: normalised scores (higher, better) and ranks (lower, better). \scorecolorlegend}
\label{tab:development_tokenization_results}
\end{table*}

\clearpage

\section{Stage-by-Stage Results}
\label{app:stage-results}

This appendix isolates the contribution of the successive adaptation stages. Table~\ref{tab:post_training_results} compares the Qwen3 baselines, the models after Wolof-centric SFT, the models after DPO, and the merged models. The SFT stage is the only one that uses Wolof data. DPO and model merging aim to recover general capabilities while preserving the Wolof skills acquired during SFT.

\begin{table*}[ht]
\centering
\scriptsize
\begin{fitwidthtable}
\begin{tabular}{@{}l@{}l@{~}|@{~~}l@{~~}l@{~~}l@{~~}l@{~~}l@{~~}l@{~~}l@{~~}l@{~~}l@{~~}l@{~~}l@{~~}l@{~~}l@{~~}l@{~~}l@{}}
\toprule
 & Avg. sc.$^\uparrow{}$ &\multicolumn{2}{@{}c@{}}{English} &\multicolumn{2}{@{}c@{}}{French} &\multicolumn{2}{@{}c@{}}{Wolof} &\multicolumn{1}{@{}c@{}}{Know.} & \multicolumn{1}{@{}c@{}}{Conv.} & \multicolumn{1}{@{}c@{}}{Reas.} & \multicolumn{1}{@{}c@{}}{Code} & \multicolumn{1}{@{}c@{}}{Format.} & \multicolumn{1}{@{}c@{}}{Instr.} & \multicolumn{1}{@{}c@{}}{Func.} & \multicolumn{1}{@{}c@{}}{Lg ctx.} & \multicolumn{1}{@{}c@{}}{Safet.} \\
\cline{3-8}
& {\tiny (\& rank$^\downarrow{}$)} & \multicolumn{1}{@{}c@{}}{Und.{\tiny (22)}} & \multicolumn{1}{@{}c@{}}{Gen.{\tiny (8)}} & \multicolumn{1}{@{}c@{}}{Und.{\tiny (16)}} & \multicolumn{1}{@{}c@{}}{Gen.{\tiny (7)}} & \multicolumn{1}{@{}c@{}}{Und.{\tiny (30)}} & \multicolumn{1}{@{}c@{}}{Gen.{\tiny (14)}} & \multicolumn{1}{@{}c@{}}{{\tiny (7)}} & \multicolumn{1}{@{}c@{}}{{\tiny (6)}} & \multicolumn{1}{@{}c@{}}{{\tiny (6)}} & \multicolumn{1}{@{}c@{}}{{\tiny (1)}} & \multicolumn{1}{@{}c@{}}{{\tiny (22)}} & \multicolumn{1}{@{}c@{}}{{\tiny (4)}} & \multicolumn{1}{@{}c@{}}{{\tiny (1)}} & \multicolumn{1}{@{}c@{}}{{\tiny (2)}} & \multicolumn{1}{@{}c@{}}{{\tiny (4)}} \\
\midrule
\multicolumn{17}{@{}l}{\textbf{Qwen3-14B family}}\\
Qwen3-14B & \scoretop{\sr{.32}{8.9}} & \scoretop{\sr{.52}{6}} & \scorehigh{\sr{.46}{9}} & \scoretop{\sr{.40}{6}} & \scorehigh{\sr{.45}{11}} & \scoretop{\sr{.10}{15}} & \scoretop{\sr{.00}{18}} & \scorehigh{\sr{.25}{13}} & \scorehigh{\sr{.33}{12}} & \scorehigh{\sr{.37}{11}} & \scoretop{\sr{.74}{3}} & \scorehigh{\sr{.30}{10}} & \textbf{\scoretop{\sr{.67}{1}}} & \textbf{\scoretop{\sr{.70}{1}}} & \scorehigh{\sr{.45}{9}} & \scorehigh{\sr{.54}{9}} \\

+ SFT & \scoretop{\sr{.40}{7.3}} & \scoretop{\sr{.52}{6}} & \scorehigh{\sr{.40}{16}} & \scorehigh{\sr{.35}{12}} & \scorehigh{\sr{.47}{9}} & \scoretop{\sr{.29}{8}} & \scoretop{\sr{.27}{7}} & \scoretop{\sr{.40}{3}} & \scoretop{\sr{.38}{9}} & \textbf{\scoretop{\sr{.46}{1}}} & \scoretop{\sr{.71}{4}} & \scorehigh{\sr{.35}{8}} & \scorehigh{\sr{.41}{12}} & \scorehigh{\sr{.35}{11}} & \textbf{\scoretop{\sr{.51}{1}}} & \scoretop{\sr{.68}{3}} \\

+ SFT + Merging (40/60) & \scoretop{\sr{.47}{2.2}} & \textbf{\scoretop{\sr{.56}{1}}} & \textbf{\scoretop{\sr{.56}{1}}} & \textbf{\scoretop{\sr{.44}{1}}} & \textbf{\scoretop{\sr{.56}{1}}} & \textbf{\scoretop{\sr{.36}{1}}} & \scoretop{\sr{.30}{2}} & \textbf{\scoretop{\sr{.41}{1}}} & \scoretop{\sr{.41}{5}} & \scoretop{\sr{.45}{3}} & \textbf{\scoretop{\sr{.79}{1}}} & \textbf{\scoretop{\sr{.43}{1}}} & \scorehigh{\sr{.57}{5}} & \scorehigh{\sr{.39}{8}} & \textbf{\scoretop{\sr{.51}{1}}} & \textbf{\scoretop{\sr{.73}{1}}} \\

+ SFT + DPO & \scoretop{\sr{.42}{6.0}} & \scoretop{\sr{.53}{4}} & \scoretop{\sr{.51}{5}} & \scoretop{\sr{.40}{6}} & \scoretop{\sr{.50}{6}} & \scoretop{\sr{.31}{5}} & \scoretop{\sr{.28}{4}} & \scoretop{\sr{.39}{4}} & \scoretop{\sr{.43}{3}} & \scoretop{\sr{.44}{4}} & \scorehigh{\sr{.68}{6}} & \scorehigh{\sr{.37}{6}} & \scorehigh{\sr{.47}{11}} & \scorehigh{\sr{.34}{15}} & \scoretop{\sr{.48}{7}} & \scoretop{\sr{.67}{4}} \\

+ SFT + DPO + Merging (final) & \scoretop{\sr{.47}{2.3}} & \textbf{\scoretop{\sr{.56}{1}}} & \textbf{\scoretop{\sr{.56}{1}}} & \textbf{\scoretop{\sr{.44}{1}}} & \scoretop{\sr{.55}{2}} & \textbf{\scoretop{\sr{.36}{1}}} & \textbf{\scoretop{\sr{.31}{1}}} & \textbf{\scoretop{\sr{.41}{1}}} & \scoretop{\sr{.44}{2}} & \textbf{\scoretop{\sr{.46}{1}}} & \textbf{\scoretop{\sr{.79}{1}}} & \textbf{\scoretop{\sr{.43}{1}}} & \scoretop{\sr{.58}{4}} & \scorehigh{\sr{.35}{11}} & \scoretop{\sr{.50}{5}} & \textbf{\scoretop{\sr{.73}{1}}} \\

\midrule
\multicolumn{17}{@{}l}{\textbf{Qwen3-8B family}}\\
Qwen3-8B & \scoretop{\sr{.29}{11.5}} & \scorehigh{\sr{.49}{10}} & \scorehigh{\sr{.46}{9}} & \scorehigh{\sr{.38}{9}} & \scorehigh{\sr{.44}{12}} & \scoretop{\sr{.07}{18}} & \scoretop{\sr{-.03}{20}} & \scoremid{\sr{.20}{16}} & \scorehigh{\sr{.31}{13}} & \scoremid{\sr{.30}{15}} & \scorehigh{\sr{.54}{14}} & \scorehigh{\sr{.26}{12}} & \scoretop{\sr{.63}{2}} & \scoretop{\sr{.67}{3}} & \scorehigh{\sr{.45}{9}} & \scoremid{\sr{.45}{10}} \\

+ SFT & \scoretop{\sr{.40}{7.6}} & \scoretop{\sr{.51}{9}} & \scorehigh{\sr{.47}{8}} & \scorehigh{\sr{.36}{11}} & \scoretop{\sr{.50}{6}} & \scoretop{\sr{.28}{9}} & \scoretop{\sr{.28}{4}} & \scoretop{\sr{.36}{6}} & \scoretop{\sr{.41}{5}} & \scorehigh{\sr{.41}{7}} & \scorehigh{\sr{.66}{7}} & \scorehigh{\sr{.33}{9}} & \scorehigh{\sr{.40}{13}} & \scorehigh{\sr{.35}{11}} & \textbf{\scoretop{\sr{.51}{1}}} & \scorehigh{\sr{.62}{8}} \\

+ SFT + Merging (40/60) & \scoretop{\sr{.43}{5.4}} & \scoretop{\sr{.53}{4}} & \scoretop{\sr{.52}{4}} & \scoretop{\sr{.42}{3}} & \scoretop{\sr{.51}{4}} & \scoretop{\sr{.31}{5}} & \scoretop{\sr{.27}{7}} & \scorehigh{\sr{.34}{8}} & \scoretop{\sr{.40}{7}} & \scorehigh{\sr{.41}{7}} & \scorehigh{\sr{.63}{9}} & \scoretop{\sr{.40}{3}} & \scorehigh{\sr{.54}{7}} & \scorehigh{\sr{.50}{6}} & \textbf{\scoretop{\sr{.51}{1}}} & \scorehigh{\sr{.63}{6}} \\

+ SFT + DPO & \scoretop{\sr{.43}{5.9}} & \scoretop{\sr{.52}{6}} & \scoretop{\sr{.51}{5}} & \scoretop{\sr{.41}{5}} & \scoretop{\sr{.51}{4}} & \scoretop{\sr{.32}{4}} & \scoretop{\sr{.29}{3}} & \scoretop{\sr{.37}{5}} & \textbf{\scoretop{\sr{.45}{1}}} & \scoretop{\sr{.43}{6}} & \scorehigh{\sr{.60}{11}} & \scoretop{\sr{.38}{5}} & \scorehigh{\sr{.52}{8}} & \scorehigh{\sr{.35}{11}} & \scorehigh{\sr{.45}{9}} & \scorehigh{\sr{.63}{6}} \\

+ SFT + DPO + Merging (final) & \scoretop{\sr{.44}{4.4}} & \scoretop{\sr{.54}{3}} & \scoretop{\sr{.53}{3}} & \scoretop{\sr{.42}{3}} & \scoretop{\sr{.53}{3}} & \scoretop{\sr{.33}{3}} & \scoretop{\sr{.28}{4}} & \scoretop{\sr{.36}{6}} & \scoretop{\sr{.42}{4}} & \scoretop{\sr{.44}{4}} & \scoretop{\sr{.69}{5}} & \scoretop{\sr{.40}{3}} & \scorehigh{\sr{.57}{5}} & \scorehigh{\sr{.46}{7}} & \scoretop{\sr{.47}{8}} & \scorehigh{\sr{.64}{5}} \\

\midrule
\multicolumn{17}{@{}l}{\textbf{Qwen3-4B family}}\\
Qwen3-4B & \scoretop{\sr{.28}{12.5}} & \scorehigh{\sr{.46}{12}} & \scorehigh{\sr{.45}{12}} & \scorehigh{\sr{.37}{10}} & \scorehigh{\sr{.44}{12}} & \scoretop{\sr{.07}{18}} & \scoretop{\sr{-.02}{19}} & \scorehigh{\sr{.24}{14}} & \scorehigh{\sr{.26}{18}} & \scorehigh{\sr{.33}{12}} & \scorehigh{\sr{.62}{10}} & \scorehigh{\sr{.25}{15}} & \scoretop{\sr{.59}{3}} & \scoretop{\sr{.68}{2}} & \scorehigh{\sr{.42}{17}} & \scoremid{\sr{.38}{13}} \\

+ SFT & \scoretop{\sr{.35}{11.2}} & \scorehigh{\sr{.45}{13}} & \scorehigh{\sr{.37}{17}} & \scorehigh{\sr{.33}{13}} & \scorehigh{\sr{.47}{9}} & \scoretop{\sr{.25}{10}} & \scoretop{\sr{.25}{10}} & \scorehigh{\sr{.32}{9}} & \scorehigh{\sr{.36}{10}} & \scorehigh{\sr{.39}{9}} & \scorehigh{\sr{.55}{12}} & \scorehigh{\sr{.30}{10}} & \scoremid{\sr{.32}{15}} & \scorehigh{\sr{.33}{16}} & \scoretop{\sr{.50}{5}} & \scoremid{\sr{.45}{10}} \\

+ SFT + Merging (40/60) & \scoremid{\sr{-1.53}{19.1}} & \scoremid{\sr{.25}{22}} & \scorelow{\sr{.10}{23}} & \scoremid{\sr{.17}{22}} & \scorelow{\sr{.15}{22}} & \scoremid{\sr{-3.49}{23}} & \scoremid{\sr{-7.74}{23}} & \scorehigh{\sr{.26}{11}} & \scoremid{\sr{.05}{22}} & \scorehigh{\sr{.32}{13}} & \scoremid{\sr{.29}{20}} & \scorehigh{\sr{.26}{12}} & \scorelow{\sr{.14}{21}} & \underline{\scorelow{\sr{-.45}{22}}} & \scorehigh{\sr{.43}{16}} & \scoremid{\sr{.35}{15}} \\

+ SFT + DPO (final) & \scoretop{\sr{.40}{8.7}} & \scorehigh{\sr{.47}{11}} & \scorehigh{\sr{.48}{7}} & \scoretop{\sr{.39}{8}} & \scorehigh{\sr{.48}{8}} & \scoretop{\sr{.30}{7}} & \scoretop{\sr{.27}{7}} & \scorehigh{\sr{.30}{10}} & \scoretop{\sr{.39}{8}} & \scorehigh{\sr{.38}{10}} & \scorehigh{\sr{.64}{8}} & \scorehigh{\sr{.36}{7}} & \scorehigh{\sr{.49}{10}} & \scorehigh{\sr{.37}{10}} & \scorehigh{\sr{.45}{9}} & \scoremid{\sr{.45}{10}} \\

+ SFT + DPO + Merging & \scoremid{\sr{-1.52}{18.8}} & \scoremid{\sr{.26}{20}} & \scorelow{\sr{.11}{22}} & \scoremid{\sr{.17}{22}} & \scorelow{\sr{.15}{22}} & \scoremid{\sr{-3.47}{22}} & \scoremid{\sr{-7.69}{22}} & \scorehigh{\sr{.26}{11}} & \scoremid{\sr{.05}{22}} & \scorehigh{\sr{.31}{14}} & \scorelow{\sr{.20}{22}} & \scorehigh{\sr{.26}{12}} & \scoremid{\sr{.16}{20}} & \underline{\scorelow{\sr{-.45}{22}}} & \scorehigh{\sr{.44}{15}} & \scoremid{\sr{.37}{14}} \\

\midrule
\multicolumn{17}{@{}l}{\textbf{Qwen3-1.7B family}}\\
Qwen3-1.7B & \scoretop{\sr{.23}{16.4}} & \scorehigh{\sr{.40}{14}} & \scorehigh{\sr{.41}{14}} & \scorehigh{\sr{.32}{14}} & \scorehigh{\sr{.39}{17}} & \scoretop{\sr{.04}{21}} & \scoretop{\sr{-.04}{21}} & \scorelow{\sr{.10}{21}} & \scorehigh{\sr{.23}{20}} & \scoremid{\sr{.26}{20}} & \scorehigh{\sr{.53}{15}} & \scoremid{\sr{.20}{19}} & \scorehigh{\sr{.50}{9}} & \scoretop{\sr{.61}{4}} & \scorehigh{\sr{.37}{20}} & \scoremid{\sr{.32}{17}} \\

+ SFT & \scoretop{\sr{.28}{15.6}} & \scorehigh{\sr{.36}{17}} & \scorehigh{\sr{.41}{14}} & \scorehigh{\sr{.29}{16}} & \scorehigh{\sr{.41}{16}} & \scoretop{\sr{.19}{12}} & \scoretop{\sr{.22}{11}} & \scoremid{\sr{.22}{15}} & \scorehigh{\sr{.30}{14}} & \scoremid{\sr{.27}{19}} & \scoremid{\sr{.32}{19}} & \scoremid{\sr{.21}{18}} & \scoremid{\sr{.20}{19}} & \scorehigh{\sr{.27}{17}} & \scorehigh{\sr{.45}{9}} & \scoremid{\sr{.31}{18}} \\

+ SFT + Merging (40/60) & \scoretop{\sr{.27}{14.3}} & \scorehigh{\sr{.38}{15}} & \scorehigh{\sr{.45}{12}} & \scorehigh{\sr{.29}{16}} & \scorehigh{\sr{.42}{15}} & \scoretop{\sr{.15}{13}} & \scoretop{\sr{.10}{15}} & \scoremid{\sr{.18}{18}} & \scorehigh{\sr{.28}{16}} & \scoremid{\sr{.30}{15}} & \scorehigh{\sr{.55}{12}} & \scoremid{\sr{.22}{17}} & \scoremid{\sr{.28}{16}} & \scorehigh{\sr{.38}{9}} & \scorehigh{\sr{.45}{9}} & \scoremid{\sr{.33}{16}} \\

+ SFT + DPO (final) & \scoretop{\sr{.31}{14.9}} & \scorehigh{\sr{.37}{16}} & \scorehigh{\sr{.46}{9}} & \scorehigh{\sr{.31}{15}} & \scorehigh{\sr{.43}{14}} & \scoretop{\sr{.22}{11}} & \scoretop{\sr{.22}{11}} & \scoremid{\sr{.20}{16}} & \scorehigh{\sr{.36}{10}} & \scoremid{\sr{.29}{17}} & \scorehigh{\sr{.48}{16}} & \scoremid{\sr{.24}{16}} & \scoremid{\sr{.25}{17}} & \scorehigh{\sr{.24}{19}} & \scorehigh{\sr{.42}{17}} & \scoremid{\sr{.30}{19}} \\

+ SFT + DPO + Merging & \scoremid{\sr{-1.68}{22.0}} & \scorelow{\sr{.18}{24}} & \underline{\scorelow{\sr{.06}{25}}} & \scorelow{\sr{.11}{24}} & \scorelow{\sr{.11}{24}} & \scoremid{\sr{-3.66}{24}} & \scoremid{\sr{-7.94}{24}} & \scoremid{\sr{.18}{18}} & \underline{\scorelow{\sr{-.04}{25}}} & \scoremid{\sr{.29}{17}} & \scoremid{\sr{.35}{18}} & \scoremid{\sr{.13}{21}} & \underline{\scorelow{\sr{.06}{24}}} & \underline{\scorelow{\sr{-.45}{22}}} & \scorehigh{\sr{.41}{19}} & \scorelow{\sr{.25}{21}} \\

\midrule
\multicolumn{17}{@{}l}{\textbf{Qwen3-0.6B family}}\\
Qwen3-0.6B & \scoretop{\sr{.20}{18.2}} & \scoremid{\sr{.33}{18}} & \scorehigh{\sr{.37}{17}} & \scorehigh{\sr{.27}{18}} & \scoremid{\sr{.32}{20}} & \scoretop{\sr{.06}{20}} & \scoretop{\sr{.03}{17}} & \scorelow{\sr{.07}{23}} & \scorehigh{\sr{.25}{19}} & \scoremid{\sr{.21}{21}} & \scoremid{\sr{.37}{17}} & \scoremid{\sr{.15}{20}} & \scorehigh{\sr{.38}{14}} & \scoretop{\sr{.53}{5}} & \underline{\scorelow{\sr{.18}{25}}} & \scoremid{\sr{.30}{19}} \\

+ SFT (final) & \scoretop{\sr{.20}{19.6}} & \scoremid{\sr{.26}{20}} & \scorehigh{\sr{.35}{20}} & \scoremid{\sr{.24}{19}} & \scorehigh{\sr{.33}{18}} & \scoretop{\sr{.12}{14}} & \scoretop{\sr{.17}{13}} & \scoremid{\sr{.14}{20}} & \scorehigh{\sr{.27}{17}} & \scoremid{\sr{.20}{22}} & \scorelow{\sr{.20}{22}} & \scoremid{\sr{.12}{22}} & \scorelow{\sr{.13}{22}} & \scorehigh{\sr{.23}{20}} & \scoremid{\sr{.26}{21}} & \scorelow{\sr{.22}{24}} \\

+ SFT + Merging (40/60) & \scoretop{\sr{.20}{18.8}} & \scoremid{\sr{.28}{19}} & \scorehigh{\sr{.37}{17}} & \scoremid{\sr{.24}{19}} & \scorehigh{\sr{.33}{18}} & \scoretop{\sr{.10}{15}} & \scoretop{\sr{.11}{14}} & \scorelow{\sr{.09}{22}} & \scorehigh{\sr{.29}{15}} & \scoremid{\sr{.20}{22}} & \scoremid{\sr{.29}{20}} & \scoremid{\sr{.12}{22}} & \scoremid{\sr{.21}{18}} & \scorehigh{\sr{.27}{17}} & \scoremid{\sr{.25}{23}} & \scorelow{\sr{.25}{21}} \\

+ SFT + DPO & \scoretop{\sr{.17}{21.7}} & \scoremid{\sr{.24}{23}} & \scorehigh{\sr{.34}{21}} & \scoremid{\sr{.21}{21}} & \scoremid{\sr{.29}{21}} & \scoretop{\sr{.09}{17}} & \scoretop{\sr{.10}{15}} & \underline{\scorelow{\sr{.06}{25}}} & \scorehigh{\sr{.21}{21}} & \scorelow{\sr{.17}{24}} & \scorelow{\sr{.20}{22}} & \scorelow{\sr{.11}{24}} & \scorelow{\sr{.12}{23}} & \scorehigh{\sr{.20}{21}} & \scoremid{\sr{.24}{24}} & \scorelow{\sr{.24}{23}} \\

+ SFT + DPO + Merging & \underline{\scorelow{\sr{-2.24}{24.1}}} & \underline{\scorelow{\sr{.13}{25}}} & \scorelow{\sr{.08}{24}} & \underline{\scorelow{\sr{.08}{25}}} & \underline{\scorelow{\sr{.09}{25}}} & \underline{\scorelow{\sr{-4.84}{25}}} & \underline{\scorelow{\sr{-10.41}{25}}} & \scorelow{\sr{.07}{23}} & \scorelow{\sr{-.01}{24}} & \underline{\scorelow{\sr{.14}{25}}} & \underline{\scorelow{\sr{.11}{25}}} & \underline{\scorelow{\sr{.06}{25}}} & \scorelow{\sr{.06}{24}} & \underline{\scorelow{\sr{-.45}{22}}} & \scoremid{\sr{.26}{21}} & \underline{\scorelow{\sr{.22}{24}}} \\

\bottomrule
\end{tabular}
\end{fitwidthtable}
\caption{Post-SFT comparison of Qwen3 baselines, Wolof-adapted Qwen3 models, and DPO/model-merging variants. DPO uses the DOLCI preference dataset; merging denotes a 40/60 interpolation between the unadapted and adapted checkpoints: normalised scores (higher, better) and ranks (lower, better). \scorecolorlegend}
\label{tab:post_training_results}
\end{table*}

The recovery mechanism changes with model capacity. At 14B, merging contributes substantially more after SFT than DPO, and adding DPO before merging leaves the overall score unchanged while slightly reshaping the skill profile. At 8B, DPO and merging independently reach the same overall score and their combination adds a smaller further gain, indicating complementarity. DPO is instead the strongest post-SFT option at 4B and 1.7B, where merging becomes unstable; at 0.6B, neither stage improves on SFT. This transition motivates the size-specific final recipes reported in the main table.

\clearpage

\section{Detailed Final Results}
\label{app:skill-detailed-results}

This appendix reports every task for the models in Table~\ref{tab:benchmark_results_abbrev_noprovider}, grouped by skill. The adapted 4B and 8B outperform their respective bases on all 24 public Wolof tasks, and the 14B on 23, showing that the aggregate Wolof gains are not driven by a small subset of benchmarks.

\newcommand{\skilldetailstable}[1]{%
  \FloatBarrier
  \input{#1}
  \FloatBarrier
}

  \FloatBarrier
  \begin{sidewaystable*}[p]
\centering
\scriptsize
\begin{fitwidthtable}
\begin{tabular}{@{}lr@{~~}r@{~~}c@{~~}c@{~~}c@{~~}c@{~~}c@{~~}c@{~~}c@{~~}c@{~~}c@{~~}c@{~~}c@{~~}c@{~~}c@{~~}c@{~~}c@{~~}c@{~~}c@{~~}c@{~~}c@{~~}c@{~~}c@{~~}c@{~~}c@{~~}c@{~~}c@{~~}c@{~~}c@{~~}c@{~~}c@{~~}c@{}}
\toprule
Model & Avg. score & Avg. rank & \rotatebox[origin=l]{90}{AfrIFact / Claim Classification} & \rotatebox[origin=l]{90}{AfrIFact / Evidence Extraction} & \rotatebox[origin=l]{90}{Aya Wolof / Culture QA} & \rotatebox[origin=l]{90}{Aya Wolof / Orthographic Correction} & \rotatebox[origin=l]{90}{Aya Wolof / POS Tagging} & \rotatebox[origin=l]{90}{Aya Wolof / Translation / FROM Wolof} & \rotatebox[origin=l]{90}{Claire Wolof / Conversation Coherence} & \rotatebox[origin=l]{90}{Corporate FAQ Wolof / Conversational MRQA} & \rotatebox[origin=l]{90}{Corporate FAQ Wolof / Long Context MRQA} & \rotatebox[origin=l]{90}{Corporate FAQ Wolof / MRQA} & \rotatebox[origin=l]{90}{FLORES / Translation / Wolof → English} & \rotatebox[origin=l]{90}{FLORES / Translation / Wolof → French} & \rotatebox[origin=l]{90}{IrokoBench / MGSM} & \rotatebox[origin=l]{90}{IrokoBench / MMLU} & \rotatebox[origin=l]{90}{IrokoBench / XNLI} & \rotatebox[origin=l]{90}{MasakhaNER / NER (News Domain)} & \rotatebox[origin=l]{90}{Corporate / Conversation Coherence} & \rotatebox[origin=l]{90}{Corporate / Dialog Act Labelling} & \rotatebox[origin=l]{90}{Corporate / Full Analysis} & \rotatebox[origin=l]{90}{Corporate / GDPR/PII Extraction} & \rotatebox[origin=l]{90}{Corporate / NER (Public Entities)} & \rotatebox[origin=l]{90}{Corporate / Resolution Classification} & \rotatebox[origin=l]{90}{Corporate / Sentiment Classification} & \rotatebox[origin=l]{90}{Corporate / Summarization} & \rotatebox[origin=l]{90}{WolBanking77 / Intent Classification / Wolof} & \rotatebox[origin=l]{90}{WolBanking77 / Translation / Wolof → English} & \rotatebox[origin=l]{90}{WolBanking77 / Translation / Wolof → French} & \rotatebox[origin=l]{90}{WolBanking77 Audio / Intent Classification / ASR} & \rotatebox[origin=l]{90}{WolBanking77 Audio / Intent Classification / Reference} & \rotatebox[origin=l]{90}{WolBanking77 Audio / Transcription Correction} \\
\bottomrule
\end{tabular}
\end{fitwidthtable}
\caption{Detailed Wolof understanding results: normalized scores (higher, better) and ranks (lower, better). \scorecolorlegend}
\label{tab:detailed_wolof_understanding_results}
\end{sidewaystable*}

  \FloatBarrier

  \FloatBarrier
  \begin{table*}[ht]
\centering
\scriptsize
\begin{fitwidthtable}
\begin{tabular}{@{}lr@{~~}r@{~~}c@{~~}c@{~~}c@{~~}c@{~~}c@{~~}c@{~~}c@{~~}c@{~~}c@{~~}c@{~~}c@{~~}c@{~~}c@{~~}c@{}}
\toprule
Model & Avg. score & Avg. rank & \rotatebox[origin=l]{90}{Aya Wolof / Culture QA} & \rotatebox[origin=l]{90}{Aya Wolof / Orthographic Correction} & \rotatebox[origin=l]{90}{Aya Wolof / Translation / TO Wolof} & \rotatebox[origin=l]{90}{Claire Wolof / Conversation Coherence} & \rotatebox[origin=l]{90}{Corporate FAQ Wolof / Conversational MRQA} & \rotatebox[origin=l]{90}{Corporate FAQ Wolof / MRQA} & \rotatebox[origin=l]{90}{FLORES / Translation / English → Wolof} & \rotatebox[origin=l]{90}{FLORES / Translation / French → Wolof} & \rotatebox[origin=l]{90}{Corporate / Conversation Coherence} & \rotatebox[origin=l]{90}{Corporate / Full Analysis} & \rotatebox[origin=l]{90}{Corporate / Summarization} & \rotatebox[origin=l]{90}{WolBanking77 / Translation / English → Wolof} & \rotatebox[origin=l]{90}{WolBanking77 / Translation / French → Wolof} & \rotatebox[origin=l]{90}{WolBanking77 Audio / Transcription Correction} \\
\midrule
\multicolumn{17}{@{}l}{\textbf{Baseline open-weight models}}\\
Qwen3-14B & \scorelow{.00} & 14.9 & \scorelow{\sr{-.47}{15}} & \scorelow{\sr{.05}{16}} & \scoremid{\sr{.03}{12}} & \scoremid{\sr{-.60}{13}} & \scoremid{\sr{.13}{15}} & \scorelow{\sr{.11}{17}} & \scorelow{\sr{.06}{16}} & \scorelow{\sr{.05}{17}} & \scorelow{\sr{.15}{19}} & \scoremid{\sr{.11}{12}} & \scoremid{\sr{-.33}{16}} & \scorelow{\sr{-.00}{15}} & \scorelow{\sr{-.01}{15}} & \scoremid{\sr{.76}{11}} \\

Qwen3-8B & \scorelow{-.03} & 16.1 & \scorelow{\sr{-.56}{17}} & \scorelow{\sr{.03}{19}} & \scorelow{\sr{.02}{17}} & \scoremid{\sr{-.76}{16}} & \scoremid{\sr{.16}{12}} & \scoremid{\sr{.12}{15}} & \scorelow{\sr{.06}{16}} & \scorelow{\sr{.06}{15}} & \scorelow{\sr{.16}{17}} & \scoremid{\sr{.09}{13}} & \scorelow{\sr{-.49}{18}} & \scorelow{\sr{-.01}{16}} & \scorelow{\sr{-.01}{15}} & \scorelow{\sr{.68}{19}} \\
\bottomrule
\end{tabular}
\end{fitwidthtable}
\caption{Detailed Wolof generation results: normalized scores (higher, better) and ranks (lower, better). \scorecolorlegend}
\label{tab:detailed_wolof_generation_results}
\end{table*}

  \FloatBarrier

  \FloatBarrier
  \begin{sidewaystable*}[p]
\centering
\scriptsize
\begin{fitwidthtable}
\begin{tabular}{@{}lr@{~~}r@{~~}c@{~~}c@{~~}c@{~~}c@{~~}c@{~~}c@{~~}c@{~~}c@{~~}c@{~~}c@{~~}c@{~~}c@{~~}c@{~~}c@{~~}c@{~~}c@{}}
\toprule
Model & Avg. score & Avg. rank & \rotatebox[origin=l]{90}{Aya Wolof / Translation / TO Wolof} & \rotatebox[origin=l]{90}{Corporate FAQ French / Conversational MRQA} & \rotatebox[origin=l]{90}{Corporate FAQ French / MRQA} & \rotatebox[origin=l]{90}{FLORES / Translation / French → English} & \rotatebox[origin=l]{90}{FLORES / Translation / French → Wolof} & \rotatebox[origin=l]{90}{GPQA Diamond / MCQA / French} & \rotatebox[origin=l]{90}{MIFEval / Instruction Following / French} & \rotatebox[origin=l]{90}{MMLU / MCQA / French} & \rotatebox[origin=l]{90}{Multi-IF / Instruction Following} & \rotatebox[origin=l]{90}{Corporate / Conversation Coherence} & \rotatebox[origin=l]{90}{Corporate / Full Analysis} & \rotatebox[origin=l]{90}{Corporate / Resolution Classification} & \rotatebox[origin=l]{90}{Corporate / Sentiment Classification} & \rotatebox[origin=l]{90}{WolBanking77 / Intent Classification / French} & \rotatebox[origin=l]{90}{WolBanking77 / Translation / French → English} & \rotatebox[origin=l]{90}{WolBanking77 / Translation / French → Wolof} \\
\midrule
\multicolumn{19}{@{}l}{\textbf{Baseline open-weight models}}\\
Qwen3-14B & \scorehigh{.40} & 9.7 & \scoremid{\sr{.03}{12}} & \scoretop{\sr{.72}{4}} & \scorehigh{\sr{.66}{6}} & \scoretop{\sr{.74}{6}} & \scorelow{\sr{.05}{17}} & \scoremid{\sr{.14}{9}} & \scorehigh{\sr{.75}{7}} & \scorehigh{\sr{.71}{7}} & \scorehigh{\sr{.80}{6}} & \scorelow{\sr{.15}{19}} & \scoremid{\sr{.11}{12}} & \scoremid{\sr{.18}{8}} & \scoremid{\sr{.02}{11}} & \scorehigh{\sr{.55}{7}} & \scorehigh{\sr{.75}{9}} & \scorelow{\sr{-.01}{15}} \\
\bottomrule
\end{tabular}
\end{fitwidthtable}
\caption{Detailed French understanding results: normalized scores (higher, better) and ranks (lower, better). \scorecolorlegend}
\label{tab:detailed_french_understanding_results}
\end{sidewaystable*}

  \FloatBarrier

  \FloatBarrier
  \begin{table*}[ht]
\centering
\scriptsize
\begin{fitwidthtable}
\begin{tabular}{@{}lr@{~~}r@{~~}c@{~~}c@{~~}c@{~~}c@{~~}c@{~~}c@{~~}c@{}}
\toprule
Model & Avg. score & Avg. rank & \rotatebox[origin=l]{90}{Aya Wolof / Translation / FROM Wolof} & \rotatebox[origin=l]{90}{Corporate FAQ French / Conversational MRQA} & \rotatebox[origin=l]{90}{Corporate FAQ French / MRQA} & \rotatebox[origin=l]{90}{FLORES / Translation / English → French} & \rotatebox[origin=l]{90}{FLORES / Translation / Wolof → French} & \rotatebox[origin=l]{90}{WolBanking77 / Translation / English → French} & \rotatebox[origin=l]{90}{WolBanking77 / Translation / Wolof → French} \\
\midrule
\multicolumn{10}{@{}l}{\textbf{Baseline open-weight models}}\\
Qwen3-14B & \scoremid{.45} & 9.9 & \scorelow{\sr{.03}{17}} & \scoretop{\sr{.72}{4}} & \scorehigh{\sr{.66}{6}} & \scorehigh{\sr{.69}{7}} & \scoremid{\sr{.18}{14}} & \scorehigh{\sr{.76}{6}} & \scoremid{\sr{.14}{15}} \\

Qwen3-8B & \scoremid{.44} & 11.9 & \scorelow{\sr{.04}{16}} & \scorehigh{\sr{.70}{8}} & \scorehigh{\sr{.65}{9}} & \scorehigh{\sr{.67}{9}} & \scoremid{\sr{.14}{16}} & \scorehigh{\sr{.74}{10}} & \scoremid{\sr{.14}{15}} \\

Qwen3-4B & \scoremid{.44} & 12.6 & \scorelow{\sr{.07}{12}} & \scorehigh{\sr{.69}{10}} & \scorehigh{\sr{.65}{9}} & \scorehigh{\sr{.66}{13}} & \scoremid{\sr{.15}{15}} & \scorehigh{\sr{.71}{12}} & \scorelow{\sr{.13}{17}} \\

Qwen3-1.7B & \scoremid{.39} & 16.0 & \scorelow{\sr{.03}{17}} & \scorehigh{\sr{.63}{13}} & \scorehigh{\sr{.59}{14}} & \scorehigh{\sr{.60}{16}} & \scorelow{\sr{.11}{18}} & \scorehigh{\sr{.64}{16}} & \scorelow{\sr{.10}{18}} \\

Qwen3-0.6B & \scorelow{.32} & 17.4 & \scorelow{\sr{.03}{17}} & \scoremid{\sr{.58}{16}} & \scoremid{\sr{.57}{15}} & \scorelow{\sr{.47}{18}} & \scorelow{\sr{.07}{19}} & \scorelow{\sr{.48}{18}} & \scorelow{\sr{.06}{19}} \\

\midrule
\multicolumn{10}{@{}l}{\textbf{Commercial API models}}\\
\bottomrule
\end{tabular}
\end{fitwidthtable}
\caption{Detailed French generation results: normalized scores (higher, better) and ranks (lower, better). \scorecolorlegend}
\label{tab:detailed_french_generation_results}
\end{table*}

  \FloatBarrier

  \FloatBarrier
  \begin{sidewaystable*}[p]
\centering
\scriptsize
\begin{fitwidthtable}
\begin{tabular}{@{}lr@{~~}r@{~~}c@{~~}c@{~~}c@{~~}c@{~~}c@{~~}c@{~~}c@{~~}c@{~~}c@{~~}c@{~~}c@{~~}c@{~~}c@{~~}c@{~~}c@{~~}c@{~~}c@{~~}c@{~~}c@{~~}c@{~~}c@{~~}c@{}}
\toprule
Model & Avg. score & Avg. rank & \rotatebox[origin=l]{90}{ARC-Challenge / Scientific Reasoning} & \rotatebox[origin=l]{90}{Aya Wolof / Translation / TO Wolof} & \rotatebox[origin=l]{90}{BBQ / Demographic Bias MCQA} & \rotatebox[origin=l]{90}{BFCL / Function Calling} & \rotatebox[origin=l]{90}{DialogSum / Conversation Summarization} & \rotatebox[origin=l]{90}{FLORES / Translation / English → French} & \rotatebox[origin=l]{90}{FLORES / Translation / English → Wolof} & \rotatebox[origin=l]{90}{GPQA Diamond / MCQA/ English} & \rotatebox[origin=l]{90}{GSM8K / Mathematical Reasoning} & \rotatebox[origin=l]{90}{HaluEval / QA Hallucination Detection} & \rotatebox[origin=l]{90}{IFBench / Instruction Following} & \rotatebox[origin=l]{90}{MIFEval / Instruction Following / English} & \rotatebox[origin=l]{90}{MMLU / MCQA / English} & \rotatebox[origin=l]{90}{Multi-IF / Instruction Following} & \rotatebox[origin=l]{90}{PersonaChat / Conversation Coherence} & \rotatebox[origin=l]{90}{RULER / MRQA} & \rotatebox[origin=l]{90}{ToxiGen / Implicit Hate Detection} & \rotatebox[origin=l]{90}{WebNLG QA / Conversational MRQA} & \rotatebox[origin=l]{90}{WildJailbreak / Harmful Request Refusal} & \rotatebox[origin=l]{90}{WolBanking77 / Intent Classification / English} & \rotatebox[origin=l]{90}{WolBanking77 / Translation / English → French} & \rotatebox[origin=l]{90}{WolBanking77 / Translation / English → Wolof} \\
\midrule
\multicolumn{25}{@{}l}{\textbf{Baseline open-weight models}}\\
\bottomrule
\end{tabular}
\end{fitwidthtable}
\caption{Detailed English understanding results: normalized scores (higher, better) and ranks (lower, better). \scorecolorlegend}
\label{tab:detailed_english_understanding_results}
\end{sidewaystable*}

  \FloatBarrier

  \FloatBarrier
  \begin{table*}[ht]
\centering
\scriptsize
\begin{fitwidthtable}
\begin{tabular}{@{}lr@{~~}r@{~~}c@{~~}c@{~~}c@{~~}c@{~~}c@{~~}c@{~~}c@{~~}c@{}}
\toprule
Model & Avg. score & Avg. rank & \rotatebox[origin=l]{90}{Aya Wolof / Translation / FROM Wolof} & \rotatebox[origin=l]{90}{DialogSum / Conversation Summarization} & \rotatebox[origin=l]{90}{FLORES / Translation / French → English} & \rotatebox[origin=l]{90}{FLORES / Translation / Wolof → English} & \rotatebox[origin=l]{90}{PersonaChat / Conversation Coherence} & \rotatebox[origin=l]{90}{WebNLG QA / Conversational MRQA} & \rotatebox[origin=l]{90}{WolBanking77 / Translation / French → English} & \rotatebox[origin=l]{90}{WolBanking77 / Translation / Wolof → English} \\
\midrule
\multicolumn{11}{@{}l}{\textbf{Baseline open-weight models}}\\
Qwen3-14B & \scoremid{.46} & 11.2 & \scorelow{\sr{.03}{17}} & \scoremid{\sr{.22}{11}} & \scoretop{\sr{.74}{6}} & \scoremid{\sr{.24}{14}} & \scoretop{\sr{.92}{7}} & \scorehigh{\sr{.69}{11}} & \scorehigh{\sr{.75}{9}} & \scoremid{\sr{.12}{15}} \\

Qwen3-8B & \scoremid{.46} & 11.8 & \scorelow{\sr{.04}{16}} & \scorelow{\sr{.21}{16}} & \scoretop{\sr{.73}{8}} & \scoremid{\sr{.21}{15}} & \scoretop{\sr{.90}{8}} & \scoretop{\sr{.71}{4}} & \scorehigh{\sr{.74}{11}} & \scorelow{\sr{.11}{16}} \\

Qwen3-4B & \scoremid{.45} & 11.9 & \scorelow{\sr{.07}{12}} & \scoremid{\sr{.22}{11}} & \scoretop{\sr{.73}{8}} & \scoremid{\sr{.20}{16}} & \scorehigh{\sr{.88}{9}} & \scorehigh{\sr{.69}{11}} & \scorehigh{\sr{.74}{11}} & \scorelow{\sr{.10}{17}} \\

Qwen3-1.7B & \scoremid{.41} & 14.2 & \scorelow{\sr{.03}{17}} & \scoremid{\sr{.23}{9}} & \scorehigh{\sr{.71}{11}} & \scorelow{\sr{.17}{17}} & \scorehigh{\sr{.80}{11}} & \scorehigh{\sr{.62}{15}} & \scoremid{\sr{.70}{16}} & \scorelow{\sr{.08}{18}} \\

Qwen3-0.6B & \scorelow{.37} & 16.8 & \scorelow{\sr{.03}{17}} & \scoremid{\sr{.22}{11}} & \scorehigh{\sr{.64}{17}} & \scorelow{\sr{.14}{18}} & \scoremid{\sr{.71}{16}} & \scoremid{\sr{.48}{18}} & \scoremid{\sr{.66}{18}} & \scorelow{\sr{.04}{19}} \\
\bottomrule
\end{tabular}
\end{fitwidthtable}
\caption{Detailed English generation results: normalized scores (higher, better) and ranks (lower, better). \scorecolorlegend}
\label{tab:detailed_english_generation_results}
\end{table*}

  \FloatBarrier

  \FloatBarrier
  \begin{table*}[ht]
\centering
\scriptsize
\begin{fitwidthtable}
\begin{tabular}{@{}lr@{~~}r@{~~}c@{~~}c@{~~}c@{~~}c@{~~}c@{~~}c@{~~}c@{}}
\toprule
Model & Avg. score & Avg. rank & \rotatebox[origin=l]{90}{ARC-Challenge / Scientific Reasoning} & \rotatebox[origin=l]{90}{Aya Wolof / Culture QA} & \rotatebox[origin=l]{90}{GPQA Diamond / MCQA / French} & \rotatebox[origin=l]{90}{GPQA Diamond / MCQA/ English} & \rotatebox[origin=l]{90}{IrokoBench / MMLU} & \rotatebox[origin=l]{90}{MMLU / MCQA / English} & \rotatebox[origin=l]{90}{MMLU / MCQA / French} \\
\midrule
\multicolumn{10}{@{}l}{\textbf{Baseline open-weight models}}\\
Qwen3-14B & \scoremid{.25} & 9.3 & \scoremid{\sr{.12}{12}} & \scorelow{\sr{-.47}{15}} & \scoremid{\sr{.14}{9}} & \scoremid{\sr{.32}{11}} & \scoremid{\sr{.18}{7}} & \scorehigh{\sr{.72}{4}} & \scorehigh{\sr{.71}{7}} \\

Qwen3-8B & \scoremid{.20} & 12.0 & \scoremid{\sr{.15}{11}} & \scorelow{\sr{-.56}{17}} & \scorelow{\sr{.01}{15}} & \scorelow{\sr{.29}{14}} & \scorelow{\sr{.05}{16}} & \scorehigh{\sr{.70}{7}} & \scorehigh{\sr{.74}{4}} \\

Qwen3-4B & \scoremid{.24} & 12.3 & \scorelow{\sr{.03}{17}} & \scoremid{\sr{-.11}{12}} & \scoremid{\sr{.12}{10}} & \scorelow{\sr{.30}{13}} & \scoremid{\sr{.06}{14}} & \scorehigh{\sr{.64}{9}} & \scorehigh{\sr{.66}{11}} \\

Qwen3-1.7B & \scorelow{.10} & 16.1 & \scorelow{\sr{.05}{13}} & \scorelow{\sr{-.64}{19}} & \scorelow{\sr{-.01}{16}} & \scorelow{\sr{.26}{18}} & \scorelow{\sr{-.04}{19}} & \scoremid{\sr{.53}{14}} & \scoremid{\sr{.58}{14}} \\

Qwen3-0.6B & \scorelow{.07} & 17.9 & \scorelow{\sr{.02}{18}} & \scorelow{\sr{-.61}{18}} & \scorelow{\sr{-.03}{18}} & \scorelow{\sr{.23}{19}} & \scorelow{\sr{.03}{18}} & \scorelow{\sr{.35}{17}} & \scoremid{\sr{.48}{17}} \\

\midrule
\multicolumn{10}{@{}l}{\textbf{Commercial API models}}\\
\bottomrule
\end{tabular}
\end{fitwidthtable}
\caption{Detailed knowledge results: normalized scores (higher, better) and ranks (lower, better). \scorecolorlegend}
\label{tab:detailed_knowledge_results}
\end{table*}

  \FloatBarrier

  \FloatBarrier
  \begin{table*}[ht]
\centering
\scriptsize
\begin{fitwidthtable}
\begin{tabular}{@{}lr@{~~}r@{~~}c@{~~}c@{~~}c@{~~}c@{~~}c@{~~}c@{}}
\toprule
Model & Avg. score & Avg. rank & \rotatebox[origin=l]{90}{Claire Wolof / Conversation Coherence} & \rotatebox[origin=l]{90}{Corporate FAQ French / Conversational MRQA} & \rotatebox[origin=l]{90}{Corporate FAQ Wolof / Conversational MRQA} & \rotatebox[origin=l]{90}{PersonaChat / Conversation Coherence} & \rotatebox[origin=l]{90}{Corporate / Conversation Coherence} & \rotatebox[origin=l]{90}{WebNLG QA / Conversational MRQA} \\
\midrule
\multicolumn{9}{@{}l}{\textbf{Baseline open-weight models}}\\
Qwen3-14B & \scoremid{.33} & 11.5 & \scoremid{\sr{-.60}{13}} & \scoretop{\sr{.72}{4}} & \scoremid{\sr{.13}{15}} & \scoretop{\sr{.92}{7}} & \scorelow{\sr{.15}{19}} & \scorehigh{\sr{.69}{11}} \\

Qwen3-8B & \scoremid{.31} & 10.8 & \scoremid{\sr{-.76}{16}} & \scorehigh{\sr{.70}{8}} & \scoremid{\sr{.16}{12}} & \scoretop{\sr{.90}{8}} & \scorelow{\sr{.16}{17}} & \scoretop{\sr{.71}{4}} \\

Qwen3-4B & \scorelow{.26} & 13.2 & \scorelow{\sr{-1.01}{19}} & \scorehigh{\sr{.69}{10}} & \scoremid{\sr{.14}{14}} & \scorehigh{\sr{.88}{9}} & \scorelow{\sr{.19}{16}} & \scorehigh{\sr{.69}{11}} \\

Qwen3-1.7B & \scorelow{.23} & 15.3 & \scorelow{\sr{-.88}{18}} & \scorehigh{\sr{.63}{13}} & \scorelow{\sr{.09}{18}} & \scorehigh{\sr{.80}{11}} & \scorelow{\sr{.16}{17}} & \scorehigh{\sr{.62}{15}} \\

Qwen3-0.6B & \scorelow{.25} & 16.2 & \scoremid{\sr{-.61}{14}} & \scoremid{\sr{.58}{16}} & \scorelow{\sr{.08}{19}} & \scoremid{\sr{.71}{16}} & \scoremid{\sr{.28}{14}} & \scoremid{\sr{.48}{18}} \\

\midrule
\multicolumn{9}{@{}l}{\textbf{Commercial API models}}\\
\bottomrule
\end{tabular}
\end{fitwidthtable}
\caption{Detailed chat results: normalized scores (higher, better) and ranks (lower, better). \scorecolorlegend}
\label{tab:detailed_chat_results}
\end{table*}

  \FloatBarrier

  \FloatBarrier
  \begin{table*}[ht]
\centering
\scriptsize
\begin{fitwidthtable}
\begin{tabular}{@{}lr@{~~}r@{~~}c@{~~}c@{~~}c@{~~}c@{~~}c@{~~}c@{}}
\toprule
Model & Avg. score & Avg. rank & \rotatebox[origin=l]{90}{ARC-Challenge / Scientific Reasoning} & \rotatebox[origin=l]{90}{GPQA Diamond / MCQA / French} & \rotatebox[origin=l]{90}{GPQA Diamond / MCQA/ English} & \rotatebox[origin=l]{90}{GSM8K / Mathematical Reasoning} & \rotatebox[origin=l]{90}{HumanEval Code Generation} & \rotatebox[origin=l]{90}{IrokoBench / MGSM} \\
\midrule
\multicolumn{9}{@{}l}{\textbf{Baseline open-weight models}}\\
Qwen3-14B & \scoremid{.37} & 10.3 & \scoremid{\sr{.12}{12}} & \scoremid{\sr{.14}{9}} & \scoremid{\sr{.32}{11}} & \scoretop{\sr{.86}{7}} & \scorehigh{\sr{.74}{10}} & \scorelow{\sr{.05}{13}} \\

Qwen3-8B & \scoremid{.30} & 13.8 & \scoremid{\sr{.15}{11}} & \scorelow{\sr{.01}{15}} & \scorelow{\sr{.29}{14}} & \scorehigh{\sr{.81}{12}} & \scoremid{\sr{.54}{14}} & \scorelow{\sr{.02}{17}} \\

Qwen3-4B & \scoremid{.33} & 12.5 & \scorelow{\sr{.03}{17}} & \scoremid{\sr{.12}{10}} & \scorelow{\sr{.30}{13}} & \scoretop{\sr{.84}{8}} & \scorehigh{\sr{.62}{13}} & \scorelow{\sr{.04}{14}} \\

Qwen3-1.7B & \scorelow{.26} & 15.7 & \scorelow{\sr{.05}{13}} & \scorelow{\sr{-.01}{16}} & \scorelow{\sr{.26}{18}} & \scorehigh{\sr{.76}{13}} & \scoremid{\sr{.53}{16}} & \scorelow{\sr{.01}{18}} \\

Qwen3-0.6B & \scorelow{.21} & 18.0 & \scorelow{\sr{.02}{18}} & \scorelow{\sr{-.03}{18}} & \scorelow{\sr{.23}{19}} & \scorehigh{\sr{.68}{16}} & \scoremid{\sr{.37}{18}} & \scorelow{\sr{.00}{19}} \\

\midrule
\multicolumn{9}{@{}l}{\textbf{Commercial API models}}\\
\bottomrule
\end{tabular}
\end{fitwidthtable}
\caption{Detailed reasoning results: normalized scores (higher, better) and ranks (lower, better). \scorecolorlegend}
\label{tab:detailed_reasoning_results}
\end{table*}

  \FloatBarrier

  \FloatBarrier
  \begin{table*}[ht]
\centering
\scriptsize
\begin{fitwidthtable}
\begin{tabular}{@{}lr@{~~}r@{~~}c@{}}
\toprule
Model & Avg. score & Avg. rank & \rotatebox[origin=l]{90}{HumanEval Code Generation} \\
\midrule
\multicolumn{4}{@{}l}{\textbf{Baseline open-weight models}}\\
Qwen3-14B & \scorehigh{.74} & 10.0 & \scorehigh{\sr{.74}{10}} \\

Qwen3-8B & \scoremid{.54} & 14.0 & \scoremid{\sr{.54}{14}} \\

Qwen3-4B & \scorehigh{.62} & 13.0 & \scorehigh{\sr{.62}{13}} \\

Qwen3-1.7B & \scoremid{.53} & 16.0 & \scoremid{\sr{.53}{16}} \\

Qwen3-0.6B & \scoremid{.37} & 18.0 & \scoremid{\sr{.37}{18}} \\

\midrule
\multicolumn{4}{@{}l}{\textbf{Commercial API models}}\\
Claude-sonnet-4-5 & \scoretop{.98} & 2.0 & \scoretop{\sr{.98}{2}} \\

Claude-haiku-4-5 & \scoretop{.93} & 6.0 & \scoretop{\sr{.93}{6}} \\

Gemini-2.5-pro & \scoretop{.94} & 5.0 & \scoretop{\sr{.94}{5}} \\

Gemini-2.5-flash & \scoretop{.96} & 4.0 & \scoretop{\sr{.96}{4}} \\

GPT-5 & \textbf{\scoretop{.99}} & \textbf{1.0} & \textbf{\scoretop{\sr{.99}{1}}} \\

GPT-5-nano & \scoretop{.98} & 2.0 & \scoretop{\sr{.98}{2}} \\

Mistral-medium-3 & \scoretop{.88} & 7.0 & \scoretop{\sr{.88}{7}} \\

\midrule
\multicolumn{4}{@{}l}{\textbf{Wolof-adapted open-weight models (ours)}}\\
Wolof-Qwen3-14B & \scorehigh{.79} & 8.0 & \scorehigh{\sr{.79}{8}} \\

Wolof-Qwen3-8B & \scorehigh{.69} & 11.0 & \scorehigh{\sr{.69}{11}} \\

Wolof-Qwen3-4B & \scorehigh{.64} & 12.0 & \scorehigh{\sr{.64}{12}} \\

Wolof-Qwen3-1.7B & \scoremid{.48} & 17.0 & \scoremid{\sr{.48}{17}} \\

Wolof-Qwen3-0.6B & \scorelow{.20} & 19.0 & \scorelow{\sr{.20}{19}} \\

\midrule
\multicolumn{4}{@{}l}{\textbf{Wolof-aware models from the literature}}\\
Oolel-v0.1 (7B) & \scorehigh{.77} & 9.0 & \scorehigh{\sr{.77}{9}} \\

tiny-aya-earth (3B) & \scoremid{.54} & 14.0 & \scoremid{\sr{.54}{14}} \\

\bottomrule
\end{tabular}
\end{fitwidthtable}
\caption{Detailed code results: normalized scores (higher, better) and ranks (lower, better). \scorecolorlegend}
\label{tab:detailed_code_results}
\end{table*}

  \FloatBarrier

  \FloatBarrier
  \begin{sidewaystable*}[p]
\centering
\scriptsize
\begin{fitwidthtable}
\begin{tabular}{@{}lr@{~~}r@{~~}c@{~~}c@{~~}c@{~~}c@{~~}c@{~~}c@{~~}c@{~~}c@{~~}c@{~~}c@{~~}c@{~~}c@{~~}c@{~~}c@{~~}c@{~~}c@{~~}c@{~~}c@{~~}c@{~~}c@{~~}c@{~~}c@{}}
\toprule
Model & Avg. score & Avg. rank & \rotatebox[origin=l]{90}{AfrIFact / Claim Classification} & \rotatebox[origin=l]{90}{Aya Wolof / POS Tagging} & \rotatebox[origin=l]{90}{BFCL / Function Calling} & \rotatebox[origin=l]{90}{GPQA Diamond / MCQA / French} & \rotatebox[origin=l]{90}{GPQA Diamond / MCQA/ English} & \rotatebox[origin=l]{90}{HumanEval Code Generation} & \rotatebox[origin=l]{90}{IrokoBench / MMLU} & \rotatebox[origin=l]{90}{IrokoBench / XNLI} & \rotatebox[origin=l]{90}{MasakhaNER / NER (News Domain)} & \rotatebox[origin=l]{90}{MMLU / MCQA / English} & \rotatebox[origin=l]{90}{MMLU / MCQA / French} & \rotatebox[origin=l]{90}{Corporate / Dialog Act Labelling} & \rotatebox[origin=l]{90}{Corporate / Full Analysis} & \rotatebox[origin=l]{90}{Corporate / GDPR/PII Extraction} & \rotatebox[origin=l]{90}{Corporate / NER (Public Entities)} & \rotatebox[origin=l]{90}{Corporate / Resolution Classification} & \rotatebox[origin=l]{90}{Corporate / Sentiment Classification} & \rotatebox[origin=l]{90}{WolBanking77 / Intent Classification / English} & \rotatebox[origin=l]{90}{WolBanking77 / Intent Classification / French} & \rotatebox[origin=l]{90}{WolBanking77 / Intent Classification / Wolof} & \rotatebox[origin=l]{90}{WolBanking77 Audio / Intent Classification / ASR} & \rotatebox[origin=l]{90}{WolBanking77 Audio / Intent Classification / Reference} \\
\midrule
\multicolumn{25}{@{}l}{\textbf{Baseline open-weight models}}\\
\bottomrule
\end{tabular}
\end{fitwidthtable}
\caption{Detailed formatting results: normalized scores (higher, better) and ranks (lower, better). \scorecolorlegend}
\label{tab:detailed_formatting_results}
\end{sidewaystable*}

  \FloatBarrier

  \FloatBarrier
  \begin{table*}[ht]
\centering
\scriptsize
\begin{fitwidthtable}
\begin{tabular}{@{}lr@{~~}r@{~~}c@{~~}c@{~~}c@{~~}c@{}}
\toprule
Model & Avg. score & Avg. rank & \rotatebox[origin=l]{90}{IFBench / Instruction Following} & \rotatebox[origin=l]{90}{MIFEval / Instruction Following / English} & \rotatebox[origin=l]{90}{MIFEval / Instruction Following / French} & \rotatebox[origin=l]{90}{Multi-IF / Instruction Following} \\
\midrule
\multicolumn{7}{@{}l}{\textbf{Baseline open-weight models}}\\
Qwen3-14B & \scorehigh{.67} & 6.0 & \scoremid{\sr{.32}{5}} & \scoretop{\sr{.83}{6}} & \scorehigh{\sr{.75}{7}} & \scorehigh{\sr{.80}{6}} \\

Qwen3-8B & \scorehigh{.63} & 8.5 & \scoremid{\sr{.23}{9}} & \scorehigh{\sr{.77}{9}} & \scorehigh{\sr{.74}{9}} & \scorehigh{\sr{.79}{7}} \\

Qwen3-4B & \scorehigh{.59} & 10.2 & \scoremid{\sr{.19}{10}} & \scorehigh{\sr{.74}{11}} & \scorehigh{\sr{.71}{10}} & \scorehigh{\sr{.72}{10}} \\

Qwen3-1.7B & \scorehigh{.50} & 14.0 & \scoremid{\sr{.13}{15}} & \scorehigh{\sr{.62}{15}} & \scorehigh{\sr{.63}{13}} & \scorehigh{\sr{.61}{13}} \\

Qwen3-0.6B & \scoremid{.38} & 15.8 & \scoremid{\sr{.13}{15}} & \scoremid{\sr{.52}{16}} & \scoremid{\sr{.41}{16}} & \scoremid{\sr{.48}{16}} \\

\midrule
\multicolumn{7}{@{}l}{\textbf{Commercial API models}}\\
\bottomrule
\end{tabular}
\end{fitwidthtable}
\caption{Detailed instruction-following results: normalized scores (higher, better) and ranks (lower, better). \scorecolorlegend}
\label{tab:detailed_instruction_following_results}
\end{table*}

  \FloatBarrier

  \FloatBarrier
  \begin{table*}[ht]
\centering
\scriptsize
\begin{fitwidthtable}
\begin{tabular}{@{}lr@{~~}r@{~~}c@{}}
\toprule
Model & Avg. score & Avg. rank & \rotatebox[origin=l]{90}{BFCL / Function Calling} \\
\midrule
\multicolumn{4}{@{}l}{\textbf{Baseline open-weight models}}\\
Qwen3-14B & \scoretop{.70} & 2.0 & \scoretop{\sr{.70}{2}} \\

Qwen3-8B & \scoretop{.67} & 6.0 & \scoretop{\sr{.67}{6}} \\

Qwen3-4B & \scoretop{.68} & 5.0 & \scoretop{\sr{.68}{5}} \\

Qwen3-1.7B & \scorehigh{.61} & 9.0 & \scorehigh{\sr{.61}{9}} \\

Qwen3-0.6B & \scorehigh{.53} & 12.0 & \scorehigh{\sr{.53}{12}} \\

\midrule
\multicolumn{4}{@{}l}{\textbf{Commercial API models}}\\
Claude-sonnet-4-5 & \textbf{\scoretop{.72}} & \textbf{1.0} & \textbf{\scoretop{\sr{.72}{1}}} \\

Claude-haiku-4-5 & \scoretop{.70} & 2.0 & \scoretop{\sr{.70}{2}} \\

Gemini-2.5-pro & \scoretop{.65} & 7.0 & \scoretop{\sr{.65}{7}} \\

Gemini-2.5-flash & \scoretop{.69} & 4.0 & \scoretop{\sr{.69}{4}} \\

GPT-5 & \scorehigh{.59} & 10.0 & \scorehigh{\sr{.59}{10}} \\

GPT-5-nano & \scorehigh{.58} & 11.0 & \scorehigh{\sr{.58}{11}} \\

Mistral-medium-3 & \scoretop{.62} & 8.0 & \scoretop{\sr{.62}{8}} \\

\midrule
\multicolumn{4}{@{}l}{\textbf{Wolof-adapted open-weight models (ours)}}\\
Wolof-Qwen3-14B & \scoremid{.35} & 15.0 & \scoremid{\sr{.35}{15}} \\

Wolof-Qwen3-8B & \scorehigh{.46} & 13.0 & \scorehigh{\sr{.46}{13}} \\

Wolof-Qwen3-4B & \scorehigh{.37} & 14.0 & \scorehigh{\sr{.37}{14}} \\

Wolof-Qwen3-1.7B & \scoremid{.24} & 16.0 & \scoremid{\sr{.24}{16}} \\

Wolof-Qwen3-0.6B & \scoremid{.23} & 17.0 & \scoremid{\sr{.23}{17}} \\

\midrule
\multicolumn{4}{@{}l}{\textbf{Wolof-aware models from the literature}}\\
Oolel-v0.1 (7B) & \scorelow{.00} & 18.0 & \scorelow{\sr{.00}{18}} \\

tiny-aya-earth (3B) & \scorelow{.00} & 18.0 & \scorelow{\sr{.00}{18}} \\

\bottomrule
\end{tabular}
\end{fitwidthtable}
\caption{Detailed function-calling results: normalized scores (higher, better) and ranks (lower, better). \scorecolorlegend}
\label{tab:detailed_function_calling_results}
\end{table*}

  \FloatBarrier

  \FloatBarrier
  \begin{table*}[ht]
\centering
\scriptsize
\begin{fitwidthtable}
\begin{tabular}{@{}lr@{~~}r@{~~}c@{~~}c@{}}
\toprule
Model & Avg. score & Avg. rank & \rotatebox[origin=l]{90}{Corporate FAQ Wolof / Long Context MRQA} & \rotatebox[origin=l]{90}{RULER / MRQA} \\
\midrule
\multicolumn{5}{@{}l}{\textbf{Baseline open-weight models}}\\
Qwen3-14B & \scorehigh{.45} & 10.0 & \scoremid{\sr{.09}{15}} & \scoretop{\sr{.80}{5}} \\

Qwen3-8B & \scorehigh{.45} & 9.5 & \scoremid{\sr{.12}{12}} & \scoretop{\sr{.78}{7}} \\

Qwen3-4B & \scorehigh{.42} & 13.0 & \scoremid{\sr{.10}{14}} & \scoretop{\sr{.73}{12}} \\

Qwen3-1.7B & \scorehigh{.37} & 16.0 & \scorelow{\sr{.03}{18}} & \scorehigh{\sr{.70}{14}} \\

Qwen3-0.6B & \scoremid{.18} & 18.5 & \scorelow{\sr{.02}{19}} & \scoremid{\sr{.33}{18}} \\

\midrule
\multicolumn{5}{@{}l}{\textbf{Commercial API models}}\\
\bottomrule
\end{tabular}
\end{fitwidthtable}
\caption{Detailed long-context results: normalized scores (higher, better) and ranks (lower, better). \scorecolorlegend}
\label{tab:detailed_long_context_results}
\end{table*}

  \FloatBarrier

  \FloatBarrier
  \begin{table*}[ht]
\centering
\scriptsize
\begin{fitwidthtable}
\begin{tabular}{@{}lr@{~~}r@{~~}c@{~~}c@{~~}c@{~~}c@{}}
\toprule
Model & Avg. score & Avg. rank & \rotatebox[origin=l]{90}{BBQ / Demographic Bias MCQA} & \rotatebox[origin=l]{90}{HaluEval / QA Hallucination Detection} & \rotatebox[origin=l]{90}{ToxiGen / Implicit Hate Detection} & \rotatebox[origin=l]{90}{WildJailbreak / Harmful Request Refusal} \\
\midrule
\multicolumn{7}{@{}l}{\textbf{Baseline open-weight models}}\\
Qwen3-14B & \scorehigh{.54} & 8.2 & \scoremid{\sr{.36}{8}} & \scorehigh{\sr{.78}{7}} & \scoretop{\sr{.71}{7}} & \scoremid{\sr{.29}{11}} \\

Qwen3-8B & \scoremid{.45} & 10.5 & \scoremid{\sr{.21}{13}} & \scorehigh{\sr{.78}{7}} & \scorehigh{\sr{.64}{10}} & \scoremid{\sr{.19}{12}} \\

Qwen3-4B & \scoremid{.38} & 12.8 & \scorelow{\sr{.06}{17}} & \scorehigh{\sr{.70}{9}} & \scorehigh{\sr{.64}{10}} & \scorelow{\sr{.11}{15}} \\

Qwen3-1.7B & \scoremid{.32} & 14.8 & \scorelow{\sr{.03}{18}} & \scoremid{\sr{.67}{10}} & \scorehigh{\sr{.54}{13}} & \scorelow{\sr{.04}{18}} \\

Qwen3-0.6B & \scorelow{.30} & 15.5 & \scoremid{\sr{.29}{10}} & \scoremid{\sr{.53}{15}} & \scoremid{\sr{.36}{18}} & \scorelow{\sr{.01}{19}} \\

\midrule
\multicolumn{7}{@{}l}{\textbf{Commercial API models}}\\
\bottomrule
\end{tabular}
\end{fitwidthtable}
\caption{Detailed safety results: normalized scores (higher, better) and ranks (lower, better). \scorecolorlegend}
\label{tab:detailed_safety_results}
\end{table*}

  \FloatBarrier

\twocolumn

\section{Detailed tasks}
\label{app:tasks}

For public datasets, the associated reference is listed inside the corresponding task subsection under a \textit{Paper} paragraph. Each description also specifies the exact split, subset, language, and sample count used for evaluation; proprietary customer-service datasets are described only at the task level and are not released. When multiple metrics are computed for a task, the metric identified for reporting is the one used in this paper and, when applicable, as the aggregate task score; additional metrics are logged for diagnostics and analysis but are not reported here.

\subsection{\texttt{afrifact\_\-claim\_\-classification}}
\paragraph{Description.}
This task evaluates evidence-grounded claim verification in Wolof using AfrIFact. The evaluation uses the Wolof subset and 1658 samples. System inputs include \texttt{language}, \texttt{domain}, \texttt{context}, and \texttt{claim}. System outputs include one fact-checking label. The metric used for reporting in this paper is \texttt{macro\_f1}.

\paragraph{Paper: } \cite{azime2026afrifact}

\paragraph{Example prompt template(s).} ~

\prompttemplate{tasks/afrifact_claim_classification.sample.json}

\subsection{\texttt{afrifact\_\-evidence\_\-extraction}}
\paragraph{Description.}
This task evaluates evidence-span extraction in Wolof using AfrIFact. The evaluation uses the Wolof subset and 1658 samples. System inputs include \texttt{language}, \texttt{domain}, \texttt{claim}, \texttt{document}, and \texttt{max\_evidences}. System outputs include minimal evidence spans copied from the document. The metric used for reporting in this paper is \texttt{partial\_f1}.

\paragraph{Paper: } \cite{azime2026afrifact}

\paragraph{Example prompt template(s).} ~

\prompttemplate{tasks/afrifact_evidence_extraction.sample.json}

\subsection{\texttt{arc\_\-challenge\_\-reasoning}}
\paragraph{Description.}
This task the Challenge Set contains questions that are difficult for retrieval and co-occurrence methods, requiring actual reasoning. The evaluation uses \texttt{allenai/ai2\_arc}, the \texttt{test} split, and 1172 samples. System inputs include \texttt{question}, \texttt{choices}, and \texttt{choice\_labels}. System outputs include \texttt{predicted\_answer\_index}. The metric used for reporting in this paper is \texttt{accuracy}.

\paragraph{Paper: } \cite{clark2018arc}

\paragraph{Example prompt template(s).} ~

\prompttemplate{tasks/arc_challenge_reasoning.sample.json}

\subsection{\texttt{aya\_\-wolof\_\-culture\_\-qa}}
\paragraph{Description.}
This task evaluates Wolof cultural knowledge QA using the Aya dataset. This task tests factual accuracy on cultural/geographical/historical questions, relevance of answers, and wolof language fluency. The evaluation uses \texttt{wolof-lm/aya-dataset-wolof-annotated}, the \texttt{test} split, and 865 samples. System inputs include \texttt{question} and \texttt{category}. System outputs include \texttt{answer}. The metric used for reporting in this paper is \texttt{llm\_overall}. Other computed metrics are logged for diagnostic analysis and include \texttt{bleu}, \texttt{meteor}, and \texttt{llm\_correctness}.

\paragraph{Paper: } \cite{singh-etal-2024-aya}

\paragraph{Example prompt template(s).} ~

\prompttemplate{tasks/aya_wolof_culture_qa.sample.json}

\subsection{\texttt{aya\_\-wolof\_\-linguistic\_\-analysis}}
\paragraph{Description.}
This task evaluates Wolof POS tagging using the Aya dataset. This task tests correct identification of parts of speech and understanding of Wolof grammatical structure. The evaluation uses \texttt{wolof-lm/aya-dataset-wolof-annotated}, the \texttt{test} split, and 159 samples. System inputs include \texttt{sentence}. System outputs include \texttt{pos\_tags}. The metric used for reporting in this paper is \texttt{token\_accuracy}. Other computed metrics are logged for diagnostic analysis and include \texttt{exact\_match} and \texttt{precision\_macro}.

\paragraph{Paper: } \cite{singh-etal-2024-aya}

\paragraph{Example prompt template(s).} ~

\prompttemplate{tasks/aya_wolof_linguistic_analysis.sample.json}

\subsection{\texttt{aya\_\-wolof\_\-orthographic\_\-correction}}
\paragraph{Description.}
This task tests detection of spelling errors, correct application of Wolof orthographic rules, and preservation of correct text. The evaluation uses \texttt{wolof-lm/aya-dataset-wolof-annotated}, the \texttt{test} split, and 244 samples. System inputs include \texttt{text\_to\_correct}. System outputs include \texttt{corrected\_text}. The metric used for reporting in this paper is \texttt{char\_edit\_distance}. Other computed metrics are logged for diagnostic analysis and include \texttt{word\_accuracy} and \texttt{exact\_match}.

\paragraph{Paper: } \cite{singh-etal-2024-aya}

\paragraph{Example prompt template(s).} ~

\prompttemplate{tasks/aya_wolof_orthographic_correction.sample.json}

\subsection{\texttt{aya\_\-wolof\_\-translation\_\-from\_\-wol}}
\paragraph{Description.}
This task evaluates translation FROM Wolof to English or French using the Aya dataset. The evaluation uses \texttt{wolof-lm/aya-dataset-wolof-annotated}, the \texttt{test} split, and 16 samples. System inputs include \texttt{source\_text} and \texttt{target\_language}. System outputs include \texttt{translation}. The metric used for reporting in this paper is \texttt{meteor}. Other computed metrics are logged for diagnostic analysis and include \texttt{bleu} and \texttt{chrf}.

\paragraph{Paper: } \cite{singh-etal-2024-aya}

\paragraph{Example prompt template(s).} ~

\prompttemplate{tasks/aya_wolof_translation_from_wol.sample.json}

\subsection{\texttt{aya\_\-wolof\_\-translation\_\-to\_\-wol}}
\paragraph{Description.}
This task evaluates translation TO Wolof from English or French using the Aya dataset. The evaluation uses \texttt{wolof-lm/aya-dataset-wolof-annotated}, the \texttt{test} split, and 31 samples. System inputs include \texttt{source\_text} and \texttt{source\_language}. System outputs include \texttt{translation}. The metric used for reporting in this paper is \texttt{meteor}. Other computed metrics are logged for diagnostic analysis and include \texttt{bleu} and \texttt{chrf}.

\paragraph{Paper: } \cite{singh-etal-2024-aya}

\paragraph{Example prompt template(s).} ~

\prompttemplate{tasks/aya_wolof_translation_to_wol.sample.json}

\subsection{\texttt{bbq\_\-demographic\_\-bias\_\-mcqa}}
\paragraph{Description.}
This task measures demographic biases in language models by presenting multiple-choice questions with ambiguous and disambiguated contexts. In ambiguous contexts, a model relying on stereotypes will make biased choices. The evaluation uses 500 samples. System inputs include \texttt{context}, \texttt{question}, \texttt{choices}, \texttt{category}, and other metadata. System outputs include \texttt{predicted\_answer\_index}. The metric used for reporting in this paper is \texttt{accuracy}. Other computed metrics are logged for diagnostic analysis and include \texttt{accuracy\_ambiguous} and \texttt{accuracy\_disambiguated}.

\paragraph{Paper: } \cite{parrish2022bbq}

\paragraph{Example prompt template(s).} ~

\prompttemplate{tasks/bbq_demographic_bias_mcqa.sample.json}

\subsection{\texttt{berkeley\_\-function\_\-calling}}
\paragraph{Description.}
This task consists in evaluating the single-turn instruction following capabilities of a model in multiple languages \url{https://arxiv.org/abs/2410.15553} (note: Multi-IF is potentially multi-turn, but here we only consider the first turn). The evaluation uses 3641 samples. System inputs include \texttt{system}, \texttt{question}, \texttt{function}, \texttt{possible\_answer}, and other metadata. System outputs include \texttt{answer}. The metric used for reporting in this paper is \texttt{micro\_accuracy}. Other computed metrics are logged for diagnostic analysis and include \texttt{macro\_accuracy} and \texttt{simple\_python}.

\paragraph{Paper: } \cite{berkeley2024bfcl}

\paragraph{Example prompt template(s).} ~

\prompttemplate{tasks/berkeley_function_calling.sample.json}

\subsection{\texttt{claire\_\-wolof\_\-conversation\_\-coherence}}
\paragraph{Description.}
This task evaluates conversational coherence on Wolof dialogues from the Claire dataset. This task tests scenario/persona following, context awareness across turns, response fluency, and language consistency (Wolof). The evaluation uses \texttt{wolof-lm/claire\_one\_to\_one\_dialogue-instructions}, the \texttt{validation} split, and 292 samples. System inputs include \texttt{history} and \texttt{current\_utterance}. System outputs include \texttt{response}. The metric used for reporting in this paper is \texttt{scenario\_coherence}. Other computed metrics are logged for diagnostic analysis and include \texttt{bleu} and \texttt{meteor}.

\paragraph{Paper: } \cite{hunter2023claire}

\paragraph{Example prompt template(s).} ~

\prompttemplate{tasks/claire_wolof_conversation_coherence.sample.json}

\subsection{\texttt{dialogsum\_\-conversation\_\-summarization}}
\paragraph{Description.}
This task tests summary quality (ROUGE metrics), coherence of generated summaries, faithfulness to the original dialogue, and conciseness of the summary. The evaluation uses \texttt{knkarthick/dialogsum}, the \texttt{test} split, and 200 samples. System inputs include \texttt{dialogue}. System outputs include \texttt{summary}. The metric used for reporting in this paper is \texttt{rouge1}. Other computed metrics are logged for diagnostic analysis and include \texttt{overall\_quality}, \texttt{faithfulness\_llm}, and \texttt{conciseness\_llm}.

\paragraph{Paper: } \cite{chen2021dialogsum}

\paragraph{Example prompt template(s).} ~

\prompttemplate{tasks/dialogsum_conversation_summarization.sample.json}

\subsection{\texttt{corporate\_\-faq\_\-french\_\-conversational\_\-mrqa}}
\paragraph{Description.}
This task consists in answering question given a source document. Questions, history of questions-answers, documents and answers are in French. The evaluation uses the \texttt{test} split and 122 samples. System inputs include \texttt{question}, \texttt{history}, and \texttt{document}. System outputs include \texttt{answer}. The metric used for reporting in this paper is \texttt{llm\_good}. Other computed metrics are logged for diagnostic analysis and include \texttt{bleu}, \texttt{meteor}, and \texttt{llm\_mos}.

\paragraph{Example prompt template(s).} ~

\prompttemplate{tasks/corporate_faq_french_conversational_mrqa.sample.json}

\subsection{\texttt{corporate\_\-faq\_\-french\_\-mrqa}}
\paragraph{Description.}
This task consists in answering question given a source document. Questions, documents and answers are in French. The evaluation uses the \texttt{test} split and 122 samples. System inputs include \texttt{question} and \texttt{document}. System outputs include \texttt{answer}. The metric used for reporting in this paper is \texttt{llm\_good}. Other computed metrics are logged for diagnostic analysis and include \texttt{bleu}, \texttt{meteor}, and \texttt{llm\_mos}.

\paragraph{Example prompt template(s).} ~

\prompttemplate{tasks/corporate_faq_french_mrqa.sample.json}

\subsection{\texttt{corporate\_\-faq\_\-wolof\_\-conversational\_\-mrqa}}
\paragraph{Description.}
This task consists in answering question given a source document. Questions, history of previous questions-answers, documents and answers are in Wolof. The evaluation uses the \texttt{test} split and 122 samples. System inputs include \texttt{question}, \texttt{history}, and \texttt{document}. System outputs include \texttt{answer}. The metric used for reporting in this paper is \texttt{meteor}. Other computed metrics are logged for diagnostic analysis and include \texttt{bleu}.

\paragraph{Example prompt template(s).} ~

\prompttemplate{tasks/corporate_faq_wolof_conversational_mrqa.sample.json}

\subsection{\texttt{corporate\_\-faq\_\-wolof\_\-long\_\-context\_\-mrqa}}
\paragraph{Description.}
This task creates long contexts by combining FAQ documents with Wolof distractors from talkwalker-v2, then tests the model's ability to find and answer questions from the relevant document. Only uses "sentence" form questions for clarity. The evaluation uses 99 samples. System inputs include \texttt{question}, \texttt{document}, \texttt{needle\_position}, \texttt{num\_distractors}, and other metadata. System outputs include \texttt{answer}. The metric used for reporting in this paper is \texttt{meteor}. Other computed metrics are logged for diagnostic analysis and include \texttt{bleu} and \texttt{bleu\_by\_position}.

\paragraph{Example prompt template(s).} ~

\prompttemplate{tasks/corporate_faq_wolof_long_context_mrqa.sample.json}

\subsection{\texttt{corporate\_\-faq\_\-wolof\_\-mrqa}}
\paragraph{Description.}
This task consists in answering question given a source document. Questions, documents and answers are in Wolof. The evaluation uses the \texttt{test} split and 122 samples. System inputs include \texttt{question} and \texttt{document}. System outputs include \texttt{answer}. The metric used for reporting in this paper is \texttt{meteor}. Other computed metrics are logged for diagnostic analysis and include \texttt{bleu}.

\paragraph{Example prompt template(s).} ~

\prompttemplate{tasks/corporate_faq_wolof_mrqa.sample.json}

\subsection{\texttt{corporate\_\-conversation\_\-coherence}}
\paragraph{Description.}
This task tests context awareness across turns, response fluency, and language respect (Wolof/French code-switching). The evaluation uses the \texttt{test} split and 287 samples. System inputs include \texttt{history} and \texttt{current\_utterance}. System outputs include \texttt{response}. The metric used for reporting in this paper is \texttt{overall\_coherence}. Other computed metrics are logged for diagnostic analysis and include \texttt{bleu}, \texttt{meteor}, and \texttt{context\_awareness}.

\paragraph{Example prompt template(s).} ~

\prompttemplate{tasks/corporate_conversation_coherence.sample.json}

\subsection{\texttt{corporate\_\-dialog\_\-act\_\-labelling}}
\paragraph{Description.}
This task evaluates LLMs on classifying dialog acts for each conversation turn. Uses macro F1, exact set match, and Jaccard similarity as metrics. The evaluation uses 149 samples. System inputs include \texttt{messages} and \texttt{turn\_count}. System outputs include \texttt{dialog\_acts}. The metric used for reporting in this paper is \texttt{macro\_f1}. Other computed metrics are logged for diagnostic analysis and include \texttt{exact\_set\_match\_accuracy} and \texttt{mean\_jaccard}.

\paragraph{Example prompt template(s).} ~

\prompttemplate{tasks/corporate_dialog_act_labelling.sample.json}

\subsection{\texttt{corporate\_\-full\_\-analysis}}
\paragraph{Description.}
This task evaluates LLMs on performing complete multi-task analysis: summarization, sentiment, dialog acts, and NER in a single pass. The evaluation uses 150 samples. System inputs include \texttt{messages} and \texttt{turn\_count}. The metric used for reporting in this paper is \texttt{overall\_score}. Other computed metrics are logged for diagnostic analysis and include \texttt{rouge1\_fmeasure} and \texttt{rouge2\_fmeasure}.

\paragraph{Example prompt template(s).} ~

\prompttemplate{tasks/corporate_full_analysis.sample.json}

\subsection{\texttt{corporate\_\-gdpr\_\-labelling}}
\paragraph{Description.}
This task evaluates GDPR/PII entity extraction from in-house call center conversations for privacy compliance. The evaluation uses 150 samples. System inputs include \texttt{messages}. System outputs include \texttt{entities}. The metric used for reporting in this paper is \texttt{exact\_f1}. Other computed metrics are logged for diagnostic analysis and include \texttt{exact\_precision} and \texttt{exact\_recall}.

\paragraph{Example prompt template(s).} ~

\prompttemplate{tasks/corporate_gdpr_labelling.sample.json}

\subsection{\texttt{corporate\_\-ner}}
\paragraph{Description.}
This task evaluates LLMs on extracting public/general named entities from conversations. Uses exact and partial match F1 as primary metrics. The evaluation uses 150 samples. System inputs include \texttt{messages}. System outputs include \texttt{entities}. The metric used for reporting in this paper is \texttt{exact\_f1}. Other computed metrics are logged for diagnostic analysis and include \texttt{exact\_precision} and \texttt{exact\_recall}.

\paragraph{Example prompt template(s).} ~

\prompttemplate{tasks/corporate_ner.sample.json}

\subsection{\texttt{corporate\_\-resolution\_\-classification}}
\paragraph{Description.}
This task evaluates LLMs on classifying the resolution/outcome of call center conversations. This is a dialogue-level (not turn-level) classification task. The evaluation uses 149 samples. System inputs include \texttt{messages} and \texttt{turn\_count}. System outputs include \texttt{resolution}. The metric used for reporting in this paper is \texttt{macro\_f1}. Other computed metrics are logged for diagnostic analysis and include \texttt{accuracy} and \texttt{weighted\_f1}.

\paragraph{Example prompt template(s).} ~

\prompttemplate{tasks/corporate_resolution_classification.sample.json}

\subsection{\texttt{corporate\_\-sentiment\_\-classification}}
\paragraph{Description.}
This task evaluates LLMs on classifying sentiment for each conversation turn. Uses macro and weighted F1 scores as primary metrics. The evaluation uses 149 samples. System inputs include \texttt{messages} and \texttt{turn\_count}. System outputs include \texttt{sentiments}. The metric used for reporting in this paper is \texttt{macro\_f1}. Other computed metrics are logged for diagnostic analysis and include \texttt{weighted\_f1} and \texttt{accuracy}.

\paragraph{Example prompt template(s).} ~
\prompttemplate{tasks/corporate_sentiment_classification.sample.json}

\subsection{\texttt{corporate\_\-summarization}}
\paragraph{Description.}
This task evaluates LLMs on generating Wolof summaries of call center conversations. Metrics rOUGE (lexical overlap, no stemming for Wolof), mETEOR (semantic similarity), lLM-as-judge semantic score, lLM-as-judge language respect score (Wolof), and overall score (average of semantic and language). The evaluation uses 150 samples. System inputs include \texttt{messages}. System outputs include \texttt{summary}. The metric used for reporting in this paper is \texttt{llm\_overall}. Other computed metrics are logged for diagnostic analysis and include \texttt{llm\_semantic} and \texttt{llm\_language}.

\paragraph{Example prompt template(s).} ~

\prompttemplate{tasks/corporate_summarization.sample.json}

\subsection{\texttt{flores\_\-translate\_\-eng\_\-to\_\-wol}}
\paragraph{Description.}
This task consists of translating an english sentence into Wolof. The evaluation uses \texttt{openlanguagedata/flores\_plus}, the \texttt{devtest} split, and 1012 samples. System inputs include \texttt{topic}, \texttt{domain}, and \texttt{sentence\_eng\_Latn}. System outputs include \texttt{translation\_pred}. The metric used for reporting in this paper is \texttt{meteor}. Other computed metrics are logged for diagnostic analysis and include \texttt{bleu}.

\paragraph{Paper: } \cite{costajussa2022nllb}

\paragraph{Example prompt template(s).} ~

\prompttemplate{tasks/flores_translate_eng_to_wol.sample.json}

\subsection{\texttt{flores\_\-translate\_\-fra\_\-to\_\-wol}}
\paragraph{Description.}
This task consists of translating a french sentence into Wolof. The evaluation uses \texttt{openlanguagedata/flores\_plus}, the \texttt{devtest} split, and 1012 samples. System inputs include \texttt{topic}, \texttt{domain}, and \texttt{sentence\_fra\_Latn}. System outputs include \texttt{translation\_pred}. The metric used for reporting in this paper is \texttt{meteor}. Other computed metrics are logged for diagnostic analysis and include \texttt{bleu}.

\paragraph{Paper: } \cite{costajussa2022nllb}

\paragraph{Example prompt template(s).} ~

\prompttemplate{tasks/flores_translate_fra_to_wol.sample.json}

\subsection{\texttt{flores\_\-translate\_\-wol\_\-to\_\-eng}}
\paragraph{Description.}
This task consists of translating a Wolof sentence into English. The evaluation uses \texttt{openlanguagedata/flores\_plus}, the \texttt{devtest} split, and 1012 samples. System inputs include \texttt{topic}, \texttt{domain}, and \texttt{sentence\_wol\_Latn}. System outputs include \texttt{translation\_pred}. The metric used for reporting in this paper is \texttt{meteor}. Other computed metrics are logged for diagnostic analysis and include \texttt{bleu} and \texttt{llm\_score}.

\paragraph{Paper: } \cite{costajussa2022nllb}

\paragraph{Example prompt template(s).} ~

\prompttemplate{tasks/flores_translate_wol_to_eng.sample.json}

\subsection{\texttt{flores\_\-translate\_\-wol\_\-to\_\-fra}}
\paragraph{Description.}
This task consists of translating a Wolof sentence into French. The evaluation uses \texttt{openlanguagedata/flores\_plus}, the \texttt{devtest} split, and 5 samples. System inputs include \texttt{topic}, \texttt{domain}, and \texttt{sentence\_wol\_Latn}. System outputs include \texttt{translation\_pred}. The metric used for reporting in this paper is \texttt{meteor}. Other computed metrics are logged for diagnostic analysis and include \texttt{bleu} and \texttt{llm\_score}.

\paragraph{Paper: } \cite{costajussa2022nllb}

\paragraph{Example prompt template(s).} ~

\prompttemplate{tasks/flores_translate_wol_to_fra.sample.json}

\subsection{\texttt{flores\_\-translation}}
\paragraph{Description.}
This task evaluates translation quality using BLEU and METEOR metrics. Supports any language pair from the FLORES+ dataset. The evaluation uses the \texttt{devtest} split and 1012 samples. System inputs include \texttt{source\_text}. System outputs include \texttt{translation\_pred}. The metric used for reporting in this paper is \texttt{meteor}. Other computed metrics are logged for diagnostic analysis and include \texttt{bleu}.

\paragraph{Paper: } \cite{costajussa2022nllb}

\paragraph{English to French prompt template.} ~
\prompttemplate{tasks/flores_translation_eng_to_fra.sample.json}

\paragraph{English to Wolof prompt template.} ~
\prompttemplate{tasks/flores_translation_eng_to_wol.sample.json}

\paragraph{French to English prompt template.} ~
\prompttemplate{tasks/flores_translation_fra_to_eng.sample.json}

\paragraph{French to Wolof prompt template.} ~
\prompttemplate{tasks/flores_translation_fra_to_wol.sample.json}

\paragraph{Wolof to English prompt template.} ~
\prompttemplate{tasks/flores_translation_wol_to_eng.sample.json}

\paragraph{Wolof to French prompt template.} ~
\prompttemplate{tasks/flores_translation_wol_to_fra.sample.json}

\subsection{\texttt{gpqa\_\-english\_\-mcqa}}
\paragraph{Description.}
This task evaluates graduate-level scientific reasoning on the English GPQA Diamond subset. GPQA is a multiple-choice benchmark written and validated by domain experts in biology, chemistry, and physics. The evaluation uses \texttt{cloned-Idavidrein/gpqa}, configuration/subset \texttt{gpqa\_diamond}, the \texttt{test} split, language coverage English, and 198 samples. System inputs include \texttt{question}, \texttt{choices}, and \texttt{choice\_labels}. System outputs include \texttt{answer\_pred}. The metric used for reporting in this paper is \texttt{accuracy}.

\paragraph{Paper: } \cite{rein2023gpqa}

\paragraph{Example prompt template(s).} ~

\prompttemplate{tasks/gpqa_english_mcqa.sample.json}

\subsection{\texttt{gpqa\_\-french\_\-mcqa}}
\paragraph{Description.}
This task evaluates graduate-level scientific reasoning on the French GPQA Diamond subset. This task mirrors the English GPQA Diamond evaluation but uses a French translation of the benchmark. The evaluation uses \texttt{cloned-le-leaderboard/gpqa-fr}, configuration/subset \texttt{gpqa\_diamond}, the \texttt{test} split, language coverage French, and 197 samples. System inputs include \texttt{question}, \texttt{choices}, and \texttt{choice\_labels}. System outputs include \texttt{answer\_pred}. The metric used for reporting in this paper is \texttt{accuracy}.

\paragraph{Paper: } \cite{rein2023gpqa}

\paragraph{Example prompt template(s).} ~

\prompttemplate{tasks/gpqa_french_mcqa.sample.json}

\subsection{\texttt{gsm8k\_\-reasoning}}
\paragraph{Description.}
This task gSM8K consists of 8.5K high-quality, linguistically diverse grade school math word problems. Each problem requires 2-8 steps to solve and involves basic arithmetic operations. The evaluation uses \texttt{openai/gsm8k}, the \texttt{test} split, and 1319 samples. System inputs include \texttt{question}. System outputs include \texttt{numerical\_answer}. The metric used for reporting in this paper is \texttt{accuracy}.

\paragraph{Paper: } \cite{cobbe2021gsm8k}

\paragraph{Example prompt template(s).} ~

\prompttemplate{tasks/gsm8k_reasoning.sample.json}

\subsection{\texttt{halueval\_\-qa\_\-hallucination\_\-detection}}
\paragraph{Description.}
This task evaluates the ability of a language model to detect hallucinations in QA responses. Given background knowledge and a question, the model must identify which of two responses contains fabricated or incorrect information. The evaluation uses 500 samples. System inputs include \texttt{user\_request}, \texttt{response\_a}, \texttt{response\_b}, and \texttt{task}. System outputs include \texttt{predicted\_hallucinated}. The metric used for reporting in this paper is \texttt{accuracy}. Other computed metrics are logged for diagnostic analysis and include \texttt{f1}.

\paragraph{Paper: } \cite{li2023halueval}

\paragraph{Example prompt template(s).} ~

\prompttemplate{tasks/halueval_qa_hallucination_detection.sample.json}

\subsection{\texttt{humaneval\_\-code\_\-generation}}
\paragraph{Description.}
This task evaluates humaneval code generation. The evaluation uses \texttt{openai/openai\_humaneval}, the \texttt{test} split, and 164 samples. System inputs include \texttt{prompt} and \texttt{entry\_point}. System outputs include \texttt{generated\_code}. The metric used for reporting in this paper is \texttt{pass\_at\_1}. Other computed metrics are logged for diagnostic analysis and include \texttt{syntax\_error\_rate} and \texttt{runtime\_error\_rate}.

\paragraph{Paper: } \cite{chen2021humaneval}

\paragraph{Example prompt template(s).} ~

\prompttemplate{tasks/humaneval_code_generation.sample.json}

\subsection{\texttt{ifbench\_\-instruction\_\-following}}
\paragraph{Description.}
This task evaluates precise instruction-following generalization on English prompts. IFBench focuses on verifiable out-of-domain constraints that are designed to test whether a model can generalize beyond the common constraints used in earlier instruction-following benchmarks. The evaluation uses \texttt{IFBench}, the \texttt{test} split, language coverage English, and 300 samples. System inputs include \texttt{prompt} and \texttt{constraint\_ids}. System outputs include \texttt{response}. The metric used for reporting in this paper is \texttt{accuracy}. Other computed metrics are logged for diagnostic analysis and include \texttt{macro\_accuracy}.

\paragraph{Paper: } \cite{pyatkin2025ifbench}

\paragraph{Example prompt template(s).} ~

\prompttemplate{tasks/ifbench_instruction_following.sample.json}

\subsection{\texttt{irokobench\_\-mgsm}}
\paragraph{Description.}
This task consists of solving a grade school math problem. The evaluation uses \texttt{masakhane/afrimgsm}, the \texttt{test} split, and 250 samples. System inputs include \texttt{question}. System outputs include \texttt{answer\_pred}. The metric used for reporting in this paper is \texttt{exact\_match}. Other computed metrics are logged for diagnostic analysis and include \texttt{mae} and \texttt{mse}.

\paragraph{Paper: } \cite{adelani2025irokobench}

\paragraph{Example prompt template(s).} ~

\prompttemplate{tasks/irokobench_mgsm.sample.json}

\subsection{\texttt{irokobench\_\-mmlu}}
\paragraph{Description.}
This task consists of responding to a question, among 4 possible answers: A, B, C or D. The evaluation uses \texttt{masakhane/afrimmlu}, the \texttt{test} split, and 500 samples. System inputs include \texttt{subject}, \texttt{question}, and \texttt{choices}. System outputs include \texttt{answer\_pred}. The metric used for reporting in this paper is \texttt{accuracy}.

\paragraph{Paper: } \cite{adelani2025irokobench}

\paragraph{Example prompt template(s).} ~

\prompttemplate{tasks/irokobench_mmlu.sample.json}

\subsection{\texttt{irokobench\_\-xnli}}
\paragraph{Description.}
This task consists of classifying the logic between a premise and a hypothesis according to three possible labels: \texttt{0:"entailment"}, \texttt{1:"neutral"}, \texttt{2:"contradiction}. The evaluation uses \texttt{masakhane/afrixnli}, the \texttt{test} split, and 600 samples. System inputs include \texttt{premise} and \texttt{hypothesis}. System outputs include \texttt{label}. The metric used for reporting in this paper is \texttt{accuracy}. Other computed metrics are logged for diagnostic analysis and include \texttt{macro\_avg}, \texttt{weighted\_avg}, and \texttt{entailment}.

\paragraph{Paper: } \cite{adelani2025irokobench}

\paragraph{Example prompt template(s).} ~

\prompttemplate{tasks/irokobench_xnli.sample.json}

\subsection{\texttt{masakha\_\-ner}}
\paragraph{Description.}
This task evaluates LLMs on extracting news-domain named entities from Wolof text. Uses exact and partial match F1 as primary metrics, following the corporate\_ner task format. Note: The source dataset is token-level BIO annotation. The evaluation uses \texttt{masakhane/masakhaner}, configuration/subset \texttt{wol}, the \texttt{test} split, and 536 samples. System inputs include \texttt{tokens} and \texttt{text}. System outputs include \texttt{entities}. The metric used for reporting in this paper is \texttt{exact\_f1}. Other computed metrics are logged for diagnostic analysis and include \texttt{exact\_precision} and \texttt{exact\_recall}.

\paragraph{Paper: } \cite{adelani2022masakhaner2}

\paragraph{Example prompt template(s).} ~

\prompttemplate{tasks/masakha_ner.sample.json}

\subsection{\texttt{mifeval\_\-instruction\_\-following}}
\paragraph{Description.}
This task evaluates verifiable instruction following in English and French. M-IFEval extends the original IFEval protocol to multiple languages. The evaluation uses \texttt{M-IFEval}, the \texttt{test} split, language coverage English, French, and 541 English, 235 French samples. System inputs include \texttt{prompt}, \texttt{language}, and \texttt{instruction\_id\_list}. System outputs include \texttt{response}. The metric used for reporting in this paper is \texttt{strict\_prompt\_accuracy}. Other computed metrics are logged for diagnostic analysis and include \texttt{loose\_prompt\_accuracy}.

\paragraph{Paper: } \cite{dussolle2025mifeval}

\paragraph{English prompt template.} ~
\prompttemplate{tasks/mifeval_instruction_following_english.sample.json}

\paragraph{French prompt template.} ~
\prompttemplate{tasks/mifeval_instruction_following_french.sample.json}

\subsection{\texttt{mmlu\_\-mcqa\_\-english}}
\paragraph{Description.}
This task evaluates broad academic knowledge and language understanding on English MMLU. MMLU is a multiple-choice benchmark covering 57 academic and professional subjects, including STEM, humanities, social sciences, medicine, and law. The evaluation uses \texttt{openai/MMMLU}, the \texttt{test} split, language coverage English, and 14042 samples. System inputs include \texttt{subject}, \texttt{question}, \texttt{choices}, and \texttt{choice\_labels}. System outputs include \texttt{answer\_pred}. The metric used for reporting in this paper is \texttt{accuracy}.

\paragraph{Paper: } \cite{hendrycks2021mmlu,openai2024mmmlu}

\paragraph{Example prompt template(s).} ~

\prompttemplate{tasks/mmlu_mcqa_english.sample.json}

\subsection{\texttt{mmlu\_\-mcqa\_\-french}}
\paragraph{Description.}
This task evaluates broad academic knowledge and language understanding on French MMLU. This task uses the French split from the multilingual MMMLU release. The evaluation uses \texttt{openai/MMMLU}, the \texttt{test} split, language coverage French, and 14042 samples. System inputs include \texttt{subject}, \texttt{question}, \texttt{choices}, and \texttt{choice\_labels}. System outputs include \texttt{answer\_pred}. The metric used for reporting in this paper is \texttt{accuracy}.

\paragraph{Paper: } \cite{hendrycks2021mmlu,openai2024mmmlu}

\paragraph{Example prompt template(s).} ~

\prompttemplate{tasks/mmlu_mcqa_french.sample.json}

\subsection{\texttt{multi\_\-if\_\-instruction\_\-following}}
\paragraph{Description.}
This task multi-IF was introduced as a multi-turn, multilingual extension of IFEval. In this evaluation suite, we use the English/French instruction-following setting as a compact test of whether a model respects verifiable user constraints across languages. The evaluation uses \texttt{Multi-IF}, the \texttt{test} split, language coverage English, French, and 808 samples. System inputs include \texttt{prompt}, \texttt{language}, \texttt{turn\_index}, and \texttt{instruction\_id\_list}. System outputs include \texttt{response}. The metric used for reporting in this paper is \texttt{accuracy}. Other computed metrics are logged for diagnostic analysis and include \texttt{accuracy\_by\_language}.

\paragraph{Paper: } \cite{he2024multiif}

\paragraph{Example prompt template(s).} ~

\prompttemplate{tasks/multi_if_instruction_following.sample.json}

\subsection{\texttt{personachat\_\-conversation\_\-coherence}}
\paragraph{Description.}
This task tests persona consistency across turns, context awareness, and language respect (English). The evaluation uses \texttt{bavard/personachat\_truecased}, the \texttt{validation} split, and 500 samples. System inputs include \texttt{history} and \texttt{current\_utterance}. System outputs include \texttt{response}. The metric used for reporting in this paper is \texttt{overall\_coherence\_llm}. Other computed metrics are logged for diagnostic analysis and include \texttt{persona\_consistency\_llm} and \texttt{context\_awareness\_llm}.

\paragraph{Paper: } \cite{zhang2018personachat}

\paragraph{Example prompt template(s).} ~

\prompttemplate{tasks/personachat_conversation_coherence.sample.json}

\subsection{\texttt{ruler\_\-mrqa}}
\paragraph{Description.}
This task rULER Benchmark for Long-Context Language Models. Implements synthetic tasks from RULER (NVIDIA) including nIAH variations (single, multikey, multivalue, multiquery), variable tracking (multi-hop tracing), and common words extraction (aggregation). The evaluation uses 60 samples. System inputs include \texttt{context}, \texttt{question}, \texttt{task\_type}, and \texttt{context\_size}. System outputs include \texttt{answer}. The metric used for reporting in this paper is \texttt{exact\_match}. Other computed metrics are logged for diagnostic analysis and include \texttt{contains\_answer} and \texttt{f1\_score}.

\paragraph{Paper: } \cite{hsieh2024ruler}

\paragraph{Example prompt template(s).} ~

\prompttemplate{tasks/ruler_mrqa.sample.json}

\subsection{\texttt{toxigen\_\-implicit\_\-hate\_\-detection}}
\paragraph{Description.}
This task evaluates the ability of a language model to detect implicit hate speech and toxic content. The model receives a text and must classify it as toxic or non-toxic. The evaluation uses 940 samples. System inputs include \texttt{text} and \texttt{target\_group}. System outputs include \texttt{is\_toxic\_pred}. The metric used for reporting in this paper is \texttt{accuracy}. Other computed metrics are logged for diagnostic analysis and include \texttt{f1} and \texttt{precision}.

\paragraph{Paper: } \cite{hartvigsen2022toxigen}

\paragraph{Example prompt template(s).} ~

\prompttemplate{tasks/toxigen_implicit_hate_detection.sample.json}

\subsection{\texttt{webnlg\_\-qa\_\-conversational\_\-mrqa}}
\paragraph{Description.}
This task consists in answering questions given a source document in a conversational setting. Questions, history of questions-answers, documents and answers are in English. The evaluation uses the public WebNLG-QA dataset, the \texttt{challenge} split, and 232 samples. System inputs include \texttt{question}, \texttt{history}, and \texttt{document}. System outputs include \texttt{answer}. The metric used for reporting in this paper is \texttt{llm\_good}. Other computed metrics are logged for diagnostic analysis and include \texttt{bleu}, \texttt{meteor}, and \texttt{llm\_mos}.

\paragraph{Paper: } \cite{lecorve2022sparql2text}

\paragraph{Example prompt template(s).} ~

\prompttemplate{tasks/webnlg_qa_conversational_mrqa.sample.json}
lecorve2022sparql2text

\subsection{\texttt{wildjailbreak\_\-harmful\_\-request\_\-refusal}}
\paragraph{Description.}
This task evaluates whether a language model appropriately refuses harmful requests while not over-refusing benign requests. Uses the WildJailbreak evaluation methodology with LLM-as-judge for robust detection of jailbreak attempts. The evaluation uses 500 samples. System inputs include \texttt{prompt}. System outputs include \texttt{response}. The metric used for reporting in this paper is \texttt{accuracy\_llm}. Other computed metrics are logged for diagnostic analysis and include \texttt{attack\_success\_rate\_llm} and \texttt{over\_refusal\_rate\_llm}.

\paragraph{Paper: } \cite{jiang2024wildteaming}

\paragraph{Example prompt template(s).} ~

\prompttemplate{tasks/wildjailbreak_harmful_request_refusal.sample.json}

\subsection{\texttt{wolbanking77\_\-audio\_\-intent\_\-classification}}
\paragraph{Description.}
This task classifies banking customer queries from Wolof transcriptions into 10 intent categories. Supports two modes 'reference': Clean reference transcriptions and 'asr': Real ASR transcriptions (from wolof-lm/wolbanking77\_transcription-documents). The evaluation uses 641 samples. System inputs include \texttt{text} and \texttt{mode}. The metric used for reporting in this paper is \texttt{macro\_f1}. Other computed metrics are logged for diagnostic analysis and include \texttt{accuracy} and \texttt{micro\_f1}.

\paragraph{Paper: } \cite{kandji2026wolbanking}

\paragraph{Reference-transcription prompt template.} ~
\prompttemplate{tasks/wolbanking77_audio_reference_intent_classification.sample.json}

\paragraph{ASR-transcription prompt template.} ~
\prompttemplate{tasks/wolbanking77_audio_asr_intent_classification.sample.json}

\subsection{\texttt{wolbanking77\_\-audio\_\-transcription\_\-correction}}
\paragraph{Description.}
This task corrects real Wolof ASR transcriptions to match clean references. Uses WER/CER metrics before and after correction. The evaluation uses 641 samples. System inputs include \texttt{asr\_text}. System outputs include \texttt{corrected\_text}. The metric used for reporting in this paper is \texttt{wer\_relative\_difference\_mean}. Other computed metrics are logged for diagnostic analysis and include \texttt{wer\_before\_mean} and \texttt{wer\_after\_mean}.

\paragraph{Paper: } \cite{kandji2026wolbanking}

\paragraph{Example prompt template(s).} ~

\prompttemplate{tasks/wolbanking77_audio_transcription_correction.sample.json}

\subsection{\texttt{wolbanking77\_\-intent\_\-classification}}
\paragraph{Description.}
This task classifies banking customer queries into 77 intent categories. Supports Wolof, French, and English via language parameter. The evaluation uses \texttt{karim155/WolBanking77}, the \texttt{test} split, and 500 samples. System inputs include \texttt{utterance}. The metric used for reporting in this paper is \texttt{macro\_f1}. Other computed metrics are logged for diagnostic analysis and include \texttt{accuracy} and \texttt{micro\_f1}.

\paragraph{Paper: } \cite{kandji2026wolbanking}

\paragraph{English intent-classification prompt template.} ~
\prompttemplate{tasks/wolbanking77_english_intent_classification.sample.json}

\paragraph{French intent-classification prompt template.} ~
\prompttemplate{tasks/wolbanking77_french_intent_classification.sample.json}

\paragraph{Wolof intent-classification prompt template.} ~
\prompttemplate{tasks/wolbanking77_wolof_intent_classification.sample.json}

\subsection{\texttt{wolbanking77\_\-translation}}
\paragraph{Description.}
This task translates banking customer queries between Wolof, French, and English. Uses BLEU, METEOR, and optionally LLM-as-judge (when target != Wolof). The evaluation uses \texttt{karim155/WolBanking77}, the \texttt{test} split, and 500 samples. System inputs include \texttt{source\_text}, \texttt{source\_lang}, and \texttt{target\_lang}. System outputs include \texttt{translation\_pred}. The metric used for reporting in this paper is \texttt{meteor}. Other computed metrics are logged for diagnostic analysis and include \texttt{bleu}.

\paragraph{Paper: } \cite{kandji2026wolbanking}

\paragraph{English to French prompt template.} ~
\prompttemplate{tasks/wolbanking77_english_to_french_translation.sample.json}

\paragraph{English to Wolof prompt template.} ~
\prompttemplate{tasks/wolbanking77_english_to_wolof_translation.sample.json}

\paragraph{French to English prompt template.} ~
\prompttemplate{tasks/wolbanking77_french_to_english_translation.sample.json}

\paragraph{French to Wolof prompt template.} ~
\prompttemplate{tasks/wolbanking77_french_to_wolof_translation.sample.json}

\paragraph{Wolof to English prompt template.} ~
\prompttemplate{tasks/wolbanking77_wolof_to_english_translation.sample.json}

\paragraph{Wolof to French prompt template.} ~
\prompttemplate{tasks/wolbanking77_wolof_to_french_translation.sample.json}

\end{document}